%% file: extended_craft.tex
\documentclass[10pt,journal,compsoc]{IEEEtran}

\usepackage{xcolor}
\usepackage{mathrsfs}
\usepackage{amsmath,amsfonts}
\usepackage{algorithmic}
\usepackage{array}
\usepackage{tikz}
\usepackage{textcomp}
\usepackage{bm}
\usepackage{arydshln}
\usepackage{pifont}
\usepackage{ragged2e}
\usepackage[linesnumbered,ruled]{algorithm2e}
\usepackage[caption=false]{subfig}
\usepackage{verbatim}
\usepackage{amssymb}
\usepackage{booktabs}
\usepackage{multirow}
\usepackage{makecell}
\usepackage{enumerate}
\usepackage{colortbl}
\usepackage{tabu}
\usepackage{mathpazo}
\usepackage{overpic}
\usepackage{breakurl}
\usepackage[nocompress]{cite}
\usepackage{graphicx}

\definecolor{light-gray}{gray}{0.9}
\definecolor{hotgreen}{RGB}{211,84,0}
\usepackage[colorlinks,urlcolor=red,citecolor=hotgreen]{hyperref}

\graphicspath{{./figs/}}
\DeclareGraphicsExtensions{.pdf,.jpeg,.png,.pdf}

\newcommand{\etal}{et al.}
\newcommand{\eg}{\textit{e.g.}} 

\makeatletter

\begin{document}
\title{{Exploring Frequency-Inspired Optimization in Transformer for Efficient Single Image Super-Resolution}}
\author{Ao~Li,~Le Zhang,~\IEEEmembership{Member,~IEEE},~Yun Liu,~and~Ce Zhu,~\IEEEmembership{Fellow,~IEEE}

\IEEEcompsocitemizethanks{
\IEEEcompsocthanksitem This work was supported by the National Natural Science Foundation of China (NSFC) under Grant 62020106011 and the Key Program for International Cooperation of Ministry of Science and Technology of China (No.2024YFE0100700).
\IEEEcompsocthanksitem Ao Li, Le Zhang and Ce Zhu are with the School of Information and Communication Engineering, University of Electronic Science and Technology of China, Chengdu 611731, Sichuan, China. (E-mail: aoli@std.uestc.edu.cn; lezhang@uestc.edu.cn; eczhu@uestc.edu.cn). Corresponding authors: Le Zhang and Ce Zhu.
\IEEEcompsocthanksitem Yun Liu is with the College of Computer Science, Nankai University, Tianjin 300350, China. (E-mail: vagrantlyun@gmail.com)
}}

\IEEEtitleabstractindextext{
\begin{abstract}
\justifying
{Transformer-based methods have exhibited remarkable potential in single image super-resolution (SISR) by effectively extracting long-range dependencies. However, most of the current research in this area has prioritized the design of transformer blocks to capture global information, while overlooking the importance of incorporating high-frequency priors, which we believe could be beneficial. In our study, we conducted a series of experiments and found that transformer structures are more adept at capturing low-frequency information, but have limited capacity in constructing high-frequency representations when compared to their convolutional counterparts.} Our proposed solution, the \textbf{c}ross-\textbf{r}efinement \textbf{a}daptive \textbf{f}eature modulation \textbf{t}ransformer (\textbf{CRAFT}), integrates the strengths of both convolutional and transformer structures. It comprises three key components: the high-frequency enhancement residual block (\textbf{HFERB}) for extracting high-frequency information, the shift rectangle window attention block (\textbf{SRWAB}) for capturing global information, and the hybrid fusion block (\textbf{HFB}) for refining the global representation. {To tackle the inherent intricacies of transformer structures, we introduce a frequency-guided post-training quantization (PTQ) method aimed at enhancing CRAFT's efficiency. These strategies incorporate adaptive dual clipping and boundary refinement. To further amplify the versatility of our proposed approach, we extend our PTQ strategy to function as a general quantization method for transformer-based SISR techniques. Our experimental findings showcase CRAFT's superiority over current state-of-the-art methods, both in full-precision and quantization scenarios. These results underscore the efficacy and universality of our PTQ strategy.} The source code is available at: \url{https://github.com/AVC2-UESTC/Frequency-Inspired-Optimization-for-EfficientSR.git}.
\end{abstract}

}

\maketitle
\IEEEdisplaynontitleabstractindextext
\IEEEpeerreviewmaketitle
\IEEEraisesectionheading{
\section{Introduction}
\label{sec:introduction}}
\IEEEPARstart{S}{ingle} image super-resolution (SISR) has garnered significant attention in recent years due to its promising applications across various fields such as enhancing surveillance videos and medical images~\cite{mudunuri2015low, greenspan2009super}, restoring old photographs~\cite{liang2021swinir, Ledig2017}, and enabling efficient image transmission~\cite{Zhang2017}. Despite its widespread applications, SISR poses a challenging problem: reconstructing a high-resolution image from a low-resolution counterpart involves a complex mapping with many possible high-resolution outputs.

In recent years, deep learning has shown remarkable success in SISR~\cite{dong2016srcnn, zhang2021edge, Liang2021}, thanks to its powerful expressive capabilities. Specifically, the introduction of convolutional neural networks (CNNs) that employ innovative architectures like residual and dense connections has significantly improved reconstruction accuracy~\cite{ledig2017photo, Tong2017iccv}. Moreover, integrating attention mechanisms, which effectively combine spatial and channel information, has further advanced SISR~\cite{Zhang2018ECCV,Xia2022,Mei2021cvpr,Mei2020cvpr}. More recently, the emergence of transformer architectures has been pivotal, demonstrating their superior ability to capture long-range dependencies and setting new benchmarks in the SISR field~\cite{liang2021swinir,Chen2022nips,chen2021pre,li2021efficient,lu2022transformer}. {However, despite these significant advancements, existing methods in SISR are still lacking. Firstly, while these methods have effectively integrated off-the-shelf advanced network structures, such as designing transformer blocks for capturing global information or complex convolutional structures for extracting discriminative features, they often overlook the impact of frequency information in the input image on the overall performance of different architectures. Therefore, we believe that a comprehensive analysis from the frequency domain could be beneficial. Secondly, the computational demands of these transformer-based SISR methods pose a significant challenge to their practical deployment, particularly in scenarios with limited computational resources.}

{Our preliminary work, presented at ICCV 2023~\cite{li2023feature}, has initiated efforts to address the first  challenge by investigating the impact of frequency information on the performance of both CNN and transformer structures in SISR.} We achieve this by discarding different ratios of frequency components from the input image and observing the corresponding performance changes. Our empirical findings indicate that transformer models exhibit limited capability in constructing high-frequency representations when compared to CNNs. To mitigate this gap, we introduce the cross-refinement adaptive feature modulation transformer (\textbf{CRAFT}), an innovative model that effectively harnesses the strengths of both CNNs and transformers. It comprises three key components: the high-frequency enhancement residual block (\textbf{HFERB}) for extracting high-frequency information, the shift rectangle window attention block (\textbf{SRWAB}) for capturing global information, and the hybrid fusion block (\textbf{HFB}) for refining the global representation. {In this paper, we enhance the analysis by delving deeper into insights and offering more intuitive explanations, thereby providing a clearer understanding of the frequency-dependent performance in both CNN and transformer structures. Furthermore, we illustrate the superiority of our strategies in reconstructing high-frequency components compared to other state-of-the-art methods. These findings not only reinforce our prior contributions but also broaden the scope and applicability of our research in the realm of image super-resolution.}

\begin{table}[!t]
\caption{{Comparison of training efficiency among various QAT and PTQ quantization methods for the EDSR model, where ``FP" denotes the full-precision model, and ``GT" indicates the use of ground truth. Notably, the {$\dag$} symbol emphasizes that our PTQ method is specifically designed for our CRAFT and features an exceptionally low training cost.}}
\vspace{-2mm}
\centering
\scriptsize
\setlength{\tabcolsep}{1.1mm}
\begin{tabular}{ccccccc}
  \toprule
  Model & Type & Data (Resolution) & GT & BatchSize & Iters & Run Time \\
  \midrule
  EDSR~\cite{Lim2017} & FP & 800 ($2048\times1080$) & \ding{51} & 16 & 15,000 & 240$\times$ \\
  \cdashline{1-7}
  PAMS~\cite{li2020pams} & QAT & 800 ($2048\times1080$) & \ding{51} & 16 & 1,500 & 24$\times$ \\
  FQSR~\cite{wang2021fully} & QAT & 800 ($2048\times1080$) & \ding{51} & 16 & 15,000 & 120$\times$ \\
  CADyQ~\cite{hong2022cadyq} & QAT & 800 ($2048\times1080$) & \ding{51} & 8 & 30,000 & 240$\times$ \\
  DAQ~\cite{hong2022daq} & QAT & 800 ($2048\times1080$) & \ding{51} & 4 & 300,000 & 1200$\times$ \\
  DDTB~\cite{zhong2022dynamic} & QAT & 800 ($2048\times1080$) & \ding{51} & 16 & 3,000 & 48$\times$ \\
  PTQ4SR~\cite{tu2023toward} & PTQ & 100 ($2048\times1080$) & \ding{55} & 2 & 500 & 1$\times$ \\
  \cdashline{1-7}
  Ours{$^\dag$} & PTQ & 100 ($120\times120$) & \ding{55} & 2 & 500 & 1$\times$ \\
  \bottomrule
\end{tabular}
\vspace{-2mm}
\label{tab:preliminary}
\end{table}

{To further enhance the efficiency of CRAFT and tackle the second challenge, in this paper we adopt quantization strategies that compress the network into low-bit representations for both weights and activations, while maintaining the original architecture. Two primary quantization approaches are considered: quantization-aware training (QAT) and post-training quantization (PTQ). QAT typically demands large datasets and substantial computational resources, whereas PTQ offers a more efficient solution, requiring only a small number of unlabeled calibration images (100 samples) for rapid deployment across various devices, as depicted in Table~\ref{tab:preliminary}. More specifically, we calculate the ``Run Time" of different quantization methods by multiplying the batch size with the number of iterations ( ``batch size $\times$ iterations"). 
Compared to the QAT method, PTQ necessitates fewer inferences, leading to greater gains in training efficiency. Additionally, in terms of the actual number of the required training samples, PTQ also demands far fewer samples than QAT. }

{Given our primary focus on devising efficient SISR methods for both training and inference, we introduce a frequency-guided PTQ strategy tailored specifically for CRAFT. This strategy involves imposing additional frequency constraints on HFERB features to safeguard the preservation of high-frequency information. Our approach integrates a dual strategy, wherein feature boundaries are adaptively calibrated and refined using a limited set of calibration images, thus enhancing quantization efficacy. Furthermore, considering that the existing PTQ super-resolution method, as outlined in~\cite{tu2023toward}, primarily concentrates on CNN-based architectures, there exists a notable absence of PTQ implementation for transformer models. Consequently, we expand our PTQ strategy to a more general form. This strategy guarantees the retention of high image quality, even with restricted representations as low as 4 bits, signifying a noteworthy breakthrough in SISR.}

Our main contributions are summarized as follows:
\begin{itemize}
\item{\textbf{Frequency-perspective Analysis:} We study the impact of CNN and transformer structures on performance from a frequency perspective and observe that transformer is more effective in capturing low-frequency information while having limited capacity for constructing high-frequency representations compared to CNN.}

\item{\textbf{Frequency-aware SR Framework:} We employ parallel structures to extract varied frequency features: HFERB for high-frequency information essential for SISR, and SRWAB for global information. Our fusion strategy leverages HFERB as high-frequency prior and SRWAB output as key and value for inter-attention, enhancing overall performance.}

\item {\textbf{Frequency-guided PTQ Strategy:} We present a frequency-guided PTQ strategy tailored specifically for CRAFT. This involves imposing a frequency constraint on HFERB, along with an adaptive dual-phase calibration and refinement process. Moreover, we extend our PTQ strategy to cover transformer-based SISR methods, thus enhancing their practical applicability.}

\item \textbf{Comprehensive Experimental Validations:} We conduct meticulous experiments, including thorough ablation studies and comparative analyses across multiple datasets, to affirm the effectiveness of our techniques and scrutinize the impact of quantization on SISR outcome fidelity.
\end{itemize}

\section{Related Works}
\label{relatedWorks}
\subsection{CNN-based SISR}
Since the pioneering work SRCNN~\cite{dong2016srcnn} has achieved significant progress in SISR, various CNN-based works have been proposed. Kim~\etal~\cite{kim2016accurate} presented an SR method using deep networks by cascading 20 layers, demonstrating promising results. Building upon this, Lim~\etal~\cite{Lim2017} introduced the enhanced deep super-resolution (EDSR) network, which achieved a significant performance boost by removing the batch normalization layer~\cite{ioffe2015batch} from the residual block and incorporating additional convolution layers. Ahn~\etal~\cite{Ahn2018} designed an architecture with an increased number of residual blocks and dense connections, further improving the SR performance. In pursuit of lightweight models, Hui~\etal~\cite{Hui2019} proposed a selective fusion approach, employing cascaded information multi-distillation blocks to construct an efficient model. Li~\etal~\cite{Li2020} introduced a method involving predefined filters and used a CNN to learn coefficients, which were then linearly combined to obtain the final results. Sun~\etal~\cite{sun2022hybrid} proposed a hybrid pixel-unshuffled network (HPUN) by introducing an efficient and effective downsampling module into the SR task.

\newcommand{\AddImg}[1]{\includegraphics[width=.245\linewidth]{#1}}
\begin{figure*}[!t]
\centering
\subfloat[{Frequency attention bias of different structures.}]{
\AddImg{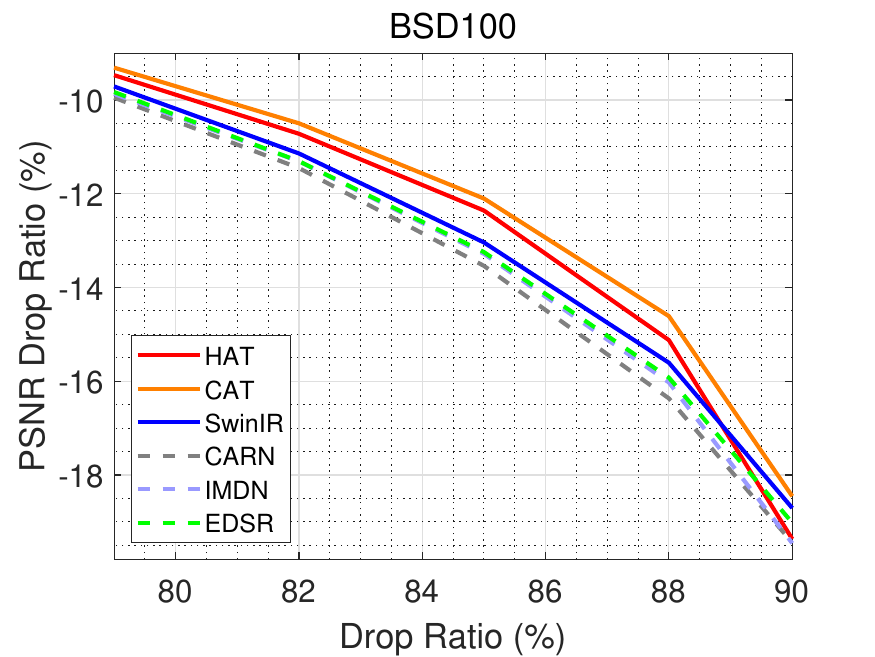} \hfill
\AddImg{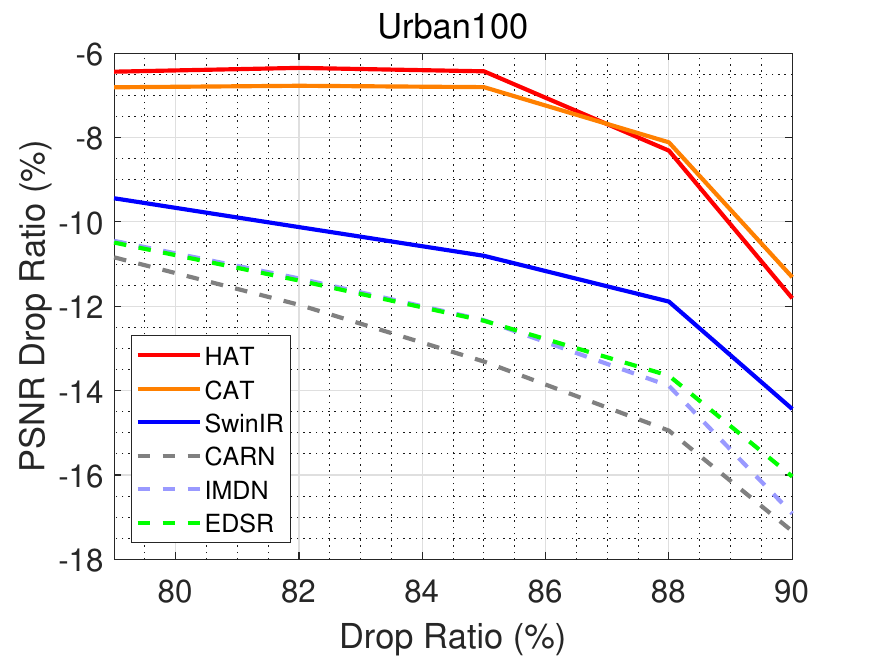} \hfill
\AddImg{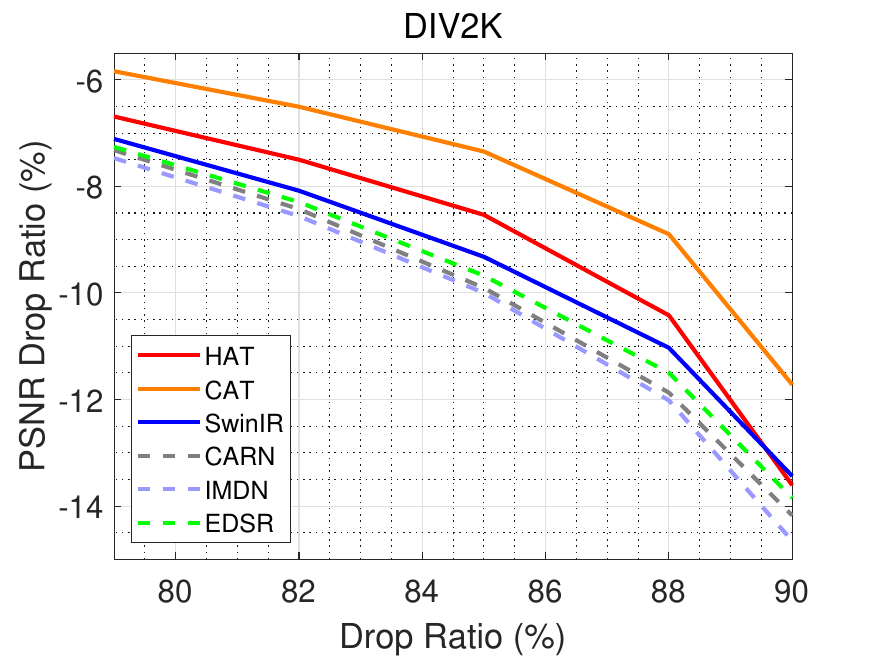} \hfill
\AddImg{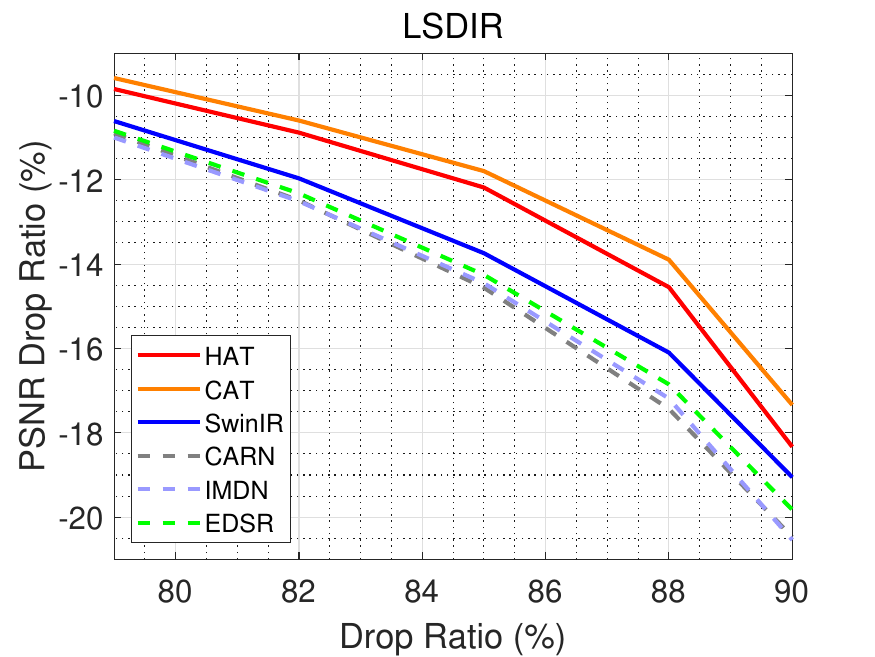} 
\label{subfig:dependence}
}\\
\subfloat[{Effectiveness of reconstructing high-frequency information.}]{
\AddImg{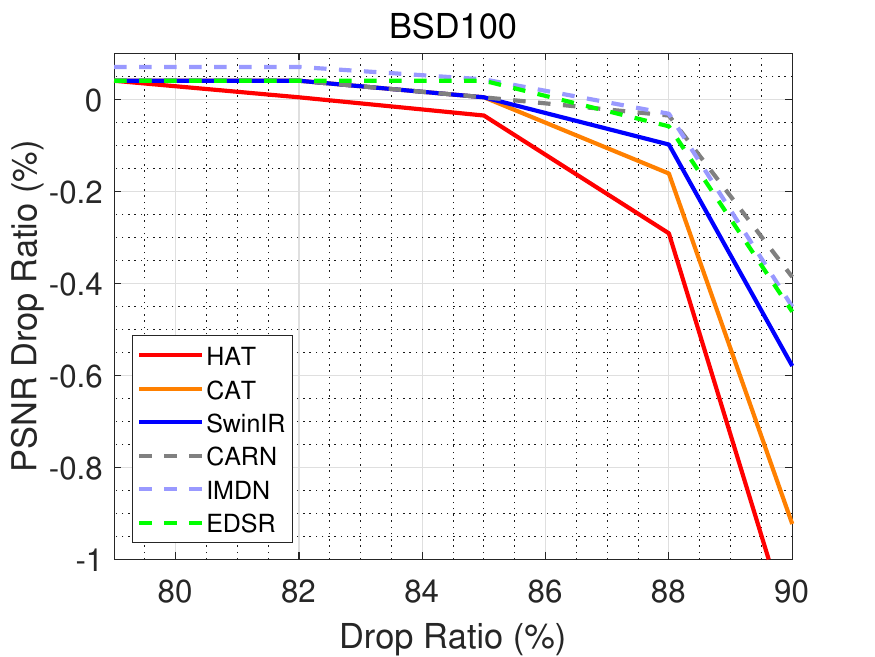} \hfill
\AddImg{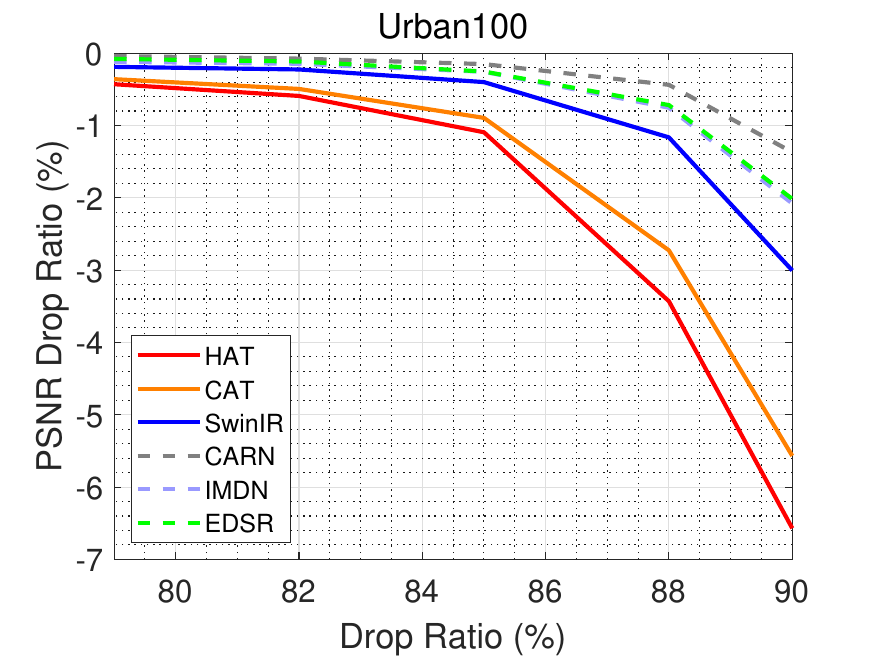} \hfill
\AddImg{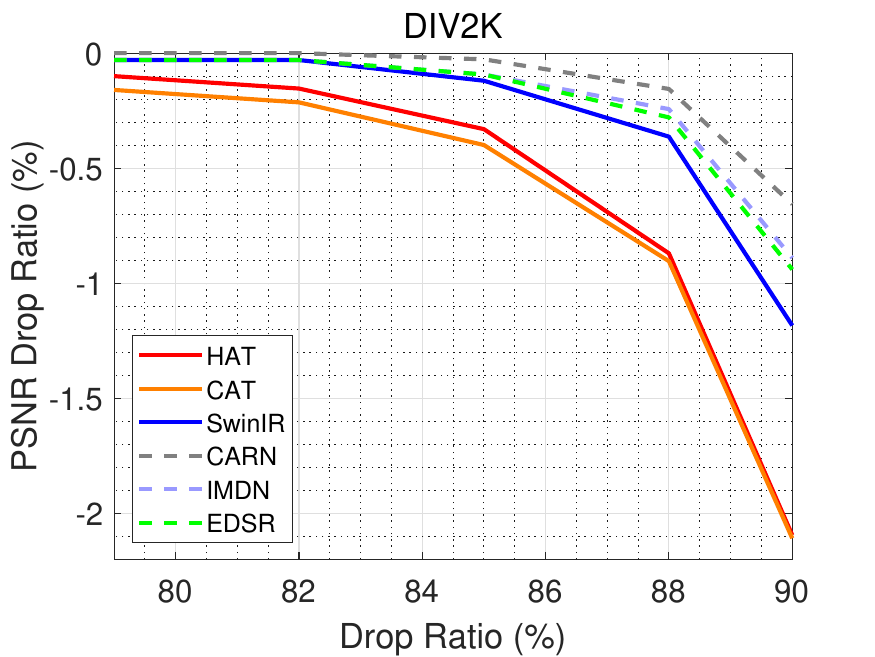} \hfill
\AddImg{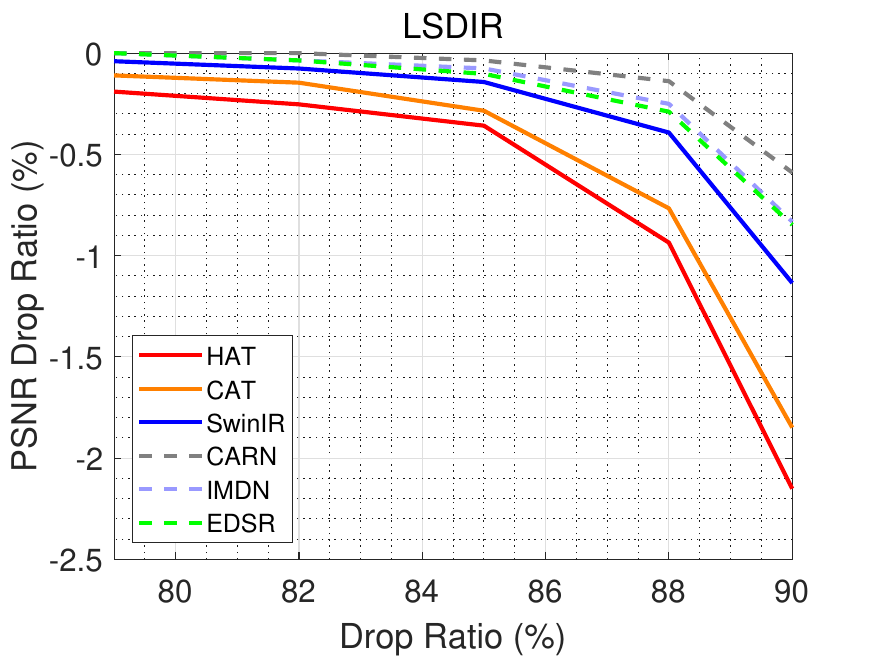}
\label{subfig:restoration}
}\\
\subfloat[{Procedure for dropping high-frequency components.}]{
\includegraphics[width=0.88\linewidth]{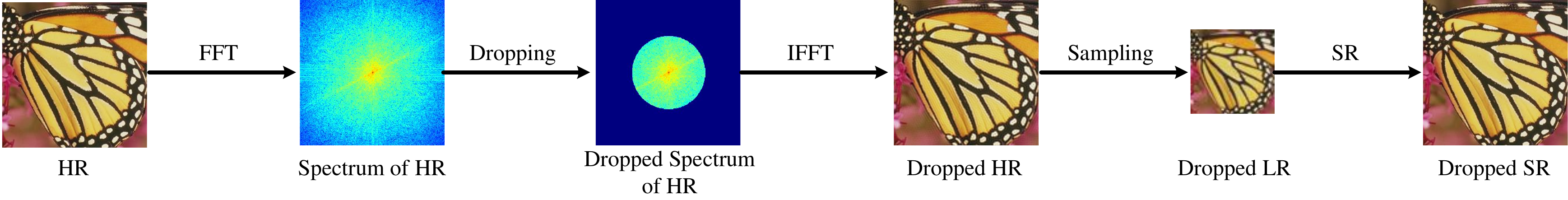}
\label{subfig:dropping}
}
\caption{{Influence of high-frequency information on the performance of CNN and transformer architectures. Dashed and solid lines correspond to CNN and transformer methods, respectively. (a) With an increase in the high-frequency drop ratio, transformer models exhibit a smaller change in PSNR compared to CNN, suggesting their superiority in capturing low-frequency information. (b) As the high-frequency drop ratio increases, transformer models show a more pronounced change in PSNR compared to CNN, indicating their limited ability to reconstruct high-frequency information from low-frequency.}}
\vspace{-3mm}
\label{fig:drop_procedure}
\end{figure*}

\subsection{Transformer-based SISR}
Liang~\etal~\cite{liang2021swinir} proposed SwinIR, a robust baseline model for image restoration, leveraging the Swin Transformer~\cite{liu2021swin}. CAT~\cite{Chen2022nips} modified the window shape and introduced a rectangle window attention to obtaining better performance. Chen~\etal~\cite{chen2021pre} proposed a pre-trained image processing transformer and showed that pre-trained mechanism could significantly improve the performance for low-level tasks. Li~\etal~\cite{li2021efficient} comprehensively analyzed the effect of pre-training and proposed a versatile model to tackle different low-level tasks. Lu~\etal~\cite{lu2022transformer} proposed a lightweight transformer to capture long-range dependencies between similar patches in an image with the help of the specially designed efficient attention mechanism. Zhang~\etal~\cite{zhang2022efficient} introduced a shift convolution and a group-wise multi-scale self-attention to reduce the complexity of transformers. HAT~\cite{chen2023activating} introduced a hybrid attention mechanism to enhance the performance of window-based transformers.

\subsection{Quantization in SISR}
Quantization plays a critical role in compressing and speeding up models, primarily divided into Quantization-aware Training (QAT) and Post-Training Quantization (PTQ). QAT fine-tunes weights and activations during training, which can enhance model performance despite requiring more training time. Ma~\etal~\cite{ma2019efficient}, explored the potential of weight binarization in SR, demonstrating the possibility of extreme compression. Similarly, BAM~\cite{xin2020binarized} introduced a bit accumulation mechanism, and PAMS~\cite{li2020pams} developed a learnable, layer-wise symmetric quantizer, both improving the precision of quantization processes. Aiming for further model compression, FQSR~\cite{wang2021fully} introduced a full-quantization approach for all layers of the SR network, using a learnable interval to retain performance. Additionally, DAQ~\cite{hong2022daq} examined a channel-wise distribution-aware quantization technique, and DDTB~\cite{zhong2022dynamic} presented a dynamic, dual trainable clip-value quantizer, effective for 2-4 bit model compression. 

In contrast to QAT, PTQ offers a fast and efficient way to quantization, only needing a small set of calibration data. This method is especially useful when there is limited access to the complete dataset or when quick model deployment is needed. PTQ4SR~\cite{tu2023toward} uses a two-step quantization approach specifically designed for the dynamic requirements of SR tasks. Yet, there is a notable gap in quantization methods, especially for transformer-based SR models. Our PTQ method can be expanded to a general quantization strategy for transformer-based SISR methods, promoting the development of more efficient and scalable SR solutions.

\section{{Frequency-perspective Analysis}}
\label{sec:motivation}
{In this section, we explore the performance of CNNs and transformers from a frequency perspective. To investigate the impact of different frequencies on these architectures, we conducted two sets of experiments using four commonly used benchmarks, as detailed in Sec.~\ref{sec:droppingprocess}. Additionally, we provide a more physically realistic analysis of the frequency-dropping method in Sec.~\ref{real-worldAnalysis}.}
\subsection{{Analysis of Frequency Impact}}
\label{sec:droppingprocess}

{For our study, we select CARN~\cite{Ahn2018}, IMDN~\cite{Hui2019}, and EDSR~\cite{Lim2017} as representative examples of CNN structures. For transformer structures, we examine SwinIR~\cite{liang2021swinir}, CAT~\cite{Chen2022nips}, and HAT~\cite{chen2023activating}. The process of dropping frequency components is depicted in Fig.~\ref{subfig:dropping}. Given a high-resolution (HR) image $X^{HR}$, we apply a Fast Fourier Transform (FFT) to obtain its frequency spectrum. In the frequency domain, the components are shifted such that the low frequencies are centered. We then calculate the Euclidean distance of each frequency component from the center of the spectrum. These distances are sorted, and based on the frequency discard ratio $\gamma$, where $0 \leq \gamma \leq 1$, we determine the number of frequencies to discard. Specifically, the most distant $\gamma \cdot L$ components (where $L$ is the total number of frequency components) are set to zero. Finally, we apply an Inverse Fast Fourier Transform (IFFT) to reconstruct the image with the high-frequency components dropped, referred to as $X_{drop}^{HR}(\gamma)$. The formulation for this process is as follows
\begin{equation}
    X_{drop}^{HR}(\gamma) = IFFT(Drop(D(FFT(X^{HR})),\gamma)),
    \label{eq:eq1}
\end{equation}
where $D(\cdot)$ represents the Euclidean distance calculation. Afterward, we downsample $X_{drop}^{HR}(\gamma)$ using bicubic interpolation to obtain the LR version $X_{drop}^{LR}(\gamma)$ (\eg~$\times4$ down-sampling). Finally, we employ CNN-based and transformer-based SR models to generate the super-resolved counterpart $X_{drop}^{SR}(\gamma)$.}

{To analyze the attention bias of different structures towards frequency components, we compute the peak signal-to-noise ratio (PSNR) $P^{D}(\gamma)$ between $X_{drop}^{SR}(\gamma)$ and $X_{drop}^{HR}$. We then plot the PSNR drop trend to visualize the difference between the two structures. As shown in Fig.~\ref{subfig:dependence}, the PSNR drop ratio for each drop ratio is defined as
\begin{equation}
    R_{drop}^{D}(\gamma) = \frac{P(0)-P^{D}(\gamma)}{P(0)},
    \label{eq:eq2}
\end{equation}
where $P(0)$ represents the PSNR without dropping, calculated between $X^{SR}$ and $X^{HR}$. The figures illustrate that the transformer model exhibits reduced sensitivity to high-frequency information and excels in capturing low-frequency information, as evidenced by the smaller PSNR change compared to the CNN model as the proportion of discarded high-frequency information increases.}

{Furthermore, we conduct another experiment to evaluate the effectiveness of different structures in reconstructing high-frequency information. Specifically, we calculate the PSNR $P^{E}(\gamma)$ between $X_{drop}^{SR}(\gamma)$ and $X^{HR}$ and plot the performance drop trend as previously depicted. The PSNR drop ratio for each drop ratio can be expressed as
\begin{equation}
R_{drop}^{E}(\gamma) = \frac{P^{E}(\gamma) - P(0)}{P(0)}.
\label{eq:eq3}
\end{equation}
From Fig.~\ref{subfig:restoration}, we observe that as the proportion of discarded high-frequency information increases, the transformer model experiences a larger PSNR change compared to the CNN model, indicating its limited ability to reconstruct high-frequency information from low-frequency.}

{Our empirical results reveal that, although transformers demonstrate impressive performance in tasks like super-resolution, they exhibit shortcomings in handling high-frequency information—critical for capturing fine details such as textures and edges. In contrast, CNNs, with their localized convolution operations, excel in processing high-frequency content more effectively. These findings suggest that transformers could benefit from the complementary strengths of CNNs to enhance their ability to recover intricate details.}

{To address this limitation, we propose a method that synergizes the strengths of both architectures. Specifically, we integrate CNN-generated information as a high-frequency prior to guide the transformer in refining its global representation. By capitalizing on the CNN’s proficiency in capturing high-frequency details and the transformer’s ability to model long-range dependencies, our approach yield improve results, as reported in} Section \ref{exp}.

\begin{figure}[!t]
\centering
\subfloat{
\includegraphics[width=0.94\linewidth]{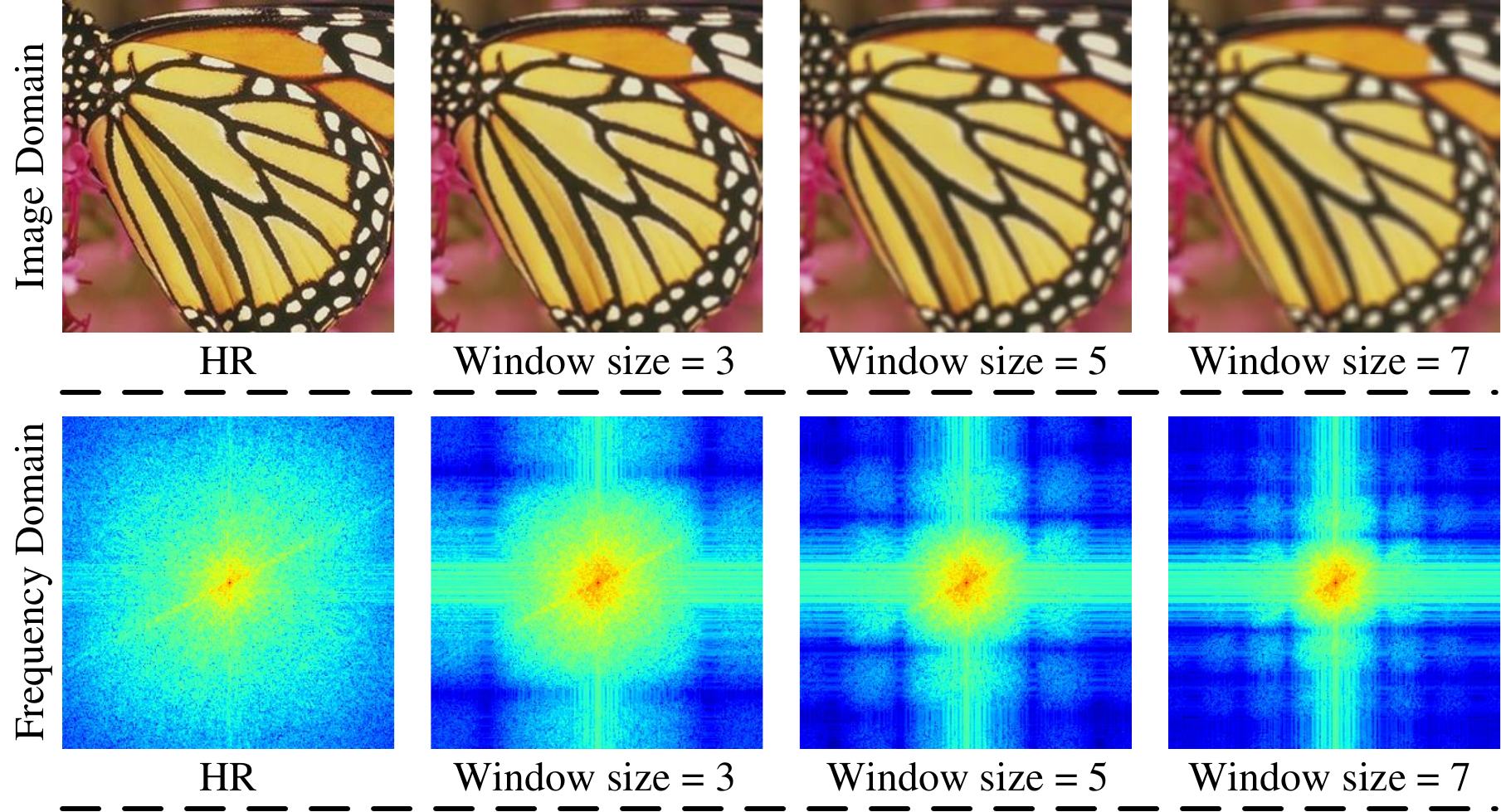} 
}
\\
\subfloat{
\includegraphics[width=0.485\linewidth]{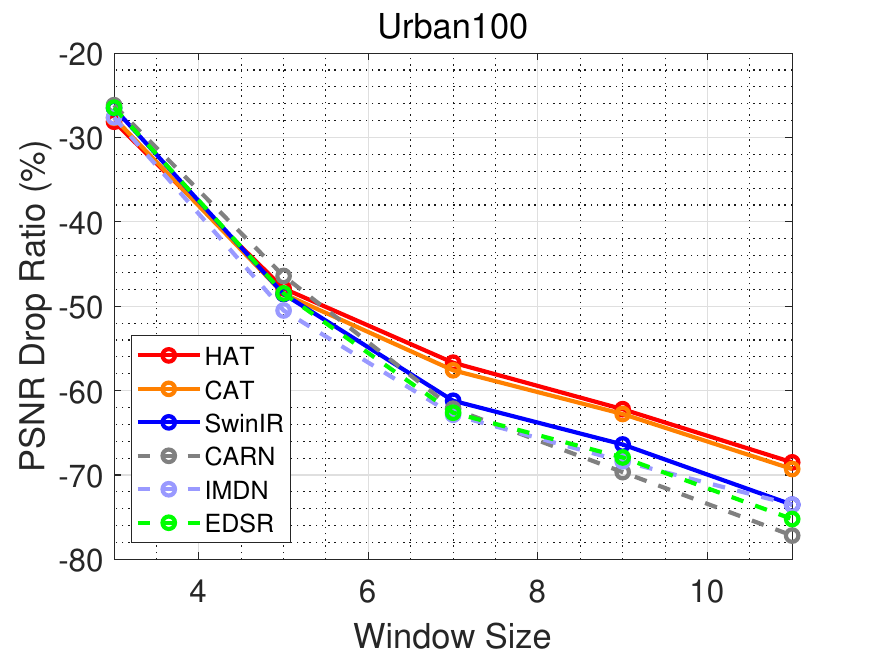} 
\includegraphics[width=0.485\linewidth]{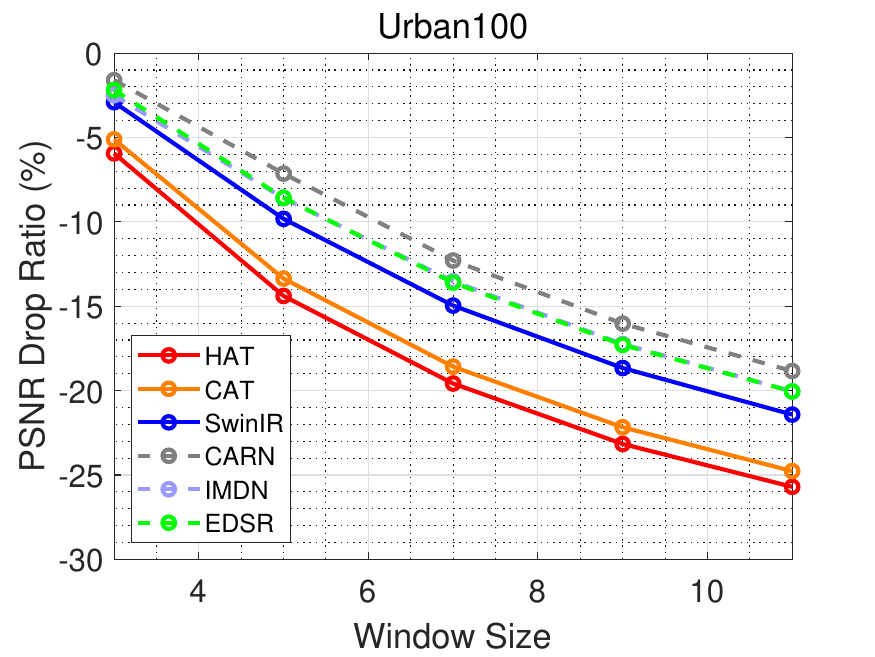} 
}
\vspace{-2mm}
\caption{{Effect of frequency dropping in the image domain using a mean filter.}}
\vspace{-4mm}
\label{fig:motivation_vis}
\end{figure}

\begin{figure*}[!t]
\centering
\begin{overpic}[scale=.33]{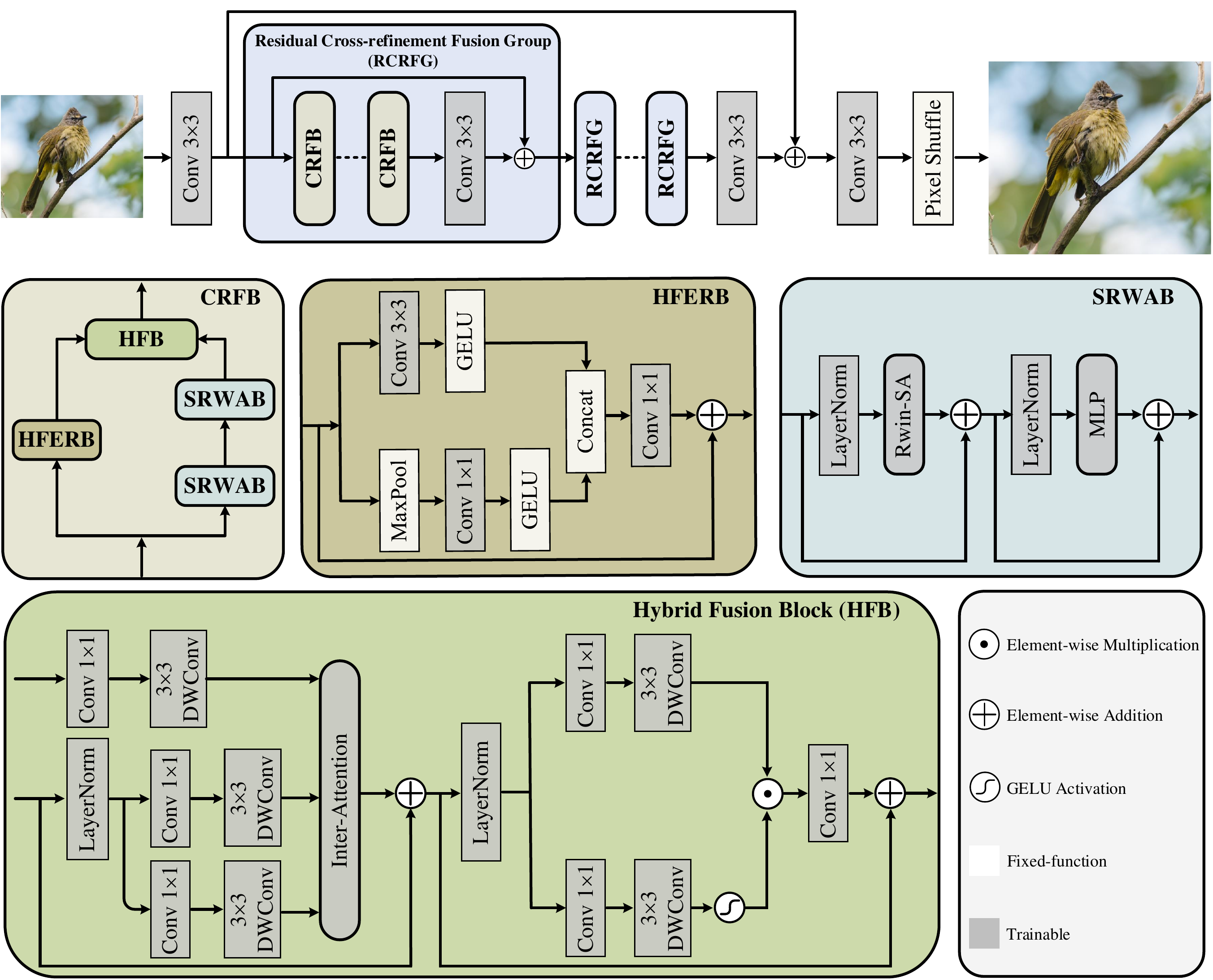}
\put(1.5,51.5){\footnotesize{$X_{\scriptscriptstyle H}$}}
\put(18.8,51.5){\footnotesize{$X_{\scriptscriptstyle S}$}}
\put(1.5,25.5){\footnotesize{$X_{\scriptscriptstyle H}$}}
\put(1.5,15.5){\footnotesize{$X_{\scriptscriptstyle S}$}}

\put(12,55){\footnotesize{C}}
\put(12,34){\footnotesize{C}}
\put(3,40){\footnotesize{C}}
\put(19,37){\footnotesize{C}}

\put(25.5,46.3){\footnotesize{C}}
\put(27.5,53){\footnotesize{C/2}}
\put(27.5,37.5){\footnotesize{C/2}}

\put(20.2,25.5){\footnotesize{${\boldsymbol{\hat{Q}}}$}}
\put(24,15.8){\footnotesize{${\boldsymbol{\hat{K}}}$}}
\put(24,6.3){\footnotesize{${\boldsymbol{\hat{V}}}$}}

\end{overpic}
\caption{{Framework of CRAFT. HFERB extracts the high-frequency information from the input features, SRWAB captures the long-range dependency of input features, and HFB integrates the output of HFERB and SRWAB to cross refine the global features. The reconstruction module employs a $3\times3$ convolutional layer to refine the features, and a shuffle layer~\cite{Shi2016} is used to obtain the final SR output. Best viewed in color.}}
\label{fig:model}
\vspace{-3mm}
\end{figure*}

\subsection{{Real-world Imaging System}}
\label{real-worldAnalysis}
{In real-world imaging systems, linear shift-invariant models are commonly used. The Point Spread Function (PSF) describes how the system responds to a point source of light, which results in blurring of the image~\cite{rossmann1969point,shechtman2014optimal}. After this blurring occurs, the signal is sampled by the sensors. This sampling process further influences the final captured image.}

{To provide a more physically realistic exploration, we conducted additional experiments using a mean filter, a linear operator, to better simulate real-world conditions. Specifically, we applied the mean filter to the input HR image. The process involves three steps. Initially, for a given HR image $X^{HR}$, we selected various window sizes $\theta\in\{3,5,7,9,11\}$ to reduce different levels of high-frequency components, leading to increased blurriness as the window size grows. The frequency dropping process can be expressed as:
\begin{equation}
X_{drop}^{HR}(\theta) = Filter(X^{HR}, \theta),
\label{eq:eq22}
\end{equation}
where $X_{drop}^{HR}(\theta)$ represents the filtered HR image. Subsequently, we downsampled $X_{drop}^{HR}(\theta)$ using bicubic interpolation to derive the LR counterpart $X_{drop}^{LR}(\theta)$. Finally, we employed CNN-based and transformer-based SR models to generate the super-resolved counterpart $X_{drop}^{SR}(\theta)$. The experimental results are presented in Fig.~\ref{fig:motivation_vis}. The top row shows the original HR image along with its filtered versions for window sizes $3$, $5$, and $7$. The second row illustrates the corresponding frequency domain representations. The bottom row presents the PSNR drop rate for various models on the Urban100 dataset. As demonstrated by this experiment, we reached the same conclusion as in Sec.~\ref{sec:droppingprocess}.}

\section{{Frequency-aware SR Framework}}
\label{sec:method}
The CRAFT network integrates three key modules: shallow feature extraction, residual cross-refinement fusion groups (RCRFGs), and a reconstruction module as shown in Fig.~\ref{fig:model}. The shallow feature extraction module consists of a single convolutional layer, while the reconstruction module is followed by SwinIR~\cite{liang2021swinir}. The RCRFG consists of several cross-refinement fusion blocks (CRFBs). Each CRFB includes three types of blocks: high-frequency enhancement residual blocks (HFERBs), shift rectangle window attention blocks (SRWABs), and hybrid fusion blocks (HFBs). We first describe the overall structures of CRAFT and then elaborate on the three key designs, including HFERB, SRWAB, and HFB.

\subsection{Model Overview}
The input LR image is processed by a $3\times3$ convolutional layer to obtain shallow features. These features are then fed into a serial of RCRFGs to learn deep features. After the last RCRFG, a $3\times3$ convolutional layer aggregates the features, and a residual connection is established between its output and the shallow features for facilitating training. The reconstruction module employs a $3\times3$ convolutional layer to refine the features, and a shuffle layer~\cite{Shi2016} is used to obtain the final SR output.

\subsection{High-Frequency Enhancement Residual Block}
The HFERB aims to enhance the high-frequency information, as shown in Fig.~\ref{fig:model}. It comprises the local feature extraction (LFE) branch and the high-frequency enhancement (HFE) branch. Specifically, we split the input features $F_{in}\hspace{-3pt}\in\hspace{-3pt}\mathbb{R}^{H\times W\times C}$ into two parts, and then processed by the two branches separately
\begin{equation}
F_{in}^{LFE},F_{in}^{HFE} = Split(F_{in}),
\label{eq:eq4}
\end{equation}
where $F_{in}^{LFE},F_{in}^{HFE}\hspace{-3pt}\in\hspace{-3pt}\mathbb{R}^{H\times W\times C/2}$ represent the input of LFE and HFE. For the LFE branch, we use a $3\times3$ convolutional layer followed by a GELU activation function to extract local high-frequency features
\begin{equation}
\hat{F}_{in}^{LFE} = f_{a}(Conv_{3\times3}(F_{in}^{LFE})),
\label{eq:eq5}
\end{equation}
where the $Conv_{3\times3}(\cdot)$ refers to the convolutional layer and the $f_{a}(\cdot)$ represents the GELU activation layer. For the HFE branch, we employ a max-pooling layer to extract high-frequency information from the input features $F_{in}^{HFE}$. Then, we use a $1\times1$ convolutional layer followed by a GELU activation function to enhance the high-frequency features,
\begin{equation}
\hat{F}_{in}^{HFE} = f_{a}(Conv_{1\times1}(MaxPooling(F_{in}^{HFE}))),
\label{eq:eq6}
\end{equation}
where the $Conv_{1\times1}(\cdot)$ indicates the convolutional layer, the $MaxPooling(\cdot)$ means the max-pooling layer and the $f_{a}(\cdot)$ represents the GELU activation layer. The outputs of the two branches are then concatenated and fed into a $1\times1$ convolutional layer to fuse the information thoroughly. To make the network benefit from multi-scale information and maintain training stability, a skip connection is introduced. The whole process can be formulated as
\begin{equation}
X_{H} = Conv_{1\times1}(Concat(\hat{F}_{in}^{LFE}, \hat{F}_{in}^{HFE})) + F_{in},
\label{eq:eq7}
\end{equation}
where the $Concat(\cdot)$ refers to the concatenation operation and the $Conv_{1\times1}(\cdot)$ represents the convolutional layer.

\subsection{Shift Rectangle Window Attention Block}
We use the shift rectangle window (SRWin) to expand the receptive field, which can benefit SISR~\cite{Chen2022nips}. Unlike square windows, the SRWin uses rectangle windows to capture more relevant information along the longer axis. In detail, given an input $X_{in}\hspace{-3pt}\in\hspace{-3pt}\mathbb{R}^{H\times W\times C}$, we divide it into $\frac{H\times W}{rh\times rw}$ rectangle windows, where $rh$ and $rw$ refer to the height and width of the rectangle window. For the $i$-th rectangle window feature $X_i\hspace{-3pt}\in\hspace{-3pt}\mathbb{R}^{(rh\times rw)\times C}$, we compute the \emph{query}, \emph{key}, and \emph{value} as follows
\begin{equation}
\boldsymbol{Q}_i = X_iW_i^Q, \boldsymbol{K}_i = X_iW_i^K, \boldsymbol{V}_i = X_iW_i^V,
\label{eq:eq8}
\end{equation}
where the $W_i^Q\hspace{-4pt}\in\hspace{-3.5pt}\mathbb{R}^{C\times d}$, $W_i^K\hspace{-4pt}\in\hspace{-3.5pt}\mathbb{R}^{C\times d}$ and $W_i^V\hspace{-4pt}\in\hspace{-3.5pt}\mathbb{R}^{C\times d}$ represent the projection matrices and $d$ is projection dimension which is commonly set to $d = \frac{C}{M}$ where the $M$ is the number of heads. The self-attention can be formulated as
\begin{equation}
Attention(\boldsymbol{Q}_i, \boldsymbol{K}_i, \boldsymbol{V}_i) \hspace{-3pt}=\hspace{-3pt} Softmax(\frac{\boldsymbol{Q}_i\boldsymbol{K}_i^T}{\sqrt{d}}\hspace{-3pt}+\hspace{-3pt}B)\boldsymbol{V}_i,
\label{eq:eq9}
\end{equation}
where $B$ is the dynamic relative position encoding~\cite{wang2021crossformericlr}. Moreover, a convolutional operation on the \emph{value} is introduced to enhance local extraction capability. 

To capture information from different axes, we use two types of rectangle windows: horizontal and vertical windows. Specifically, we split the attention heads into two equal groups and compute the self-attention within each group separately. We then concatenate the outputs of the two groups to obtain the final output. The procedure can be expressed as
\begin{equation}
Rwin\mbox{-}SA(X) = Concat(V\mbox{-}Rwin, H\mbox{-}Rwin)W^p,
\label{eq:eq10}
\end{equation}
where the $W^p\hspace{-3pt}\in\hspace{-3pt}\mathbb{R}^{C\times C}$ represents the linear projection to fuse the features, $V\mbox{-}Rwin$ and $H\mbox{-}Rwin$ indicate the vertical and horizontal rectangle window attention. {Unlike traditional methods that use attention masks to restrict computations within the same window, we empirically found that removing the mask significantly reduces computation time without negatively impacting performance, as indicated in Table~\ref{tab:ablationlmask}.} In addition, a multi-layer perceptron (MLP) is used for further feature transformations. The whole process can be formulated as
\begin{equation}
\begin{split}
   &X = Rwin\mbox{-}SA(LN(X_{in}))+X_{in}\\
   &X_{S} = MLP(LN(X)) + X,
   \label{eq:eq11}
\end{split}
\end{equation}
where the $LN$ represents the LayerNorm layer. 

\subsection{Hybrid Fusion Block}
To better integrate the merits of CNN and transformer (HFERB and SRWAB), we have designed a hybrid fusion block (HFB), which is illustrated in Fig.~\ref{fig:model}. We formulate the output of HFERB as the high-frequency prior \emph{query} and the output of SRWAB as \emph{key}, \emph{value} and calculate the inter-attention to refine the global features which are obtained from SRWAB. Moreover, most existing methods focus on spatial relations and overlook channel information. To overcome this limitation, we perform inter-attention based on the channel dimension to explore channel dependencies. In addition, this design will significantly reduce complexity. Traditional methods that use spatial attention tend to result in significant computational complexity (\eg, $O(N^2C), N\gg C$), where $N$ represents the length of the sequence and $C$ represents the channel dimension. In contrast, our channel attention design can transfer the quadratic component to the channel dimension (\eg, $O(NC^2)$), effectively reducing complexity.

Specifically, as shown in Fig.~\ref{fig:model}, we use a $1\times1$ convolutional layer followed by a $3\times3$ depth-wise convolutional layer to generate the high-frequency query $\boldsymbol{Q}\hspace{-3pt}\in\hspace{-3pt}\mathbb{R}^{H\times W\times C}$ based on the output of HFERB, $X_{H}$. As to the output of SRWAB, $X_{S}$, we first normalize the features by LayerNorm layer and then use the same operation as the query $\boldsymbol{Q}$ to get the key $\boldsymbol{K}\hspace{-3pt}\in\hspace{-3pt}\mathbb{R}^{H\times W\times C}$ and the value $\boldsymbol{V}\hspace{-4pt}\in\hspace{-3pt}\mathbb{R}^{H\times W\times C}$. Following the~\cite{Zamir2022}, we perform the reshape operation on $\boldsymbol{Q}$, $\boldsymbol{K}$ and $\boldsymbol{V}$ to get the $\boldsymbol{\hat{Q}}\hspace{-3pt}\in\hspace{-3pt}\mathbb{R}^{C\times(HW)}$, $\boldsymbol{\hat{K}}\hspace{-3pt}\in\hspace{-3pt}\mathbb{R}^{C\times(HW)}$ and $\boldsymbol{\hat{V}}\hspace{-3pt}\in\hspace{-3pt}\mathbb{R}^{C\times(HW)}$. After that, we compute the inter-attention as
\begin{equation}
Attention(\boldsymbol{\hat{Q}}, \boldsymbol{\hat{K}}, \boldsymbol{\hat{V}}) = Softmax(\frac{\boldsymbol{\hat{Q}}\boldsymbol{\hat{K}^T}}{\alpha})\boldsymbol{\hat{V}},
\label{eq:eq12}
\end{equation}
where the $\alpha$ represents the learnable parameter. Meanwhile, we add the refinement features to the $X_{S}$ to get the fusion output $X_{fuse}$. In addition, we feed $X_{fuse}$ to an improved feed-forward network~\cite{Zamir2022} to aggregate the features further. The details of this structure are shown in Fig.~\ref{fig:model}. It introduced a gate mechanism to fully extract the spatial and channel information and gain better performance. The whole process can be formulated as
\begin{equation}
\begin{split}
   X_{fuse} &\hspace{-3pt}=\hspace{-3pt} Inter\mbox{-}Atten(LN(X_{S}), X_{H})\hspace{-3pt}+\hspace{-3pt}X_{S}\\
   X_{HFB} &\hspace{-3pt}=\hspace{-3pt} IMLP(LN(X))\hspace{-3pt}+\hspace{-3pt} X_{fuse},
   \label{eq:eq13}
\end{split}
\end{equation}
where the $LN$ means LayerNorm operation, $IMLP$ represents the improved MLP, and $Inter\mbox{-}Atten$ indicates the proposed inter-attention mechanism, which introduces high-frequency prior to refining the global representations.

\section{{Frequency-guided PTQ Strategy}}
\label{sec:quantization}
\subsection{{Preliminary}}
{Quantizing a neural network means turning the float weights and activations into integers to boost efficiency. Suppose $X_i$ is an input with full precision. We define its quantized version as $X_i^{int}$ and its approximate float form as $\hat{X_i}$. When it comes to PTQ, the quantization process can be formulated as:}
{\begin{equation}
X_{i}^{int} = \emph{clip}\left(\left\lfloor\frac{X_i}{\emph{scale}}\right\rceil + \emph{zero point}; 0, 2^b - 1\right),
\label{eq:eq15}
\end{equation}
where $b$ is the number of bits used in quantization, $\lfloor \cdot \rceil$ indicates the rounding operation, and \emph{clip} is defined as:
\begin{equation}
\emph{clip}(x; a, c) = 
\begin{cases} 
a, & \text{if } x < a, \\
x, & \text{if } a \leq x \leq c, \\
c, & \text{if } x > c.
\end{cases}
\end{equation}
The $\emph{scale}$ and $\emph{zero point}$ parameters are essential for a precise quantization. They are calculated as:
\begin{equation}
\emph{scale} = \frac{u - l}{2^b - 1}, \quad \emph{zero point} = \emph{clip}\left(\left\lfloor-\frac{l}{\emph{scale}}\right\rceil; 0, 2^b - 1\right),
\label{eq:eq16}
\end{equation}
with $l$ and $u$ representing the minimal and maximal bounds of the input data. The dequantization process is given by:
\begin{equation}
\hat{X_{i}} = \emph{scale} \cdot (X_i^{int} - \emph{zero point}).
\label{eq:eq17}
\end{equation}
Since the minimal and maximal bounds ($l$ and $u$) determine the quantization scale and range, the appropriate setting of them is the key to preserving the integrity of the quantized data and minimizing performance loss.}

\begin{figure}[!t]
\centering
\subfloat{
\includegraphics[width=0.485\linewidth]{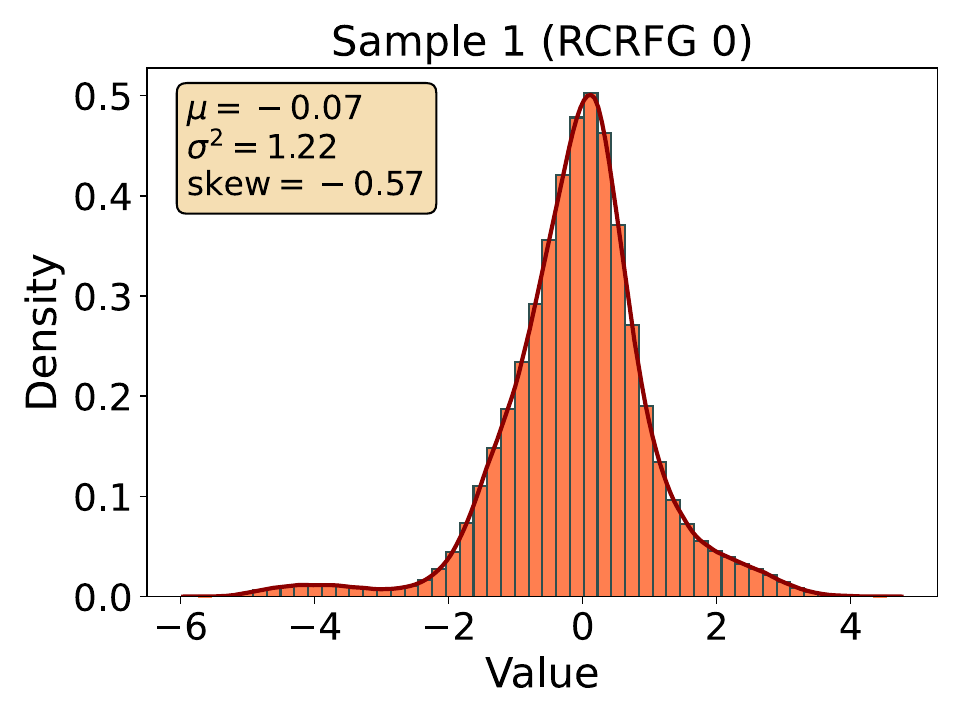} \hfill
\includegraphics[width=0.485\linewidth]{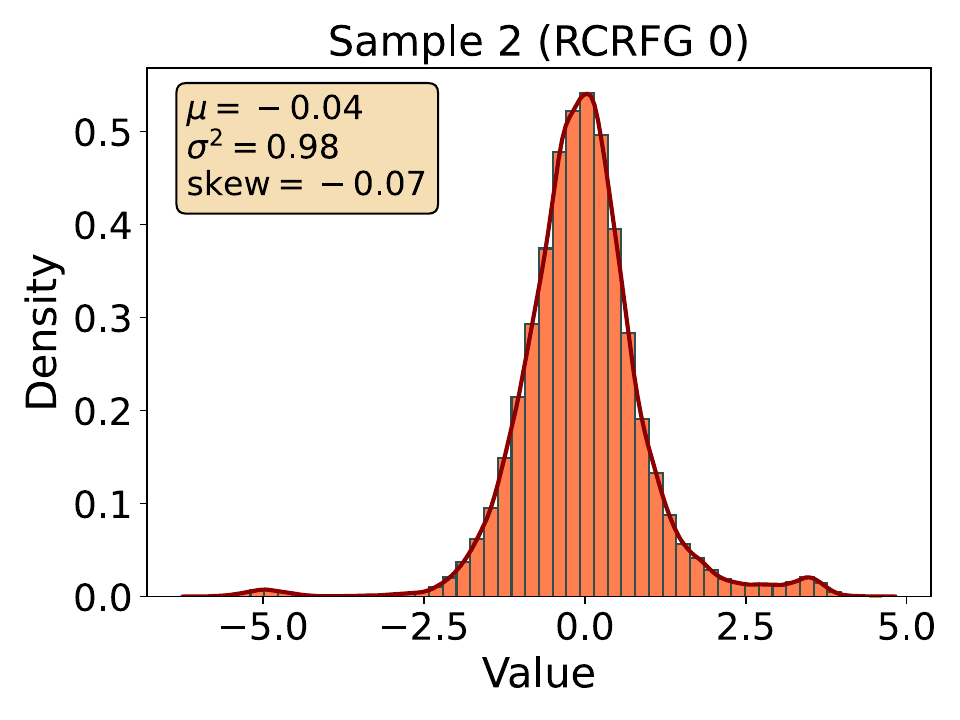}
}
\vspace{-3mm}
\caption{Illustration of asymmetric and high-dynamic phenomenon.}
\vspace{-1mm}
\label{fig:quant_pre}
\end{figure}

\begin{table}[!t]
\caption{Effects of different post-training quantization settings.}
\vspace{-2mm}
\centering
\footnotesize
\setlength{\tabcolsep}{1.5mm}
\begin{tabular}{cccccccc}
  \toprule
  \multirow{2}*{\vspace{-0.23cm}Model} & \multirow{2}*{\vspace{-0.23cm}FP32} & \multicolumn{3}{c}{Per-layer} & \multicolumn{3}{c}{Per-channel} \\
  \cmidrule(lr){3-5} \cmidrule(lr){6-8}
  &    &   W8A8   &  W4A8  &  W8A4  &  W8A8  &  W4A8  &  W8A4   \\
  \midrule
  CRAFT  &  32.52 &  32.39  & 30.18 & 22.48 & 32.38 & 30.26 & 27.96 \\
  \bottomrule
\end{tabular}
\vspace{-3mm}
\label{tab:quantObservation}
\end{table}

\subsection{{Effects of PTQ on CRAFT}}
{In this section, we analyze the effects of PTQ on the CRAFT through extensive experiments. By applying the MinMax\cite{jacob2018quantization} method as our PTQ strategy, we assess the effects of PTQ on CRAFT across various settings.}

From the perspective of granularity, we distinguish between per-layer and per-channel quantization in Table~\ref{tab:quantObservation}. Per-layer quantization applies a uniform scale across the entire layer, enhancing computational efficiency but potentially compromising precision. In contrast, per-channel quantization assigns a unique scale to each channel within a layer, offering higher granularity and accuracy at the expense of greater computational complexity. Despite the error-minimizing advantages of per-channel quantization, its substantial computational requirements make it less practical for deployment on source-constrained devices. This drives our inclination towards per-layer quantization in search of an optimal balance between computational efficiency and model accuracy.

From the perspective of quantizing weights and activations, as depicted in Table~\ref{tab:quantObservation}, reducing the precision of activation values to 4 bits leads to a notable decrease in the quality of image reconstruction, showing an average drop of 10.04 dB with per-layer quantization and 4.56 dB with per-channel quantization. This contrasts sharply with the lesser impact observed when quantizing weights, which results in a drop of 2.34 dB and 2.26 dB, respectively. {A closer look at the activation distributions within CRAFT, especially at the output of the first RCRFG block (referred to as RCRFG 0) as illustrated in Fig.~\ref{fig:quant_pre}, reveals considerable asymmetry and high dynamic. The activation histograms for RCRFG 0's output display pronounced skewness, indicating variable asymmetry across different input samples within the same layer.}

\renewcommand{\AddImg}[1]{\includegraphics[width=.3\linewidth]{#1}}
\begin{figure}[!t]
\centering
\subfloat[Full-precise]{
\AddImg{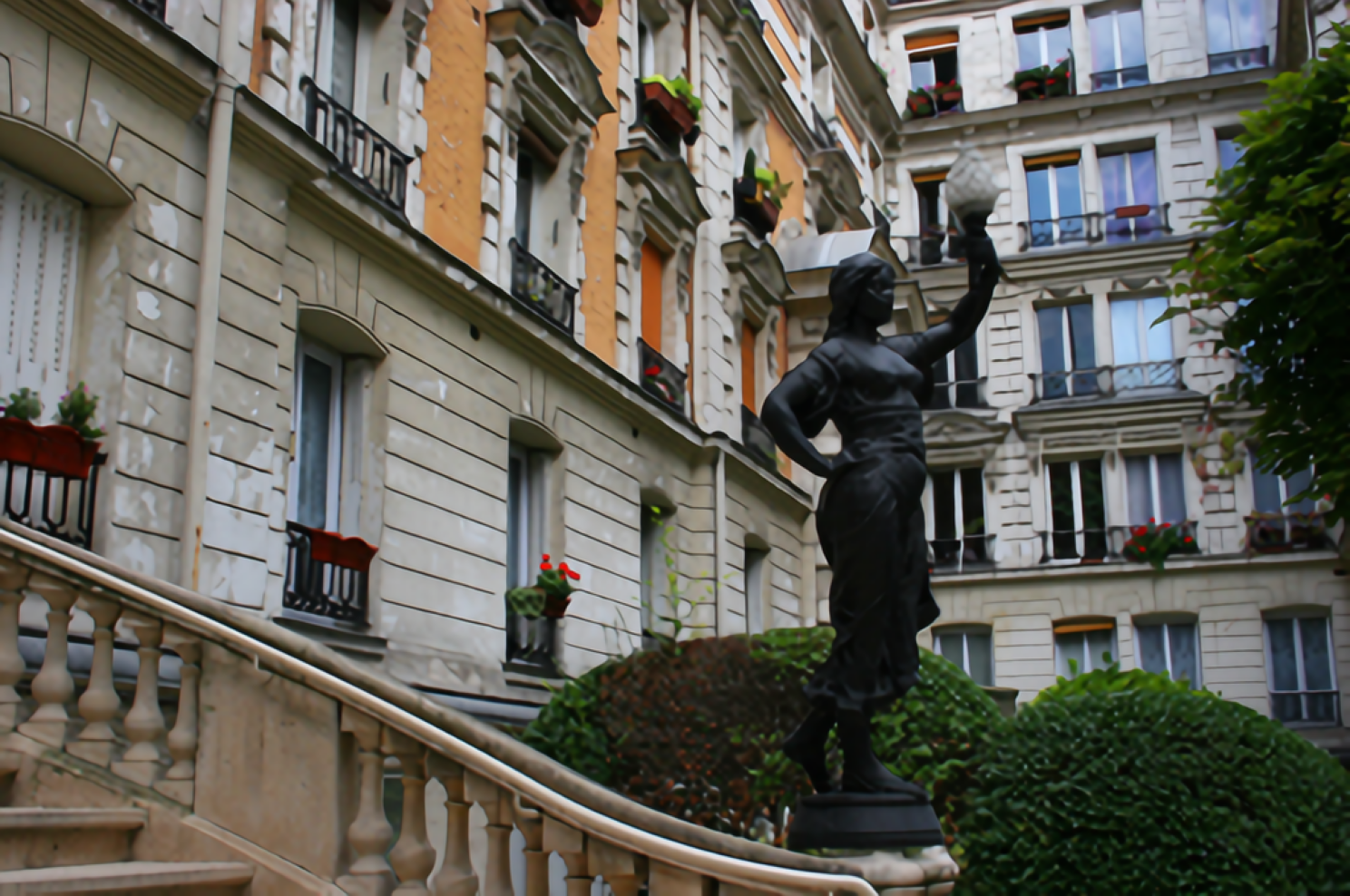}
\AddImg{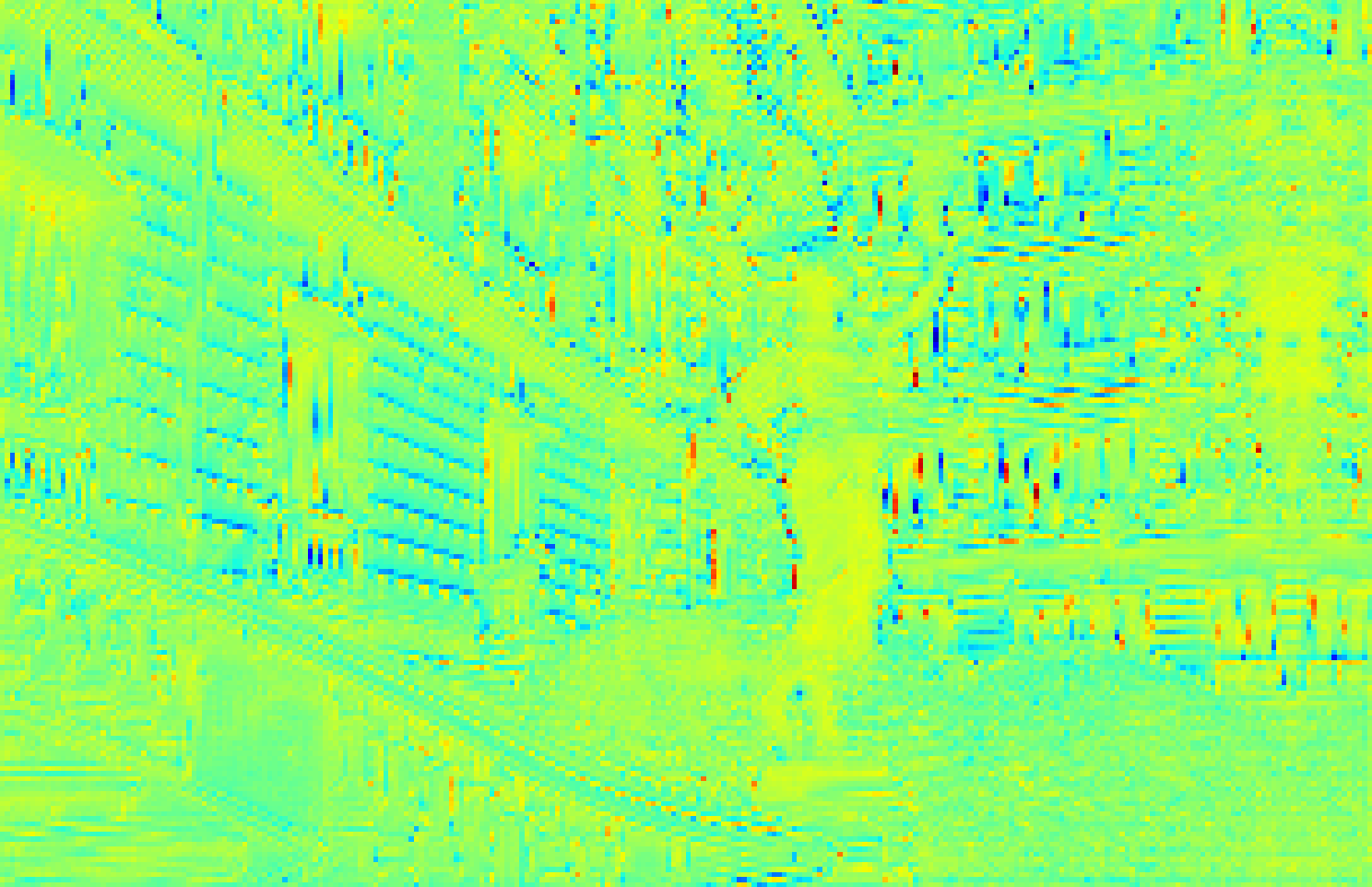}
\AddImg{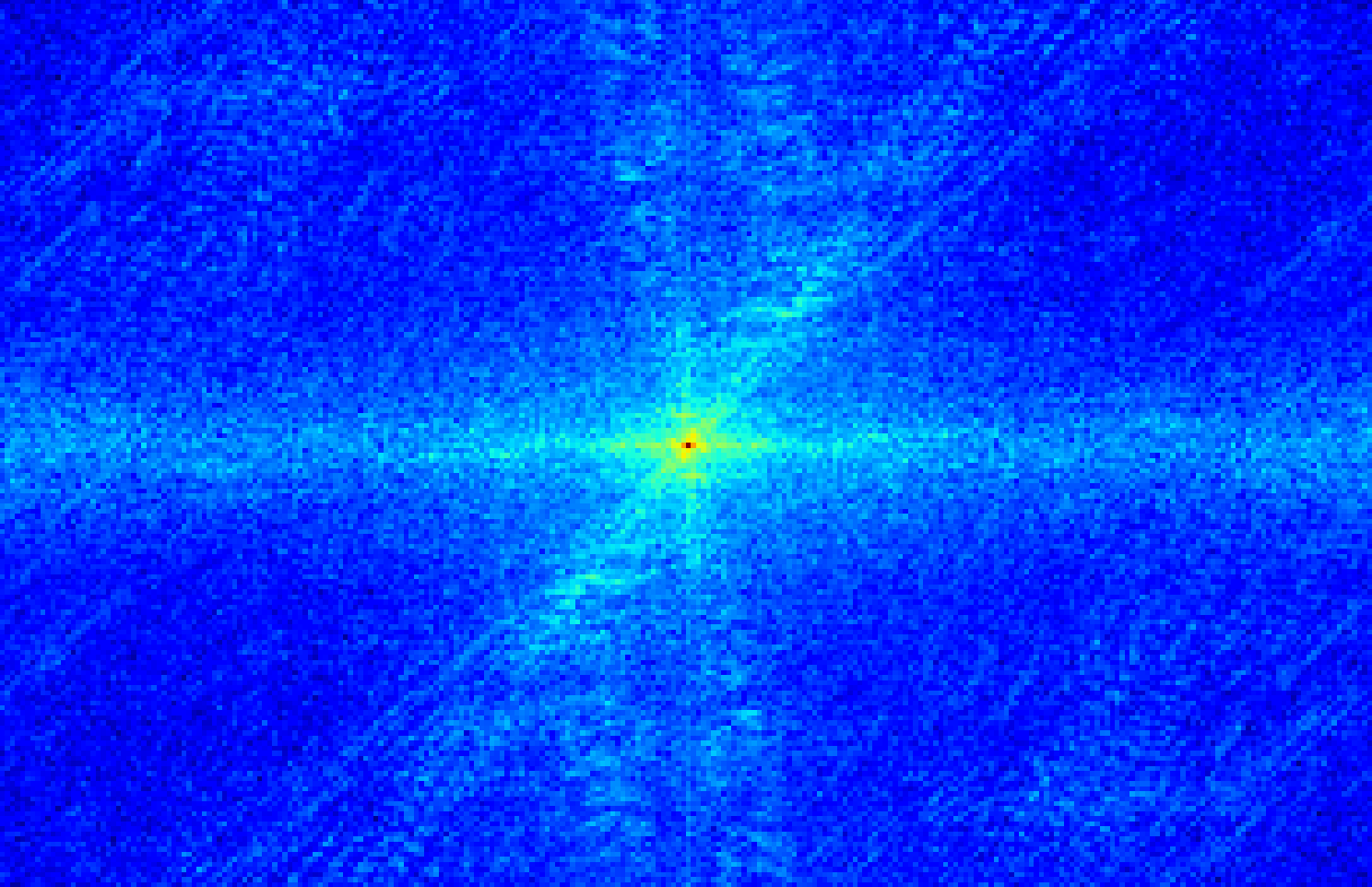}
\label{fig:filterfloat}
}
\\
\subfloat[Quantization]{
\AddImg{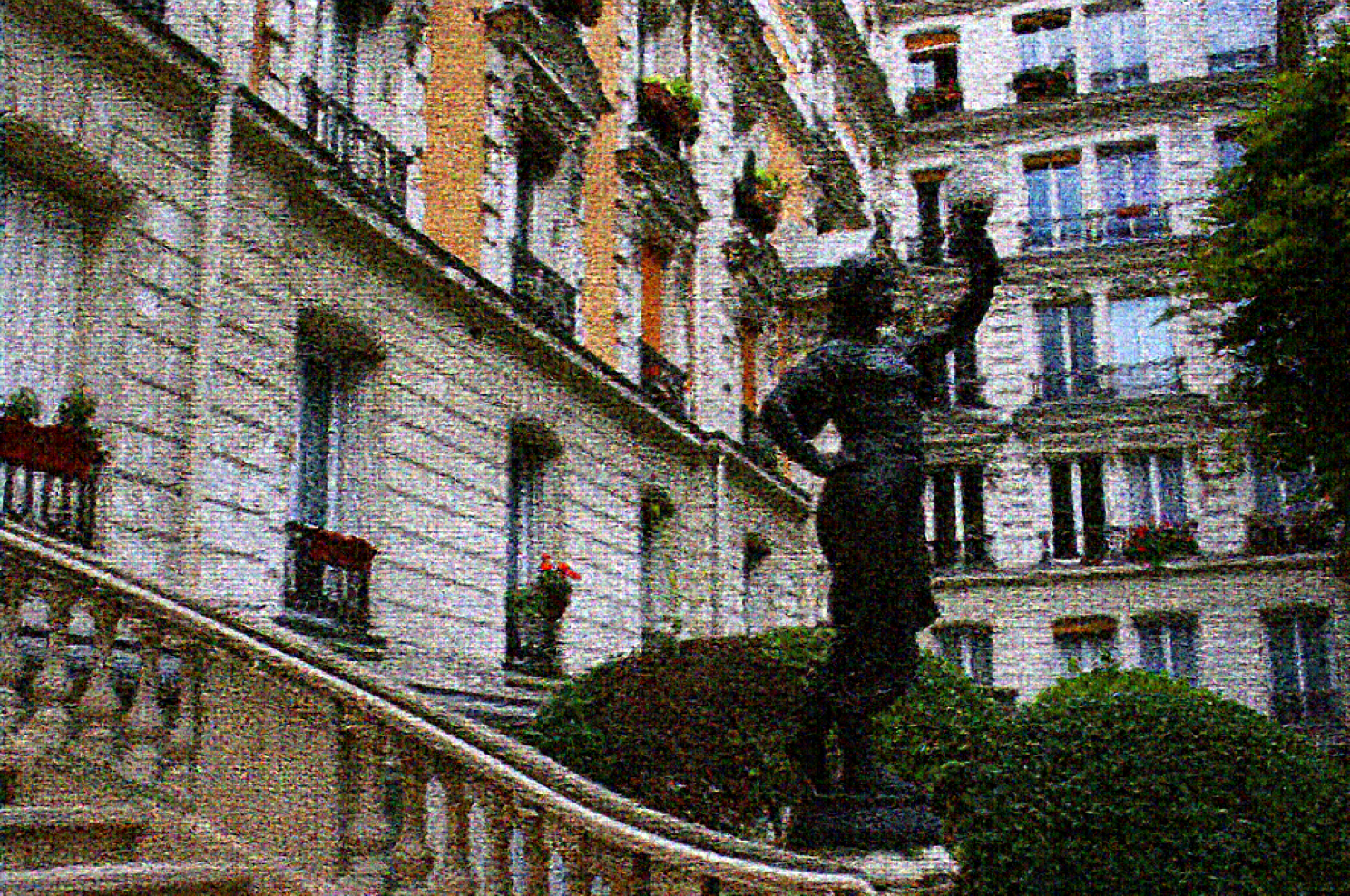}
\AddImg{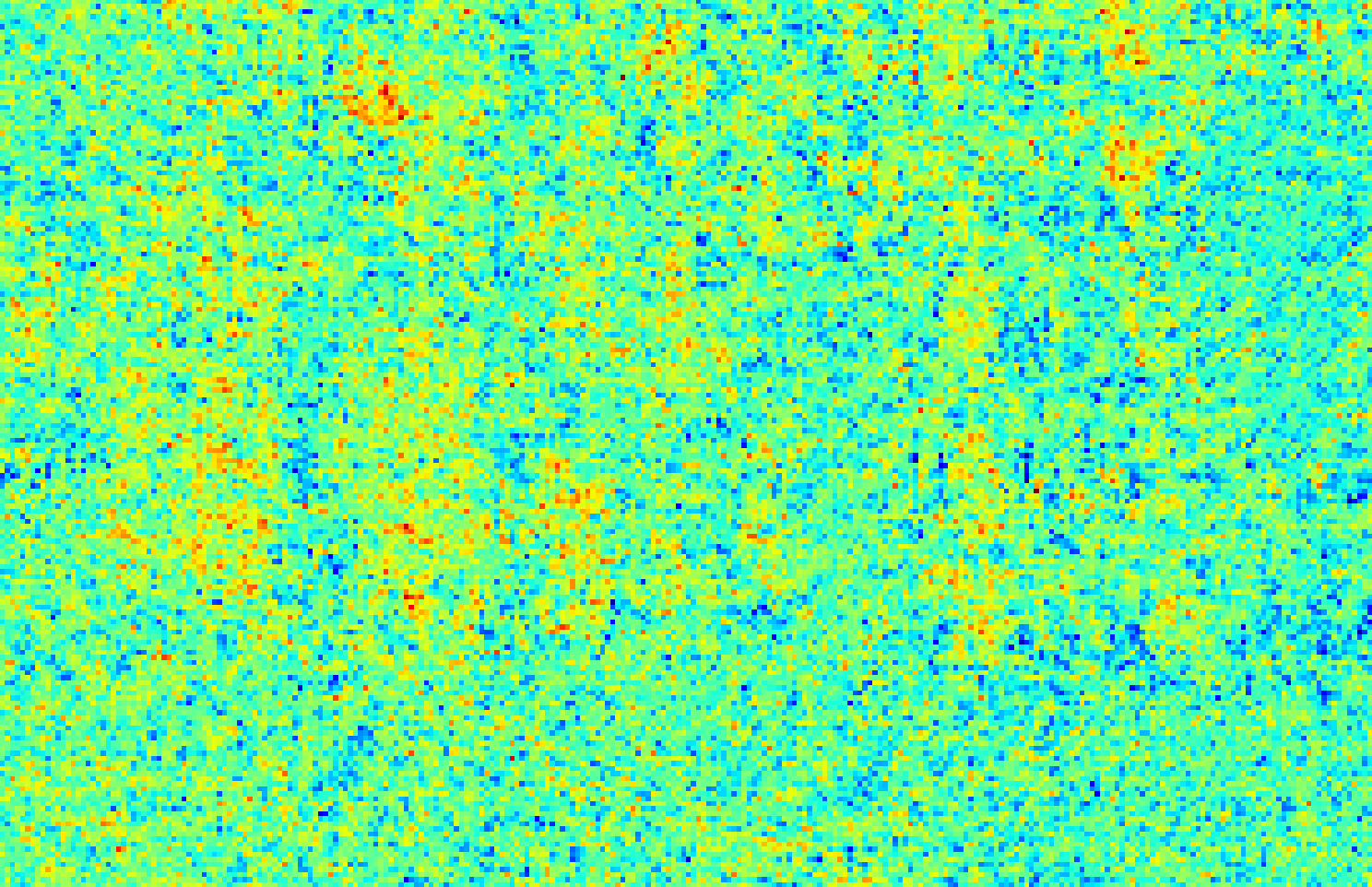}
\AddImg{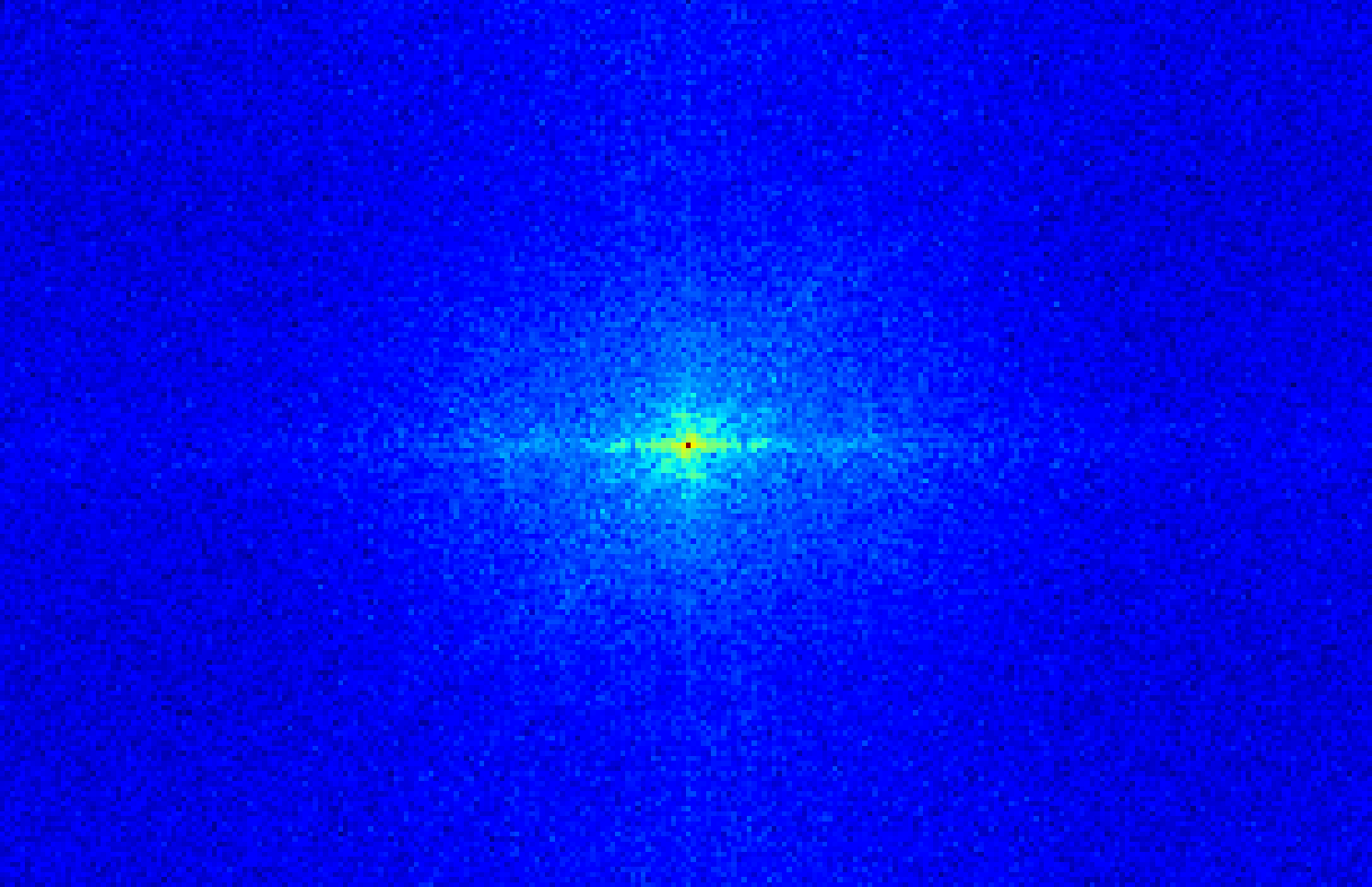}
\label{fig:filtered_quant}
}
\caption{{Demonstrating the impact of quantization on frequency and image representation. Here, we showcase the output of HFERB with 4-bit quantization to highlight the loss of high-frequency components.}}
\vspace{-5mm}
\label{fig:hf_loss_ori}
\end{figure}

{From the frequency perspective, we exhibit the outputs of HFERB in both the frequency domain and feature space, as well as the original image, under 4-bit quantization, as shown in Fig.~\ref{fig:hf_loss_ori}. The first row displays the results with full precision, while the second row presents the outcomes after quantization. The figure demonstrates that quantization leads to a significant loss of high-frequency components (illustrated in the third column) and fails to generate meaningful features (shown in the second column), compared to the original image (first column). Considering HFERB is tailored to introduce high-frequency priors into CRAFT, the reduction of high-frequency content crucially affects this block's function, resulting in compromised quality of the super-resolved output.}

\begin{algorithm}[!t]
\caption{{Frequency-guided Criteria Measuring Processing (FCMP)}}
\label{alg1}
\KwIn{Quantization bit $b$, minimal boundary $l$, maximal boundary $u$, {measuring type $t$, FFT operation $\mathcal{F}(\cdot)$}, and feature maps $X$ with $C$ channels.}
\KwOut{Criteria score $\gamma$.}
\BlankLine
{$scale \gets \frac{u - l}{2^b - 1}$\;
$zero\ point \gets clip(\left\lfloor -\frac{l}{scale} \right\rceil; 0, 2^b - 1)$\;
$X^{int} \gets clip(\left\lfloor \frac{X}{scale} \right\rceil + zero\ point; 0, 2^b - 1)$\;
$\hat{X} \gets (X^{int} - zero\ point) \cdot scale$\;
\eIf{$t = \text{FGO}$}{
$\gamma \gets \frac{1}{C} \sum\limits_{i = 1}^{C} \left|( \left| \mathcal{F}(X_i) \right| - \left| \mathcal{F}(\hat{X}_i) \right| )\right|$\;
}{
$\gamma \gets \frac{1}{C} \sum\limits_{i = 1}^C \left| X_i - \hat{X}_i \right|$\;
}}
\Return{$\gamma$}
\end{algorithm}

\subsection{{Adaptive Dual Clipping with Frequency-guided Optimization}}
To address the asymmetry in feature distributions, we propose a quantization method that combines an adaptive dual clipping (ADC) strategy with frequency-guided optimization (FGO). For each layer requiring quantization, we begin by plotting its histogram and then gradually adjust the range to determine a more accurate quantization range. This method differs from previous approaches that adjust both the lower and upper boundaries simultaneously or based on statistical calculations. Our strategy relies on comparing the features before and after quantization to refine the quantization range.

{Our strategy is based on the frequency-guided criteria measuring process (FCMP), as described in Algorithm~\ref{alg1}. It plays a crucial role in assessing the fidelity loss caused by quantization, mainly by calculating the mean absolute error (MAE) between the pre-quantized and post-quantized features. Importantly, our strategy stops adjusting when an increase in MAE indicates a possible deterioration in the quality of image reconstruction. Furthermore, given the noticeable reduction in high-frequency components attributable to quantization, as shown in Fig.~\ref{fig:hf_loss_ori}, which impairs HFERB, we have implemented a frequency domain MAE constraint not only on HFERB but also on the model's input and output layers. This strategy is designed to more effectively regulate the quantization process, preventing excessively narrow quantization ranges.}

\begin{algorithm}[!t]
\caption{Adaptive Dual Clipping Processing}
\label{alg2}
\KwIn{Quantization layer $F$, calibration data $d$, Quantization bit $b$, {measuring type $t$} and the number of bins $2^b$.}
\KwOut{$\{l, u\}$.}
\BlankLine
Initialize $\{l, u\}$ with the minimum and maximum values of $F$\;
$H(x_1),\ldots,H(x_{2^b})\leftarrow Histogram(F(d))$\;
$l_{_{\text{best}}} \leftarrow \min(F(d))$, $u_{_{\text{best}}} \leftarrow \max(F(d))$\;
$\Delta \leftarrow (u_{_{\text{best}}}-l_{_{\text{best}}})/{2^b}$\;
$\gamma_{_{\text{best}}} \leftarrow \infty$\;
\Repeat{{increasing $\gamma_{_{\text{best}}}$ detected}}{
{$\gamma_{_{\text{new}}}^{_l} = FCMP(b, l_{_{\text{best}}}+\Delta, u_{_{\text{best}}}, t, F(d))$}\;
{$\gamma_{_{\text{new}}}^{_r} = FCMP(b, l_{_{\text{best}}}, u_{_{\text{best}}}-\Delta, t, F(d))$}\;
\eIf{$\gamma_{_{\text{new}}}^{_l} < \gamma_{_{\text{new}}}^{_r}$}{
    $\gamma_{_{\text{new}}} = \gamma_{_{\text{new}}}^{_l}$\;
    $l_{_{\text{new}}} = l_{_{\text{best}}}+\Delta$\;
}{
    $\gamma_{_{\text{new}}} = \gamma_{_{\text{new}}}^{_r}$\;
    $u_{_{\text{new}}} = u_{_{\text{best}}}-\Delta$\;
}
$\gamma_{_{\text{best}}} \leftarrow \gamma_{_{\text{new}}}$\;
$l_{_{\text{best}}} \leftarrow l_{_{\text{new}}}$, $u_{_{\text{best}}} \leftarrow u_{_{\text{new}}}$\;
}
$l=l_{_{\text{best}}}$\;
$u=u_{_{\text{best}}}$\;
\Return{$\{l, u\}$}
\end{algorithm}

Expanding on this concept, Algorithm~\ref{alg2} details our ADC process, wherein the values from FCMP play a pivotal role in incrementally adjusting the clipping boundaries. Beginning with the initial clipping boundaries at the feature extremes for each layer of the full-precision CRAFT, we use the calibration dataset to iteratively refine these boundaries.
\subsection{Boundary Refinement Process}
{Addressing the high dynamic range of features presents a significant challenge, particularly under conditions of extremely low-bit (4-bit) quantization. We further introduce a boundary refinement (BR) strategy that enhances the preliminary boundaries set by the ADC process.}

\begin{algorithm}[!t]
\caption{Boundary Refinement Process}
\label{alg3}
\KwIn{Coarse boundary values $\{l_{coarse}, u_{coarse}\}^K$ of $K$ layers, calibration data $d$, learning rate $\eta$, full-precision model $F$.}
\KwOut{Refined boundary values $\{l_{refined}, u_{refined}\}^K$.}
\BlankLine
Initialize $\{l_{refined}, u_{refined}\}^K$ with $\{l_{coarse}, u_{coarse}\}^K$\;
{According to the input data $d$, perform a full-precision and quantized model forward pass to obtain $\hat{X}$ and $X$}\;
{Calculate reconstruction loss $\mathcal{L}_{rec.}$ using Eq.~\eqref{eq:eq14}\;}
{Perform backpropagation and update $\{l_{refined}, u_{refined}\}^K$ with learning rate $\eta$}\;
\Return{$\{l_{refined}, u_{refined}\}^K$}
\end{algorithm}
{The BR strategy is detailed in Algorithm~\ref{alg3}. We start with the coarse boundary values obtained from the ADC stage as our initial inputs. These initial boundaries are then transformed into learnable parameters, allowing the model to fine-tune them through training. The quantizer's discrete nature results in direct differentiation leading to zero gradients, which hinders backpropagation. To address this issue, we employ straight-through estimation (STE)~\cite{bengio2013estimating}, which enables us to approximate gradient calculation for these parameters.}

Our strategy diverges from the method outlined by~\cite{tu2023toward}, which applies intermediate feature regularization and iterative updates the boundaries of weights and activations. Instead, we focus our regularization efforts solely on the output, synchronizing the updates to the boundaries of weights and activations. This difference is critically important as it removes the necessity to store intermediate feature representations, greatly reducing the memory usage during training and contributing to more stable training.

During the boundary refinement process, we use the MAE as the reconstruction loss for optimization, which is defined as:
\begin{equation}
\mathcal{L}_{rec.} = \frac{1}{B}\sum_{i=1}^{B}\left\lVert X_{i}-\hat{X}_{i}\right\rVert_{1},
\label{eq:eq14}
\end{equation}
where \(B\) denotes the batch size, \(X_i\) represents the output from the full-precision model, and \(\hat{X}_i\) indicate the output from the quantization model.

\begin{algorithm}[!t]
\caption{{Frequency-guided PTQ Process}}
\label{alg4}
\KwIn{{Full-precision SR model $F$ of $K$ layers, calibration dataset $\mathcal{D}_{cal}$, quantization bit $b$, measuring type $\{t\}^K$ of $K$ layers, number of bins $2^b$, total epochs $E$, learning rate $\eta$.}}
\KwOut{{Refined boundary values $\{l_{refined}, u_{refined}\}^K$.}}

\BlankLine
\tcc{{Use the ADC (Algorithm \ref{alg2}) with the FCMP (Algorithm \ref{alg1}))}}
{Initialize $\{l_{coarse}, u_{coarse}\}^K$ with the minimum and maximum values of $K$ layers\;}
\For{$d \in \mathcal{D}_{cal}$}{
    \For{$k = 1$ to $K$}{
        $l_{_{\text{best}}}, u_{_{\text{best}}} = ADC(F^k, d, b, t^k, 2^b)$\;
        $l_{coarse}^k = \beta \cdot l_{coarse}^k + (1-\beta) \cdot l_{_{\text{best}}}$\;
        $u_{coarse}^k = \beta \cdot u_{coarse}^k + (1-\beta) \cdot u_{_{\text{best}}}$\;
    }
}
\BlankLine
\tcc{{BR Process (Algorithm \ref{alg3})}}
\For{$i = 1$ to $E$}{
    \For{$d \in \mathcal{D}_{cal}$}{
        $\{l_{coarse}, u_{coarse}\}^K = BR(\{l_{coarse},u_{coarse}\}^K,d,\eta,F)$\;
    }
}
$\{l_{refined}, u_{refined}\}^K = \{l_{coarse}, u_{coarse}\}^K$\;
\Return{$\{l_{refined}, u_{refined}\}^K$}
\end{algorithm}

\subsection{Overview of Frequency-guided PTQ Strategy}
{Based on the two strategies described above, we conduct the overall process of our frequency-guided PTQ process, as illustrated in Algorithm~\ref{alg4}. First, we set the layers of HFERB and the input and output layers to $t = \text{FGO}$, and then initialize $l$ and $u$ for each layer with their corresponding minimal and maximal values. After that, we perform ADC with the calibration set $\mathcal{D}_{cal}$ to refine the boundaries. This refinement process involves assessing two potential new boundaries at each iteration, utilizing a moving average method to establish more stable boundary values. Specifically, for those layers where $t$ is equal to FGO, we apply frequency constraints instead of feature domain constraints. We then obtain the coarse boundaries for each layer, $\{l_{coarse},u_{coarse}\}$. Following this, we perform BR to further refine the coarse boundaries using the same calibration set $\mathcal{D}_{cal}$ over $E$ epochs. Notably, with improved initial boundaries, a limited number of epochs (\eg, 10 epochs) is sufficient for fine-tuning these boundaries to achieve desired outcomes. Finally, we obtain the refined boundaries for each layer, $\{l_{refined},u_{refined}\}$.}
\input{tables/craft_cmp_sota}
\section{Experimental Results}
\label{exp}
\subsection{Data and Metrics}
Our CRAFT model is trained on the DIV2K dataset~\cite{agustsson2017ntire}, which is composed of 800 high-resolution images. Meanwhile, five benchmarks are used for evaluation, including Set5~\cite{bevilacqua2012low}, Set14~\cite{zeyde2010single}, BSD100~\cite{martin2001database}, Urban100~\cite{huang2015single}, and Manga109~\cite{matsui2017sketch} with three magnification factors: $\times2$, $\times3$, and $\times4$. The quality of the reconstructed images is evaluated using PSNR, and SSIM~\cite{wang2004image}. The complexity of the model is indicated by its parameters.

{For quantization, 100 low-resolution patches of $120\times120$ pixels are randomly selected from the DIV2K dataset to form a calibration dataset. This calibration dataset is consistently used across both stages of quantization. Furthermore, the high-resolution counterparts are not included to simulate scenarios often encountered in practice, where ground truths may not be available. The effectiveness of the proposed quantization strategies is further validated across seven benchmark datasets: Set5~\cite{bevilacqua2012low}, Set14~\cite{zeyde2010single}, BSD100~\cite{martin2001database}, Urban100~\cite{huang2015single}, Manga109~\cite{matsui2017sketch}, DIV2K~\cite{agustsson2017ntire}, and LSDIR~\cite{li2023lsdir}, at a $\times4$ magnification factor, using the PSNR and SSIM metrics for a comprehensive assessment.}

\subsection{Implementation Details}
Following the general setting, we use bicubic to obtain the corresponding LR images from the original HR images. During training, we randomly crop the images into $64 \times 64$ patches, and the total training iterations are 500K. Meanwhile, data augmentation is performed, such as random horizontal flipping and $90^\circ$ rotation. The Adam optimizer with $\beta_1=0.9$ and $\beta_2=0.999$ is adopted to minimize the $\mathcal{L}_1$ Loss. The batch size is set to 64, the initial learning rate is set to $2\times 10^{-4}$ and reduced by half at the milestone [250K, 400K, 450K, 475K]. In addition, the model is trained on 4 NVIDIA 3090 GPUs using the PyTorch toolbox. 

During the quantization, we adjusted the batch size to 2 and proceeded with training for 10 epochs. The learning rate for 8-bit quantization was set at $2\times 10^{-4}$ and increased to $2\times 10^{-3}$ for 6-bit and 4-bit quantizations to boost their performance. The smoothing parameter $\beta$ was set to 0.9. {Throughout this phase, simulated quantization was applied to all convolutional and linear layers, as well as matrix multiplication operations.} Additionally, we ensured that the input and output precision remained at 8 bits. This quantization stage was performed on a single NVIDIA 3090 GPU, utilizing the PyTorch framework.

\begin{table}[!t]
  \caption{Effect of HFERB, SRWAB, and HFB on SISR. The results ($\times 4$) are obtained from the Manga109 dataset.}
  \vspace{-3mm}
  \centering
  \footnotesize
  \setlength{\tabcolsep}{1.5mm}
  \begin{tabular}{cccccc}
      \toprule
      Model & HFERB & SRWAB & HFB & Concat & PSNR \\
      \midrule
      CRAFT$_{conv}$ & $\checkmark$ & & $\checkmark$ &  & 30.79\\
      CRAFT$_{tranformer}$ &  & $\checkmark$ & $\checkmark$ &  & 31.12 \\
      CRAFT$_{concat}$ & $\checkmark$ & $\checkmark$ &  & $\checkmark$ & 30.92\\
      CRAFT & $\checkmark$ & $\checkmark$ & $\checkmark$ &  & 31.18\\
      \bottomrule
  \end{tabular}
  \vspace{-3mm}
  \label{tab:ablationmodules}
\end{table}

Within the CRAFT architecture, we integrate 4 RCRFGs and include 2 CRFBs in each RCRFG, with each CRFB comprising 1 HFERB and 2 SRWABs to balance efficiency and performance. We set the feature channels and the attention head to 48 and 6, and set the MLP expansion ratio to 2. Based on the work by Zamir~\etal~\cite{Zamir2022}, we select an IMLP expansion ratio of 2.66. To accommodate diverse receptive fields, two distinct rectangle window sizes are used in our model, specifically $[sh, sw] = [4, 16]$ and $[16, 4]$.
\subsection{Comparison with State-of-the-Art Methods}
We benchmark our proposed CRAFT model against various leading SISR methods, including EDSR~\cite{Lim2017}, CARN~\cite{Ahn2018}, IMDN~\cite{Hui2019}, LatticeNet~\cite{luo2020latticenet}, LAPAR~\cite{Li2020}, SwinIR~\cite{liang2021swinir}, HPUN~\cite{sun2022hybrid}, ESRT~\cite{lu2022transformer}, LBNet~\cite{gao2022lightweight}, and ELAN~\cite{zhang2022efficient}.

\textbf{Quantitative Results.}
As shown in Table \ref{tab:resultsLightweight}, CRAFT outperforms conventional CNN-based methods such as EDSR, achieving gains of 0.85dB, 0.84dB, and 0.83dB at $\times2$, $\times3$, and $\times4$ magnification factors, respectively, on the Manga109 dataset, while requiring 46\%, 52\%, and 50\% fewer parameters. Against channel attention methods like CARN, CRAFT demonstrates improvements of 1.03dB, 0.79dB, and 0.71dB for the same magnification factors, accompanied by a 54\%, 53\%, and 52\% parameter reduction. Compared to transformer-based models~\cite{lu2022transformer, liang2021swinir, zhang2022efficient}, CRAFT offers enhancements of 0.34dB, 0.31dB, and 0.29dB at the $\times3$ factor, with a similar parameter count.

\begin{figure*}[!t]
    \centering
    \footnotesize
    \begin{overpic}[scale=.35]{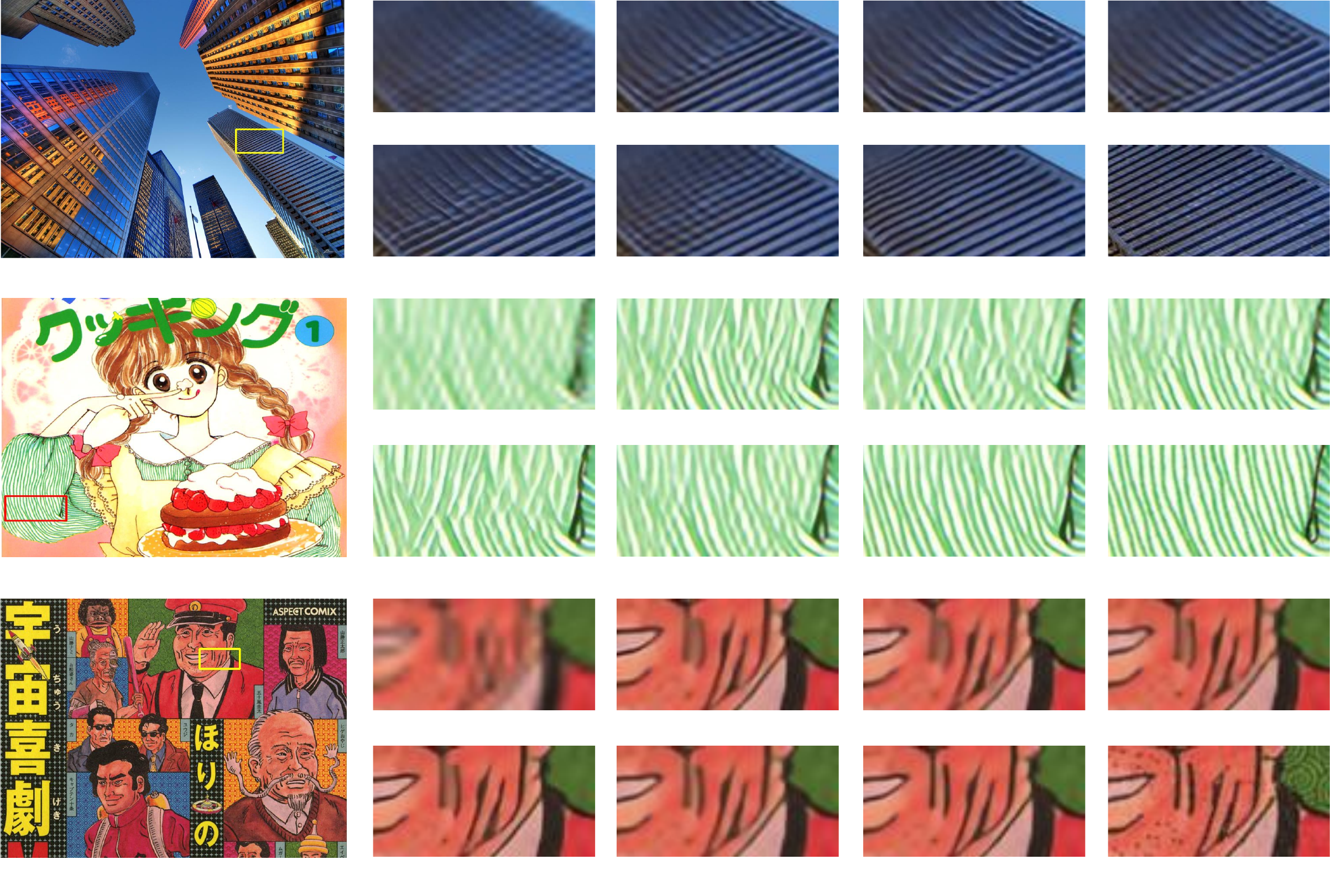}
    \put(9,46.3){Img012 ($\times4$)}
    \put(6.5,23.7){YumeiroCook ($\times4$)}
    \put(5.5,1){UchuKigekiM774 ($\times4$)}

    \put(33.7,57.3){Bicubic}
    \put(51.3,57.3){EDSR~\cite{Lim2017}}
    \put(69,57.3){CARN~\cite{Ahn2018}}
    \put(88,57.3){LBNet~\cite{gao2022lightweight}}
    \put(32.8,46.3){SwinIR~\cite{liang2021swinir}}
    \put(51.3,46.3){ESRT~\cite{lu2022transformer}}
    \put(67,46.3){CRAFT (Ours)}
    \put(90,46.3){HR}
    
    \put(33.7,34.8){Bicubic}
    \put(51.3,34.8){EDSR~\cite{Lim2017}}
    \put(69,34.8){CARN~\cite{Ahn2018}}
    \put(88,34.8){LBNet~\cite{gao2022lightweight}}
    \put(32.8,23.7){SwinIR~\cite{liang2021swinir}}
    \put(51.3,23.7){ESRT~\cite{lu2022transformer}}
    \put(67,23.7){CRAFT (Ours)}
    \put(90,23.7){HR}
    
    \put(33.7,12.3){Bicubic}
    \put(51.3,12.3){EDSR~\cite{Lim2017}}
    \put(69,12.3){CARN~\cite{Ahn2018}}
    \put(88,12.3){LBNet~\cite{gao2022lightweight}}
    \put(32.8,1){SwinIR~\cite{liang2021swinir}}
    \put(51.3,1){ESRT~\cite{lu2022transformer}}
    \put(67,1){CRAFT (Ours)}
    \put(90,1){HR}
    \end{overpic}
    \vspace{-3mm}
    \caption{Qualitative comparison with SOTA methods. CRAFT achieves better restoration quality in both line directions and details.}
    \label{fig:visulizationlightweight}
\end{figure*}

\begin{figure*}[!t]
    \centering
    \footnotesize
    \begin{overpic}[scale=.27]{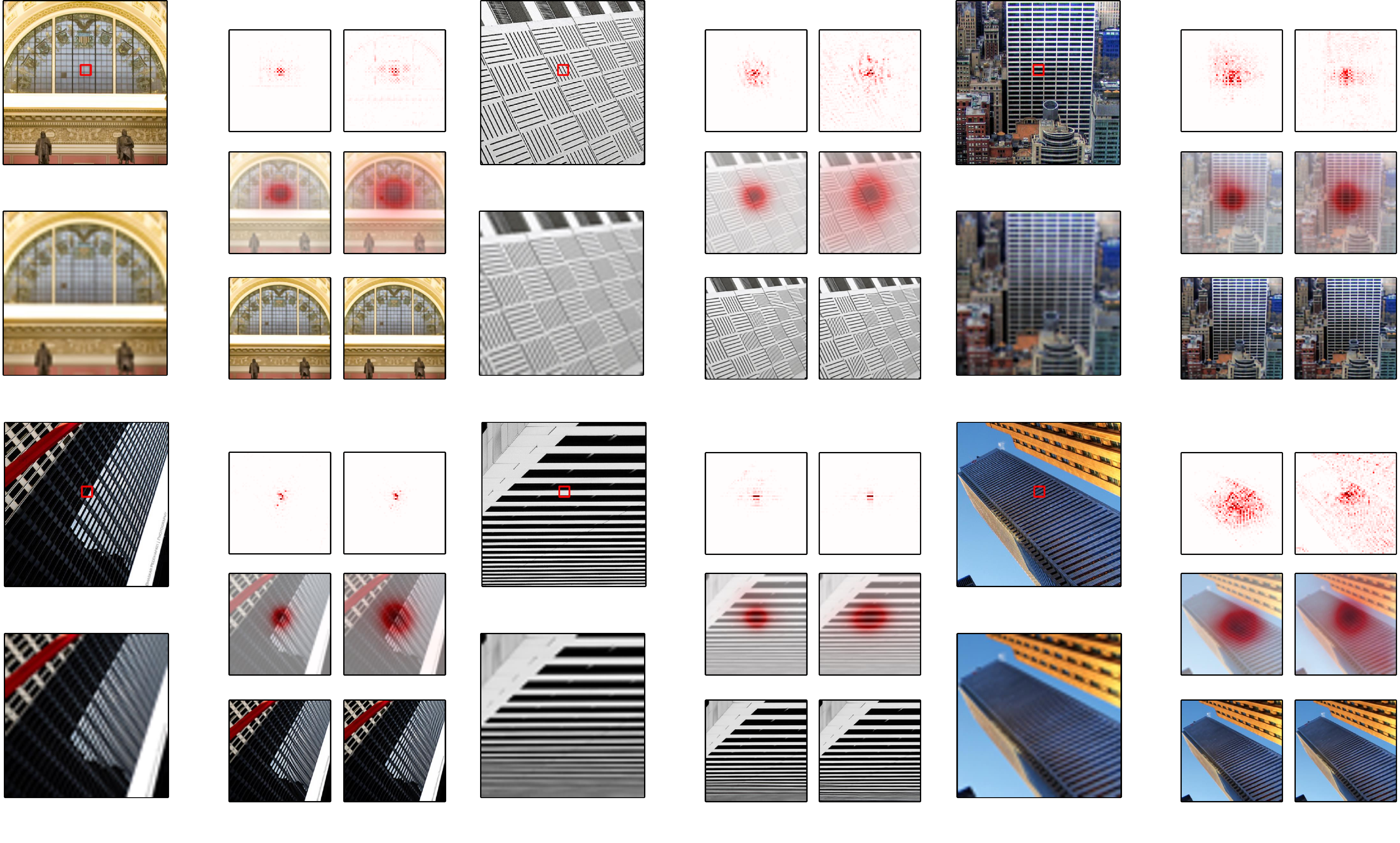}
    \put(5,46.3){HR}
    \put(5,31){LR}
    \put(17.5,58.5){SwinIR}
    \put(25.6,58.5){CRAFT} 
    \put(16.4,31){DI: 10.78} 
    \put(24.8,31){DI: 30.34} 
    \put(12.6,52.2){\rotatebox{90}{\shortstack{LAM\\Results}}} 
    \put(12.6,41.4){\rotatebox{90}{\shortstack{Contribution\\Area}}} 
    \put(12.6,34.2){\rotatebox{90}{\shortstack{SR\\Results}}} 

    \put(39.2,46.3){HR}
    \put(39.2,31){LR}
    \put(51.3,58.5){SwinIR}
    \put(59.5,58.5){CRAFT} 
    \put(51,31){DI: 9.37} 
    \put(58.9,31){DI: 30.94} 
    \put(46.6,52.2){\rotatebox{90}{\shortstack{LAM\\Results}}} 
    \put(46.6,41.4){\rotatebox{90}{\shortstack{Contribution\\Area}}} 
    \put(46.6,34.2){\rotatebox{90}{\shortstack{SR\\Results}}} 

    \put(73,46.3){HR}
    \put(73,31){LR}
    \put(85.4,58.5){SwinIR}
    \put(93.5,58.5){CRAFT} 
    \put(84.6,31){DI: 13.03} 
    \put(92.9,31){DI: 30.87} 
    \put(80.6,52.2){\rotatebox{90}{\shortstack{LAM\\Results}}} 
    \put(80.6,41.4){\rotatebox{90}{\shortstack{Contribution\\Area}}} 
    \put(80.6,34.2){\rotatebox{90}{\shortstack{SR\\Results}}} 

    \put(5,16.3){HR}
    \put(5,0.5){LR}
    \put(17.5,28.3){SwinIR}
    \put(25.6,28.3){CRAFT} 
    \put(16.4,0.5){DI: 7.36} 
    \put(24.8,0.5){DI: 18.89} 
    \put(12.6,22){\rotatebox{90}{\shortstack{LAM\\Results}}} 
    \put(12.6,11.2){\rotatebox{90}{\shortstack{Contribution\\Area}}} 
    \put(12.6,4.2){\rotatebox{90}{\shortstack{SR\\Results}}} 

    \put(39.2,16.3){HR}
    \put(39.2,0.5){LR}
    \put(51.3,28.3){SwinIR}
    \put(59.5,28.3){CRAFT} 
    \put(51,0.5){DI: 9.30} 
    \put(58.9,0.5){DI: 17.98} 
    \put(46.6,22){\rotatebox{90}{\shortstack{LAM\\Results}}} 
    \put(46.6,11.2){\rotatebox{90}{\shortstack{Contribution\\Area}}} 
    \put(46.6,4.2){\rotatebox{90}{\shortstack{SR\\Results}}} 

    \put(73,16.3){HR}
    \put(73,0.5){LR}
    \put(85.4,28.3){SwinIR}
    \put(93.5,28.3){CRAFT} 
    \put(84.6,0.5){DI: 12.88} 
    \put(92.9,0.5){DI: 32.17} 
    \put(80.6,22){\rotatebox{90}{\shortstack{LAM\\Results}}} 
    \put(80.6,11.2){\rotatebox{90}{\shortstack{Contribution\\Area}}} 
    \put(80.6,4.2){\rotatebox{90}{\shortstack{SR\\Results}}} 
    \end{overpic}
    \caption{Comparison of the LAM results of SwinIR~\cite{liang2021swinir} and CRAFT. LAM indicates the correlation between the significance of each pixel in LR and the SR of the patch that is outlined with the red box. CRAFT uses a broader range of information to obtain better performance. DI quantifies the LAM results, CRAFT has a higher DI score, indicating its ability to capture more contextual information.} 
    \vspace{-4mm}
    \label{fig:ablation_lam}
\end{figure*}

\textbf{Qualitative Results.}
We present a visual comparison ($\times4$) in Fig.~\ref{fig:visulizationlightweight} and analyze the results. Our proposed CRAFT model integrates the strengths of both CNN and transformer structures, leading to accurate line direction recovery while preserving image details. To further investigate the performance, we compare the local attribution map (LAM) \cite{gu2021interpreting} between CRAFT and SwinIR, as shown in Fig.~\ref{fig:ablation_lam}. LAM indicates the correlation between the significance of each pixel in LR and the SR of the patch that is outlined with the red box. By leveraging a broader range of information, our model achieves improved results. Furthermore, we examine the diffusion index (DI), which signifies the range of pixels involved. A larger DI indicates a wider scope of attention. Compared to SwinIR, our model exhibits a higher DI, implying that it can capture more contextual information. These results demonstrate the effectiveness of the proposed CRAFT method.

\subsection{Effect of CRAFT Components}

\begin{figure*}[!t]
    \centering
    \footnotesize
    \includegraphics[width=0.85\linewidth]{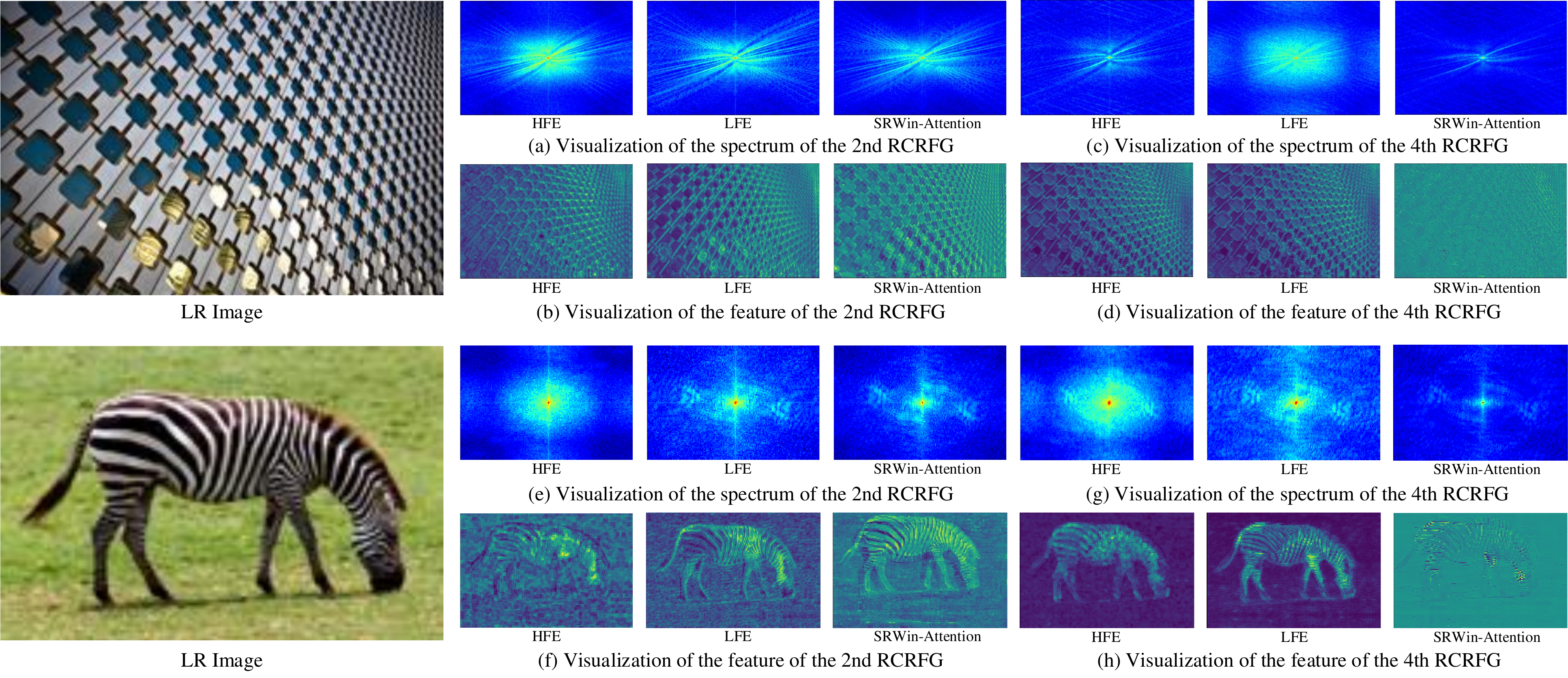}
    \vspace{-3mm}
    \caption{Visualization of HFERB and SRWAB. The LFE indicates the local feature extraction branch in HFERB, the HFE means the high-frequency enhancement branch in HFERB, and the SRWin-Attention represents the self-attention part in SRWAB.}
    \vspace{-5mm}
    \label{fig:visulizationFrequency}
\end{figure*}

\subsubsection{Effect of HFERB and SRWAB}
We conducted several experiments to demonstrate the effectiveness of HFERB and SRWAB, as presented in Table~\ref{tab:ablationmodules}. Specifically, we removed SRWAB and HFERB separately to assess their contributions. We observed that relying solely on local or global information, as denoted by CRAFT$_{conv}$ and CRAFT$_{transformer}$, respectively, resulted in inadequate representation learning, leading to lower performance. Furthermore, we found that SRWAB provides the most significant performance improvement, demonstrating the benefits of the long-range dependencies learned by the transformer. In addition, high-frequency priors from the CNN are also helpful in improving performance, cross-refining the learned features and further improving performance. 

\subsubsection{Frequency Analysis of HFERB and SRWAB}
We visualized the features extracted from two blocks in different RCRFGs and plotted the Fourier spectrum to observe what each block learns. The results, shown in Fig.~\ref{fig:visulizationFrequency}, indicate that HFERB focuses more on high-frequency information, while SRWAB extracts more global information. Specifically, the top row of each image indicates the Fourier spectrum of each block, and the bottom row indicates the feature maps of each block. The figure shows that SRWAB has a weaker response and focuses more on the low-frequency parts, which correspond to flat regions, while HFERB shows a stronger response and focuses more on intricate parts of features, such as edges and corners. The feature maps on the bottom row also support this conclusion. HFERB captures more details such as window edges and cornices, while SRWAB pays more attention to flat areas such as windows and walls.

\begin{table}[!t]
  \caption{Effect of high-frequency prior. The results ($\times 4$) are obtained from the Manga109 dataset.}
  \vspace{-3mm}
  \centering
  \footnotesize
  \setlength{\tabcolsep}{2mm}
  \begin{tabular}{cccccc}
      \toprule
      Model & Regular & Swap & Cascade & PSNR & SSIM \\
      \midrule
      CRAFT$_{swap}$ &  & $\checkmark$ & & 30.67 & 0.9113\\
      CRAFT$_{cascade}$ &  & & $\checkmark$ & 30.88 & 0.9141\\
      CRAFT & $\checkmark$ & & & 31.18 & 0.9168\\
      \bottomrule
  \end{tabular}
  \vspace{-4mm}
  \label{tab:ablationcrs}
\end{table}

\subsubsection{Effect of HFB}
To evaluate the effectiveness of HFB, we conducted an experiment where we modified the fusion method to a concatenation formulation. This involved concatenating the outputs of HFERB and SRWAB and replacing the HFB with a $3 \times 3$ convolutional layer to obtain the final output. The results are presented in Table \ref{tab:ablationmodules}, where CRAFT$_{concat}$ denotes the modified version. The result shows that our proposed method outperforms the concatenation structure by 0.26dB, demonstrating the effectiveness of our HFB. This improvement can be attributed to SRWAB and HFERB focusing on disparate frequency information. Stacking features directly impedes the ability of the network to learn the relationship between high-frequency and low-frequency components. Conversely, the inter-attention mechanism presents a viable solution for integrating features with different distributions.

\begin{figure}[!t]
\centering
\footnotesize
\begin{overpic}[scale=.27]{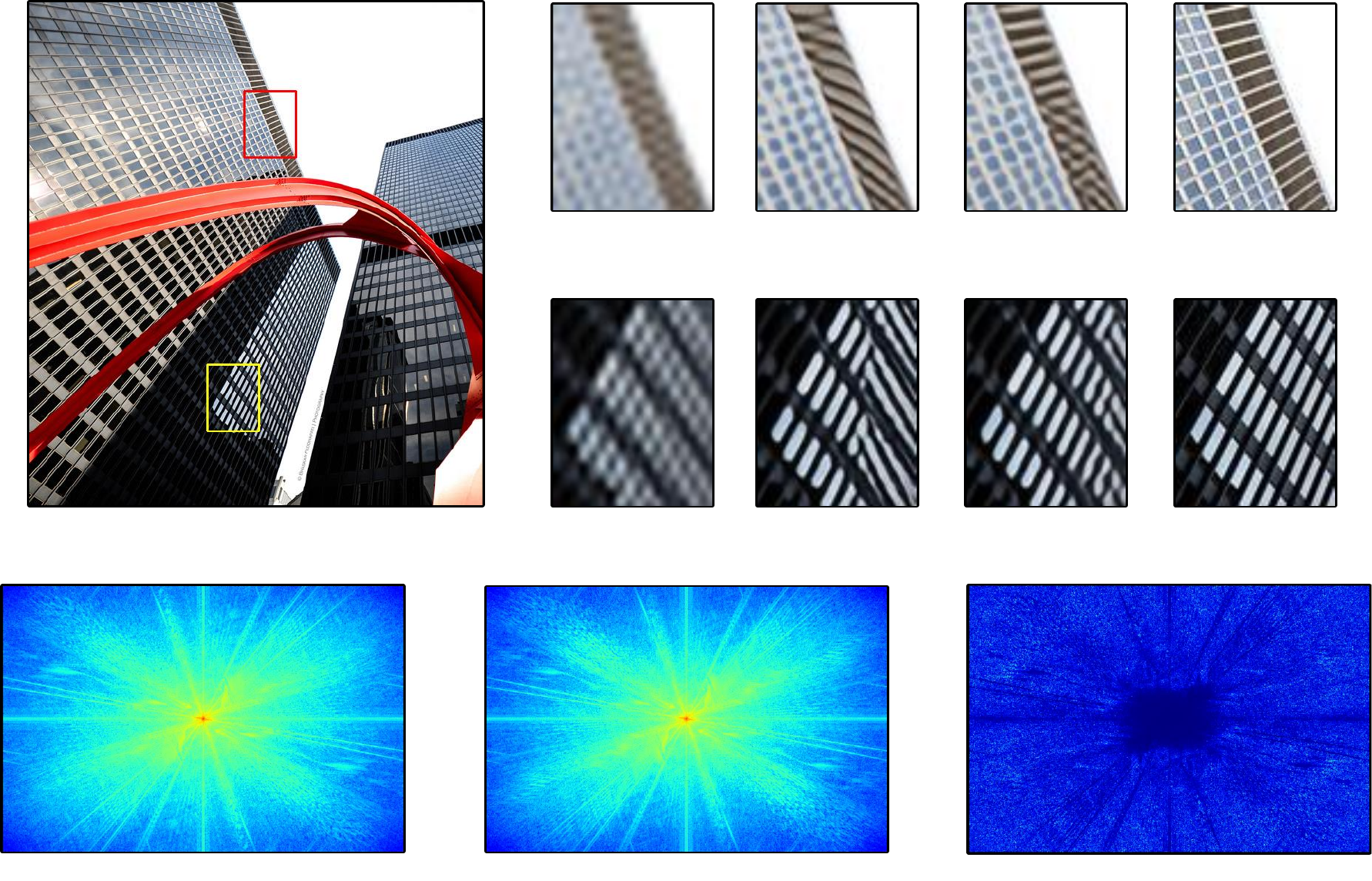}
\put(11.3,24){\scriptsize{Img062 ($\times4$)}}
\put(41.4,45.5){\scriptsize{Bicubic}}
\put(57.2,45.5){\scriptsize{\emph{w/o} H}}
\put(73,45.5){\scriptsize{\emph{w/} H}}
\put(89,45.5){\scriptsize{HR}}
\put(41.4,24){\scriptsize{Bicubic}}
\put(57.2,24){\scriptsize{\emph{w/o} H}}
\put(73,24){\scriptsize{\emph{w/} H}}
\put(89.5,24){\scriptsize{HR}}
\put(9.5,-1.5){\scriptsize{$\Phi(\emph{w/}\hspace{3pt}H)$}}
\put(43,-1.5){\scriptsize{$\Phi(\emph{w/o}\hspace{3pt}H)$}}
\put(68,-1.5){\scriptsize {$|\Phi(\emph{w/}\hspace{3pt}H)\!-\!\Phi(\emph{w/o}\hspace{3pt}H)|$}}
\end{overpic}
\caption{Qualitative comparison of the model with and without HFERB. Models with and without HFERB are denoted as $\emph{w/}\hspace{3pt}H$ and $\emph{w/o}\hspace{3pt}H$, respectively. The symbols $\Phi(\emph{w/o}\hspace{3pt}H)$ and $\Phi(\emph{w/}\hspace{3pt}H)$ represent the spectra of the models without and with HFERB, respectively. The evaluation is conducted using a magnification factor of $\times4$.}
\vspace{-4mm}
\label{fig:visulizationmore3}
\end{figure}

\subsubsection{Effect of the High-Frequency Prior}
We conducted several experiments to investigate the effectiveness of high-frequency prior. Firstly, we swapped the input of $\boldsymbol{Q}$ and $\boldsymbol{K}$, $\boldsymbol{V}$ in HFB and treated the output of SRWAB as $\boldsymbol{Q}$ and the output of HFERB as $\boldsymbol{K}$, $\boldsymbol{V}$ to verify whether global features are dominant in restoration and high-frequency features only serve as a prior for refining the global representation. As shown in Table~\ref{tab:ablationcrs}, compared to the original design, swapping the input leads to a significant drop in performance, with a 0.51dB decrease in PSNR. Furthermore, we also performed an experiment to formulate the model as a cascade structure to verify the effectiveness of the design introducing high-frequency priors. As shown in Table~\ref{tab:ablationcrs}, the CRAFT$_{cascade}$ structure resulted in a performance drop, with a 0.3dB decrease in PSNR compared to CRAFT. These results demonstrate the effectiveness of high-frequency priors in the CRAFT model. Additionally, we visualized the frequency domain representations in Fig.~\ref{fig:visulizationmore3} to demonstrate that integrating a high-frequency prior enhances detail sharpness. The high-frequency improvements were quantified using Fourier transforms, where $\Phi(\emph{w/}\hspace{3pt}H)$ and $\Phi(\emph{w/o}\hspace{3pt}H)$ denote the spectra of models with and without the high-frequency prior, respectively. Specifically, the spectrum of both models was defined as:
\begin{equation}
\begin{split}
   &\Phi(\emph{w/}\hspace{3pt}H) = FFT(\emph{w/}\hspace{3pt}H),\\
   &\Phi(\emph{w/o}\hspace{3pt}H) = FFT(\emph{w/o}\hspace{3pt}H).
   \label{eq:suppeq1}
\end{split}
 \end{equation}
From there, we derived the residual spectrum map between $\Phi(\emph{w/}\hspace{3pt}H)$ and $\Phi(\emph{w/o}\hspace{3pt}H)$, formulated as:
\begin{equation}
R(\emph{w/}\hspace{3pt}H,\emph{w/o}\hspace{3pt}H) = |\Phi(\emph{w/}\hspace{3pt}H)-\Phi(\emph{w/o}\hspace{3pt}H)|,
\end{equation}
where $R(\cdot)$ represents the residual spectrum map. This map shows a stronger high-frequency response when a high-frequency prior is included, indicating improved restoration of high-frequency components. Concurrently, visual assessment reveals that the introduction of high-frequency priors leads to the restoration of more accurate details, underscoring the significance of HFERB.

\begin{table}[!t]
  \centering
  \footnotesize
  \setlength{\tabcolsep}{2.8mm}
  \caption{Complexity analysis compared to SwinIR with an magnification factor of $\times4$. The FLOPs and inference time are measured under the setting of generating $512\times 512$ image.}
  \vspace{-3mm}
  \label{tab:abcomplex}
  \begin{tabular}{ccccc} \toprule
      Model & \makecell[c]{\#Params.\\(K)} & \makecell[c]{\#FLOPs\\(G)} & \makecell[c]{\#GPU Mem.\\(M)} & \makecell[c]{Ave. Time\\(ms)}\\
      \midrule
      SwinIR & 897 & 32.2 & 141.2 & 79.11 \\
      CRAFT & 753 & 26.0 & 79.5 & 56.40 \\
      \bottomrule
  \end{tabular}
  \vspace{-3mm}
\end{table}

\begin{table}[!t]
  \centering
  \footnotesize
  \caption{Complexity analysis of each block with an magnification factor of $\times4$. The FLOPs is measured under the setting of generating $512\times 512$ image.}
  \vspace{-3mm}
  \label{tab:abcomponets}
  \resizebox{\linewidth}{!}{
  \begin{tabular}{ccccc} \toprule
     Model &\makecell[c]{ CRAFT \\w/o HFERB} & \makecell[c]{CRAFT\\ w/o SRWAB} & \makecell[c]{CRAFT\\ w/o HFB} & CRAFT
     \\ \midrule
     \#Params. (K) & 688 & 441 & 503 & 753 \\
     \#FLOPs (G) & 23.8 & 14.2 & 20.0 & 26.0
     \\ \bottomrule
  \end{tabular}}
  \vspace{-2mm}
\end{table}

\subsubsection{{Exploring CRAFT‘s Capability in Restoring High-Frequency Components}}
{Given our findings regarding the limited ability of transformers to reconstruct high-frequency information, as discussed in Sec.~\ref{sec:motivation}, we conducted an experiment using log-amplitude spectral analysis to assess CRAFT’s capabilities. Specifically, we performed this experiment on the image \textit{img092} from the Urban100 dataset, applying a magnification factor of 4. Bicubic interpolation and SwinIR~\cite{liang2021swinir} were set as baselines to evaluate CRAFT's performance. The experimental results are depicted in Fig.~\ref{fig:quant_log_power_frequency}. The x-axis represents increasing frequency values, with higher values indicating higher frequencies. For clearer comparison, we have magnified a localized region of the curves, shown in the top-right inset of the figure. As observed in the figure, the HR image retains the entire range of high-frequency components, while the bicubic interpolation loses nearly all high-frequency details. SwinIR~\cite{liang2021swinir} is able to restore some high-frequency information, but CRAFT demonstrates superior performance, recovering a greater extent of high-frequency components.}
\input{tables/ab_capability_recon_hf}
\begin{table}[!t]
  \caption{{Runtime tested on an NVIDIA GeForce RTX 3090 GPU with a $128\times 128$ input image. The PSNR results are obtained from the Manga109 dataset.}}
  \vspace{-3mm}
  \centering
  \footnotesize
  \setlength{\tabcolsep}{2mm}
  \begin{tabular}{ccccccc}
      \toprule
      Scale & Model & Mask & w/o Mask  & PSNR & Time (ms) \\
      \midrule
      \multirow{2}{*}{${\times 2}$} & CRAFT$_{mask}$ & $\checkmark$ & & 31.18 & 87.44\\
      &CRAFT &  & $\checkmark$ & 31.18 & 55.36\\
      \midrule
      \multirow{2}{*}{${\times 3}$} & CRAFT$_{mask}$ & $\checkmark$ & & 31.18 & 86.67\\
      &CRAFT &  & $\checkmark$ & 31.18 & 51.50\\
      \midrule
      \multirow{2}{*}{${\times 4}$} & CRAFT$_{mask}$ & $\checkmark$ & & 31.18 & 93.08\\
      &CRAFT &  & $\checkmark$ & 31.18 & 56.40\\
      \bottomrule
  \end{tabular}
  \vspace{-4mm}
  \label{tab:ablationlmask}
\end{table}
\input{tables/ptq_cmp_sota}

\input{tables/ptq_vsual_cmp}

\subsubsection{Analysis of CRAFT's Complexity}
We compared CRAFT with SwinIR in terms of complexity using an input size of $128\times128$, as shown in Table~\ref{tab:abcomplex}. The analysis considered parameters, FLOPs, GPU memory consumption, and average inference time. Compared to SwinIR, CRAFT has fewer parameters and FLOPs, and requires less memory consumption and inference time. Furthermore, we analyzed the complexity of our CRAFT framework and summarized the findings in Table~\ref{tab:abcomponets}. We observed that SRWAB contributes approximately 46\% of the total complexity, while HFERB involves fewer convolution operations, resulting in reduced FLOPs. Moreover, the HFB module's channel-wise attention effectively reduces the computational burden. {In addition, we investigated the impact of employing a mask mechanism in rectangle attention on the model's efficiency, as illustrated in Table~\ref{tab:ablationlmask}. Our findings suggest that omitting the mask does not compromise performance but significantly boosts model inference speed. The results indicate that removing the mask leads to a reduction in inference time by approximately 37\%, 41\%, and 39\% for magnification factors of 2, 3, and 4, respectively.}

\subsection{Comparison with quantization strategies}
This section presents a comparative analysis of our quantization strategies against other PTQ methods, including MinMax~\cite{jacob2018quantization}, Percentile~\cite{li2019fully}, and PTQ4SR~\cite{tu2023toward}. The evaluations are performed across various quantization bit widths, with magnification factor of $\times 4$.

{\textbf{Quantitative Results.} We compare the performance of our CRAFT after applying various PTQ strategies, showcased in Table~\ref{tab:resultsQuantization}. Our evaluation spans across different quantization bit-widths (8-bit, 6-bit, and 4-bit) for both model weights and activations. Generally, the results show a significant drop in performance as the quantization bit level decreases, particularly at 4-bit quantization. Significantly, methods like MinMax~\cite{jacob2018quantization} and Percentile~\cite{li2019fully}, which lack boundary refinement, show considerable degradation in performance compared to the full-precision model. Despite this challenge, our PTQ strategies exhibit superior performance across all quantization levels. Specifically, when compared to MinMax~\cite{jacob2018quantization} and Percentile~\cite{li2019fully}, our approach results in significant PSNR improvements of up to 8.32dB and 8.70dB at 4-bit quantization. Even against method that use a boundary refinement strategy like PTQ4SR~\cite{tu2023toward}, our approach achieves performance gains of up to 2.22dB at 4-bit quantization.}

{\textbf{Qualitative Results.} We present the visual results obtained by applying different quantization methods to CRAFT in Fig.~\ref{fig:visulizationQuantization}. We selected two samples from the Urban100 and Set14 datasets and applied 4-bit quantization with a $\times4$ magnification factor. From the images, it's evident that both MinMax~\cite{jacob2018quantization} and Percentile~\cite{li2019fully} fail to accurately recreate edges and context, introducing more noise during quantization. While PTQ4SR~\cite{tu2023toward} improves SR quality to some extent through further refinement of boundaries, it still exhibits a slightly distorted appearance with less detailed clarity compared to full-precision results. In contrast, our proposed PTQ strategies yield clearer and more accurate results. Notably, even when compared to full-precision models, the outputs from our quantized model are not only acceptable but also convey meaningful visual information. This highlights the practical utility of our approach.}
\input{tables/ptq_ab_strategy}
\input{tables/ptq_ab_dual_clip}
\input{tables/ptq_ab_quantization_recon_hf}
\input{tables/ptq_ab_general_quant}
\vspace{-3mm}
\subsection{{Analysis of PTQ Strategies}}
\subsubsection{{Effect of Frequency-guided Optimization}}
{We conducted a detailed comparison with PTQ4SR~\cite{tu2023toward}, as shown in Table~\ref{tab:ablationquantstrategy}, to highlight the benefits of our frequency-guided optimization (FGO). We used PTQ4SR~\cite{tu2023toward} as a baseline for comparing our method's effectiveness in dual clipping and boundary refinement. The strategies of DBDC and PaC, as described in PTQ4SR~\cite{tu2023toward}, were evaluated against our ADC and BR strategies. Initially, we compared our ADC strategy without FGO to PTQ4SR~\cite{tu2023toward}'s DBDC and found that our ADC achieved better starting values, outperforming DBDC by up to 3.54dB. Following this, we assessed the performance of DBDC against our ADC with the FGO strategy included, clearly demonstrating that incorporating FGO into our strategy further improved performance, exceeding DBDC by up to 3.70dB.} {Furthermore, we visualize the boundary values of the two different clipping strategies across 4-bit, 6-bit, and 8-bit quantizations with magnification factor of 4 in Fig.~\ref{fig:boudary_comparison}. We choose the output of CRFB, HFERB, SRWAB, and HFB as the probe. From the visual results, it's evident that our ADC demonstrates a more constrained clipping range, providing a more precise initial value for subsequent training stages. This narrower range implies that a more accurate representation of the small range of values is sufficient to represent the full-precision counterparts.}

{Regarding the boundary refinement stage, we initially compare our training strategy with PTQ4SR~\cite{tu2023toward}. From the table, it's evident that with the same initial boundary values, our proposed strategy achieves better performance across all test datasets. Furthermore, the introduction of FGO improves boundary refinement performance by providing better initial boundary values. Additionally, we visualize the HFERB output to further evaluate the superiority of our FGO method over CRAFT, as depicted in Fig.~\ref{fig:ablation_QUANTIZATION}. From the figure, it's apparent that implementing FGO preserves more high-frequency information in the quantized model, making it more similar to the full-precision model. This observation suggests that combining quantization strategies with FGO leads to improved performance.}

\subsubsection{{General Transformer-based SR PTQ Strategy}}
{In this section, we extend our frequency-guided PTQ strategies to a general quantization approach for transformer-based SISR methods. Specifically, CRAFT utilizes high-frequency priors to enhance its performance, a characteristic derived from its reliance on specialized modules designed to exploit high-frequency information. In contrast, standard transformer models for SISR do not integrate such modules, making high-frequency optimization unnecessary for them. Therefore, we choose to exclude the FGO from our generalized PTQ strategy for these models. Instead, we enforce MAE constraints within the feature domain across all layers during the ADC stage, while retaining all other previously established strategies. Subsequently, we apply the generalized PTQ strategy to transformer-based SR models such as SwinIR~\cite{liang2021swinir} and CRAFT. We conduct comprehensive performance analyses at quantization levels of 8-bit, 6-bit, and 4-bit using seven commonly used test datasets, as outlined in Table~\ref{tab:ablationgeneralQuantization}. From the table, it is evident that all models maintain a similar level of performance, demonstrating the effectiveness of our general PTQ strategies in adapting to transformer-based SR methods. Particularly noteworthy is the comparison between CRAFT and SwinIR under the same quantization conditions, where CRAFT consistently exhibits superior performance across nearly all datasets, thus affirming its effectiveness and robustness.}
\begin{figure}[!t]
    \centering
    \includegraphics[width=0.98\linewidth]{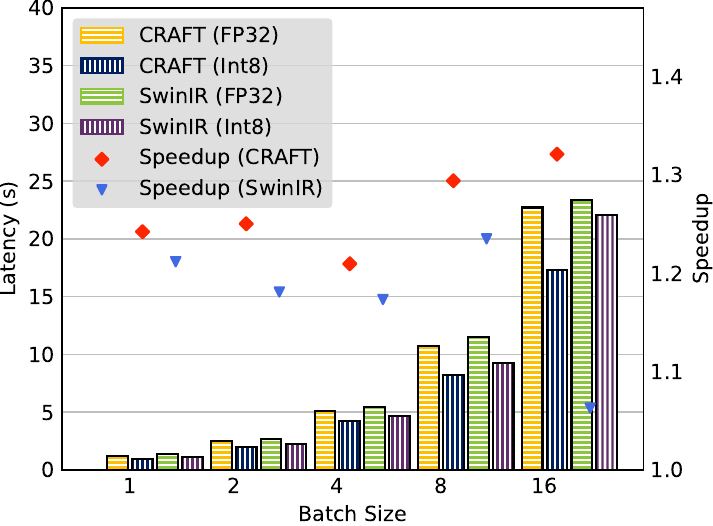}
    \vspace{-3mm}
    \caption{{Hardware performance of CRAFT and SwinIR on AMD Ryzen 5 5600U CPU. The red diamond and blue triangle represent the speedup ratio of CRAFT and SwinIR. The quantized model consistently accelerates processing across various batch sizes.}}
    \vspace{-3mm}
    \label{fig:quant-cmp}
\end{figure}
\begin{table}[!t]
  \centering
  \footnotesize
  \setlength{\tabcolsep}{1.1mm}
  \caption{{Comparison of PTQ with different training strategies. ``Regular'' refers to full-precision training, while ``AMP'' stands for Automatic Mixed Precision training. ``Prec.'' denotes the precision value.}}
  \vspace{-3mm}
  \begin{tabular}{cccccc} \toprule
      Model & Prec. & Model Size & Set14 & BSD100 & Manga109\\
        \midrule
      Regular & FP32 & 3.37M & 28.85/.7872 & 27.72/.7418  & 31.18/.9168 \\
        \midrule
       \multirow{2}{*}{AMP} & FP16 & 1.81M & 28.83/.7871 & 27.71/.7416 & 31.13/.9160 \\
        & FP8 & 1.51M & 28.15/.7708 & 27.29/.7266 & 29.16/.8888 \\
        \midrule
      PTQ & INT8 & 1.13M & 28.80/.7850 & 27.68/.7397 & 31.06/.9141\\
      \bottomrule
  \end{tabular}
  \vspace{-3mm}
  \label{tab:amp_cmp}
\end{table}
\subsubsection{{Analysis of Quantized Acceleration}}
\label{quantization_analysis}
{The primary goal of this evaluation was to assess the effects of quantization on resource-constrained platforms, such as portable PC CPUs, which more closely resemble the typical deployment scenarios for quantized models. While 8-bit quantization is also applicable to GPUs, our initial focus was to investigate the performance of quantized models on hardware with more limited resources.} Due to limitations in current hardware, which offers limited support for fully quantized models, especially for lower-bit operations (\eg, 6-bit or 4-bit), we restricted our quantization to 8-bit for this study. The tests were performed on input images sized $128 \times 128$, applying a $\times 4$ super-resolution. The acceleration results are presented in Fig.~\ref{fig:quant-cmp}, where we compared our CRAFT model with SwinIR~\cite{liang2021swinir} to evaluate the effectiveness of our quantization technique. The quantized INT8 model consistently exhibited lower latency compared to the full-precision FP32 model, achieving a speed improvement of up to 24\%. However, the actual speedup fell short of theoretical expectations. We believe this is due to several factors, including limitations in INT8 batch matrix multiplication and the presence of non-quantizable operations, such as softmax and LayerNorm, which reduce the overall acceleration potential of INT8 quantization during inference. To further highlight the advantages of PTQ, we performed additional comparisons with models trained using the Automatic Mixed Precision (AMP) strategy, which is widely used in model training~\cite{son2022real, liu2021swin, micikevicius2017mixed}. As shown in Table~\ref{tab:amp_cmp}, the models were trained with AMP in both FP16 and FP8\footnote{{FP8 training was conducted using NVIDIA’s Transformer Engine, which only supports converting the linear layers to FP8, while other layers remain in FP16.}}. While AMP-trained FP16 models demonstrated only a slight performance decrease compared to full-precision models, FP8 showed significant degradation. In contrast, PTQ achieved notably smaller model sizes while maintaining competitive performance. These results emphasize PTQ’s superior efficiency, offering clear advantages in terms of both performance and model size over floating-point models.

\section{{Conclusion and Future Works}}
\label{sec:conclusion}
{This study systematically explores how frequency information affects the performance of CNN and transformer structures in single image super-resolution (SISR). It reveals that while transformer structures excel at capturing low-frequency information, they struggle with reconstructing high-frequency details compared to CNNs. To tackle this limitation, we introduce a novel transformer variant called cross-refinement adaptive feature modulation transformer (CRAFT). CRAFT comprises three main components: the high-frequency enhancement residual block (HFERB) for extracting high-frequency features, the shift rectangle window attention block (SRWAB) for capturing global representations, and the hybrid fusion block (HFB) which refines global representation by treating HFERB output as a high-frequency prior and SRWAB output as key and value for inter-attention. Furthermore, to simplify the inherent complexity of transformers, we propose a frequency-guided post-training quantization (PTQ) strategy. It involves adaptive dual clipping and boundary refinement, enhance CRAFT's efficiency. Additionally, we extend our PTQ strategies to serve as general quantization methods for transformer-based SISR approaches. Experimental results demonstrate CRAFT's superiority over existing methods in both full-precision and quantization scenarios. Further experiments validate the effectiveness of our PTQ approach and the versatility of its extended version. Future work include but are not limited extending the application of our frequency-guided PTQ strategies to a wide range of models and tasks, aiming to conduct thorough efficacy evaluations.}
\ifCLASSOPTIONcaptionsoff
  \newpage
\fi

\bibliographystyle{IEEEtran}
\bibliography{IEEEabrv,./reference.bib}
\end{document}

%% file: tables/craft_cmp_sota.tex
\begin{table*}[!t]
  \caption{Performance comparison of different SISR models on five benchmarks. ``\#Params'' represents the total number of network parameters. The best and second best results for each setting are \textbf{highlighted} and \underline{underlined}, respectively.}
  \vspace{-3mm}
  \centering
  \footnotesize
  \setlength{\tabcolsep}{3.2mm}
  \begin{tabular}{cccccccc}
      \toprule
      Scale & Model & \#Params & \makecell[c]{Set5\\(PSNR/SSIM)} & \makecell[c]{Set14\\(PSNR/SSIM)} & \makecell[c]{BSD100\\(PSNR/SSIM)} & \makecell[c]{Urban100\\(PSNR/SSIM)} & \makecell[c]{Manga109\\(PSNR/SSIM)}\\
      \midrule
      \multirow{10}{*}{${\times 2}$} & EDSR-baseline \cite{Lim2017} & 1370K & 37.99/0.9604 & 33.57/0.9175 & 32.16/0.8994 & 31.98/0.9272 & 38.54/0.9769\\
      & CARN \cite{Ahn2018} & 1592K & 37.76/0.9590 & 33.52/0.9166 & 32.09/0.8978 & 31.92/0.9256 & 38.36/0.9765\\
      & IMDN \cite{Hui2019} & 694K & 38.00/0.9605 & 33.63/0.9177 & 32.19/0.8996 & 32.17/0.9283 & 38.88/0.9774\\
      & LatticeNet \cite{luo2020latticenet} & 756K & 38.06/0.9607 & 33.70/0.9187 & 32.20/0.8999 & 32.25/0.9288 & -/-\\
      & LAPAR-A \cite{Li2020} & 548k & 38.01/0.9605 & 33.62/0.9183 & 32.19/0.8999 & 32.10/0.9283 & 38.67/0.9772\\
      & HPUN-L \cite{sun2022hybrid} & 714K & 38.09/0.9608 & 33.79/0.9198 & 32.25/0.9006 & 32.37/0.9307 & 39.07/0.9779\\
      \cdashline{2-8}[1pt/1pt]
      & SwinIR-light \cite{liang2021swinir} & 878K & 38.14/\underline{0.9611} & 33.86/0.9206 & \underline{32.31}/\underline{0.9012} & \underline{32.76}/\underline{0.9340} & \underline{39.12}/\underline{0.9783}\\
      & ESRT \cite{lu2022transformer} & 777K & 38.03/0.9600 & 33.75/0.9184 & 32.25/0.9001 & 32.58/0.9318 & \underline{39.12}/0.9774\\
      & ELAN-light \cite{zhang2022efficient} & 582K & \underline{38.17}/\underline{0.9611} & \textbf{33.94}/\underline{0.9207} & 32.30/\underline{0.9012} & \underline{32.76}/\underline{0.9340} & 39.11/0.9782\\
      & CRAFT (Ours) & 737K & \textbf{38.23}/\textbf{0.9615} & \underline{33.92}/\textbf{0.9211} & \textbf{32.33}/\textbf{0.9016} & \textbf{32.86}/\textbf{0.9343} & \textbf{39.39}/\textbf{0.9786}\\
      \midrule
      \multirow{11}{*}{${\times 3}$} & EDSR-baseline \cite{Lim2017} & 1555K & 34.37/0.9270 & 30.28/0.8417 & 29.09/0.8052 & 28.15/0.8527 & 33.45/0.9439\\
      & CARN \cite{Ahn2018} & 1592K & 34.29/0.9255 & 30.29/0.8407 & 29.06/0.8034 & 28.06/0.8493 & 33.50/0.9440\\
      & IMDN \cite{Hui2019} & 703K & 34.36/0.9270 & 30.32/0.8417 & 29.09/0.8046 & 28.17/0.8519 & 33.61/0.9445\\
      & LatticeNet \cite{luo2020latticenet} & 765K & 34.53/0.9281 & 30.39/0.8424 & 29.15/0.8059 & 28.33/0.8538 & -/-\\
      & LAPAR-A \cite{Li2020} & 544k & 34.36/0.9267 & 30.34/0.8421 & 29.11/0.8054 & 28.15/0.8523 & 33.51/0.9441\\
      & HPUN-L \cite{sun2022hybrid} & 723K & 34.56/0.9281 & 30.45/0.8445 & 29.18/0.8072 & 28.37/0.8572 & 33.90/0.9463\\
      \cdashline{2-8}[1pt/1pt]
      & SwinIR-light \cite{liang2021swinir} & 886K & \underline{34.62}/\underline{0.9289} & 30.54/\underline{0.8463} & 29.20/\underline{0.8082} & 28.66/\underline{0.8624} & 33.98/\underline{0.9478}\\
      & LBNet \cite{gao2022lightweight} & 736K & 34.47/0.277 & 30.38/0.8417 & 29.13/0.8061 & 28.42/0.8559 & 33.82/0.9406\\
      & ESRT \cite{lu2022transformer} & 770K & 34.42/0.9268 & 30.43/0.8433 & 29.15/0.8063 & 28.46/0.8574 & 33.95/0.9455\\
      & ELAN-light \cite{zhang2022efficient} & 590K & 34.61/0.9288 & \underline{30.55}/\underline{0.8463} & \underline{29.21}/0.8081 & \underline{28.69}/\underline{0.8624} & \underline{34.00}/\underline{0.9478}\\
      & CRAFT (Ours) & 744K & \textbf{34.71}/\textbf{0.9295} & \textbf{30.61}/\textbf{0.8469} & \textbf{29.24}/\textbf{0.8093} & \textbf{28.77}/\textbf{0.8635} & \textbf{34.29}/\textbf{0.9491}\\
      \midrule
      \multirow{11}{*}{${\times 4}$} & EDSR-baseline \cite{Lim2017} & 1518K & 32.09/0.8938 & 28.58/0.7813 & 27.57/0.7357 & 26.04/0.7849 & 30.35/0.9067\\
      & CARN \cite{Ahn2018} & 1592K & 32.13/0.8937 & 28.60/0.7806 & 27.58/0.7349 & 26.07/0.7837 & 30.47/0.9084\\
      & IMDN \cite{Hui2019} & 715K & 32.21/0.8948 & 28.58/0.7811 & 27.56/0.7353 & 26.04/0.7838 & 30.45/0.9075\\
      & LatticeNet \cite{luo2020latticenet} & 777K & 32.18/0.8943 & 28.61/0.7812 & 27.57/0.7355 & 26.14/0.7844 & -/-\\
      & LAPAR-A \cite{Li2020} & 659k & 32.15/0.8944 & 28.61/0.7818 & 27.61/0.7366 & 26.14/0.7871 & 30.42/0.9074\\
      & HPUN-L \cite{sun2022hybrid} & 734K & 32.31/0.8962 & 28.73/0.7842 & 27.66/0.7386 & 26.27/0.7918 & 30.77/0.9109\\
      \cdashline{2-8}[1pt/1pt]
      & SwinIR-light \cite{liang2021swinir} & 897K &\underline{32.44}/\underline{0.8976} & 28.77/0.7858 & \underline{27.69}/\underline{0.7406} & 26.47/0.7980 & \underline{30.92}/\underline{0.9151}\\
      & LBNet \cite{gao2022lightweight} & 742K & 32.29/0.8960 & 28.68/0.7832 & 27.62/0.7382 & 26.27/0.7906 & 30.76/0.9111\\
      & ESRT \cite{lu2022transformer} & 751K & 32.19/0.8947 & 28.69/0.7833 & \underline{27.69}/0.7379 & 26.39/0.7962 & 30.75/0.9100\\
      & ELAN-light \cite{zhang2022efficient} & 601K & 32.43/0.8975 & \underline{28.78}/\underline{0.7858} & \underline{27.69}/\underline{0.7406} & \underline{26.54}/\underline{0.7982} & \underline{30.92}/0.9150\\
      & CRAFT (Ours) & 753K &\textbf{32.52}/\textbf{0.8989} & \textbf{28.85}/\textbf{0.7872} & \textbf{27.72}/\textbf{0.7418} & \textbf{26.56}/\textbf{0.7995} & \textbf{31.18}/\textbf{0.9168}\\
      \bottomrule
  \end{tabular}
  \vspace{-3mm}
  \label{tab:resultsLightweight}
\end{table*}

%% file: tables/ab_capability_recon_hf.tex
\begin{figure}[!t]
  \centering
    \includegraphics[width=0.95\linewidth]{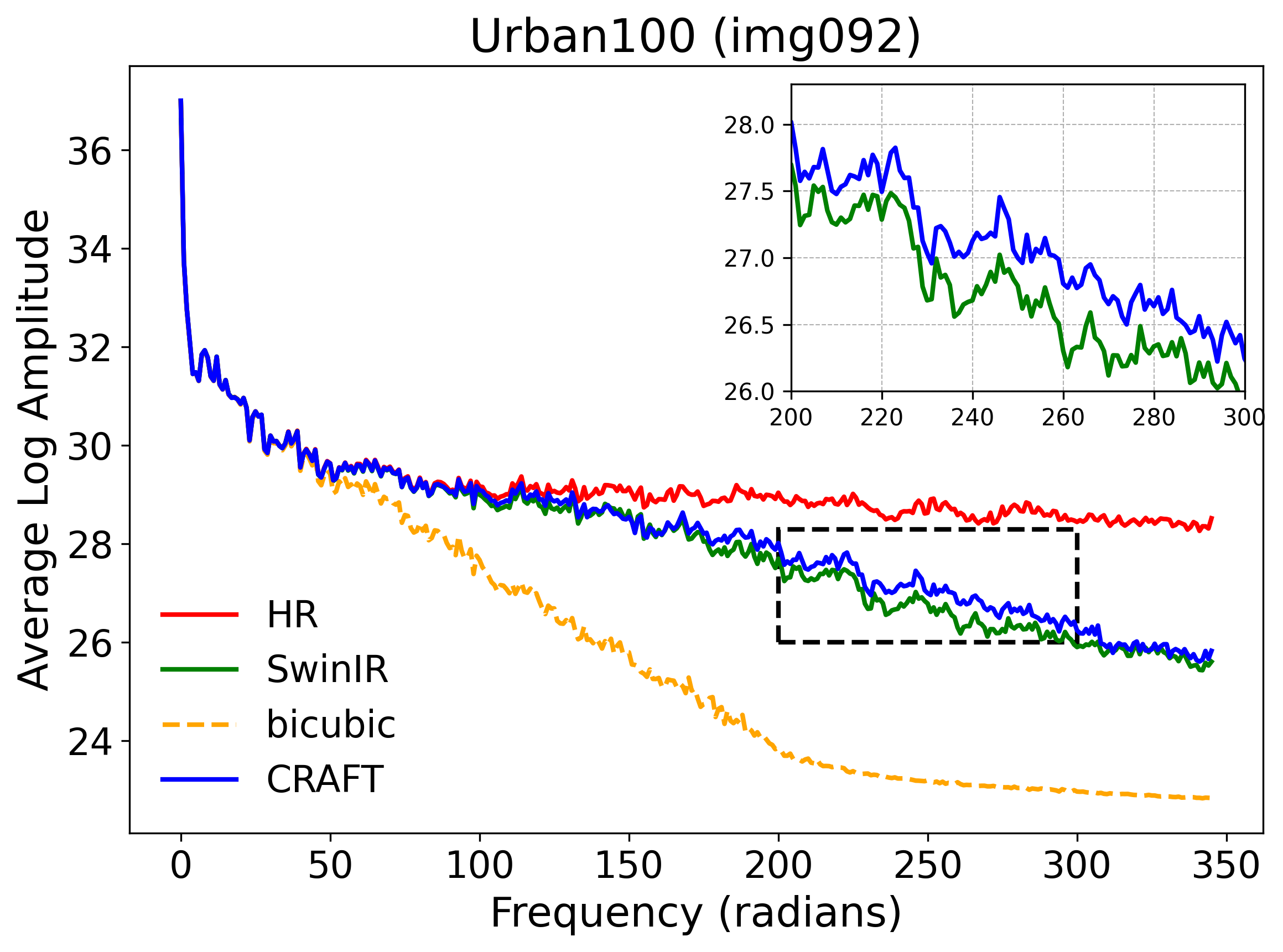}
  \vspace{-3mm}
    \caption{{Comparison of average log-amplitude spectra with different methods. The results clearly demonstrate that our CRAFT method effectively restores high-frequency components.}}
    \vspace{-4mm}
  \label{fig:quant_log_power_frequency}
\end{figure}

%% file: tables/ptq_cmp_sota.tex
\begin{table*}[!t]
  \caption{{PSNR/SSIM comparison between existing PTQ methods and ours by applying them to CRAFT for ${\times 4}$ SISR. ``W/A'' represents weight and activation bit-width, respectively. The symbol $\dag$ indicates our reproduced results.}}
  \vspace{-2mm}
  \centering
  \footnotesize 
  \setlength{\tabcolsep}{2.2mm}
  \begin{tabular}{cccccccccc}
      \toprule
      Scale & Model & W/A  & Set5 & Set14 & BSD100 & Urban100 & Manga109 & DIV2K & LSDIR\\
      \midrule
      \multirow{13}{*}{${\times 4}$} & Full-Precision & 32/32 & 32.52/.8989 & 28.85/.7872 & 27.72/.7418 & 26.56/.7995 & 31.18/.9168 & 30.67/.8428 & 26.43/.7685\\
      
      \cdashline{2-10}
      
      & MinMax \cite{jacob2018quantization} & 8/8 & 32.39/.8950 & 28.77/.7837 & 27.65/.7384 & \textcolor{red}{26.49}/.7957 & 30.95/.9119 & 30.54/.8388 & 26.37/.7650\\
      & Percentile \cite{li2019fully} & 8/8 & 32.42/.8957 & 28.75/.7843 & 27.66/.7392 & 26.42/.7960 & 30.72/.9111 & 30.54/.8394 & 26.35/.7655\\
      & PTQ4SR{$^\dag$}~\cite{tu2023toward} & 8/8 & 32.33/.8954 & 28.69/.7834 & 27.62/.7380 & 26.34/.7927 & 30.81/.9119 & 30.50/.8386 & 26.31/.7633\\
      & Ours & 8/8 & \textcolor{red}{32.45/.8965} & \textcolor{red}{28.80/.7850} & \textcolor{red}{27.68/.7397} & \textcolor{red}{26.49/.7967} & \textcolor{red}{31.06/.9141} & \textcolor{red}{30.59/.8401} & \textcolor{red}{26.39/.7660}  \\
      
      \cdashline{2-10}
      
      & MinMax \cite{jacob2018quantization} & 6/6 & 31.19/.8556 & 27.98/.7502 & 27.13/.7085 & 25.81/.7622 & 29.62/.8713 & 29.56/.8025 & 25.85/.7319\\
      & Percentile \cite{li2019fully} & 6/6 & 31.29/.8597 & 28.08/.7536 & 27.19/.7112 & 25.80/.7605 & 29.52/.8640 & 25.90/.7316 & 25.90/.7316\\
      & PTQ4SR{$^\dag$}~\cite{tu2023toward} & 6/6 & 31.53/.8749 & 28.15/.7653 & 27.23/.7213 & 25.60/.7654 & 28.93/.8791 & 29.68/.8157 & 25.90/.7422 \\
      & Ours & 6/6 & \textcolor{red}{32.01/.8820} & \textcolor{red}{28.51/.7717} & \textcolor{red}{27.49/.7272} & \textcolor{red}{26.16/.7808} & \textcolor{red}{30.11/.8922} & \textcolor{red}{30.18/.8246} & \textcolor{red}{26.17/.7508} \\
      
      \cdashline{2-10}
      
      & MinMax \cite{jacob2018quantization} & 4/4 & 21.14/.3715 & 19.84/.2979 & 20.07/.2771 & 18.58/.3041 & 21.06/.4174 & 20.37/.2862 & 19.36/.2979\\
      & Percentile \cite{li2019fully} & 4/4 & 20.76/.3407 & 20.44/.3007 & 20.45/.2784 & 19.32/.3046 & 20.51/.3529 & 20.33/.2622 & 19.64/.2806\\
      & PTQ4SR{$^\dag$}~\cite{tu2023toward} & 4/4 & 28.15/.7555 & 26.08/.6591 & 25.82/.6280 & 23.77/.6405 & 24.81/.7149 & 27.47/.6961 & 24.47/.6257 \\
      & Ours & 4/4 & \textcolor{red}{29.46/.7854} & \textcolor{red}{26.92/.6891} & \textcolor{red}{26.32/.6492} & \textcolor{red}{24.50/.6787} & \textcolor{red}{27.03/.7653} & \textcolor{red}{28.10/.7214} & \textcolor{red}{24.97/.6593} \\
      
      \bottomrule
  \end{tabular}
  \label{tab:resultsQuantization}
\end{table*}

%% file: tables/ptq_vsual_cmp.tex
\begin{figure*}[!t]
    \centering
    \footnotesize
    \begin{overpic}[scale=.49]{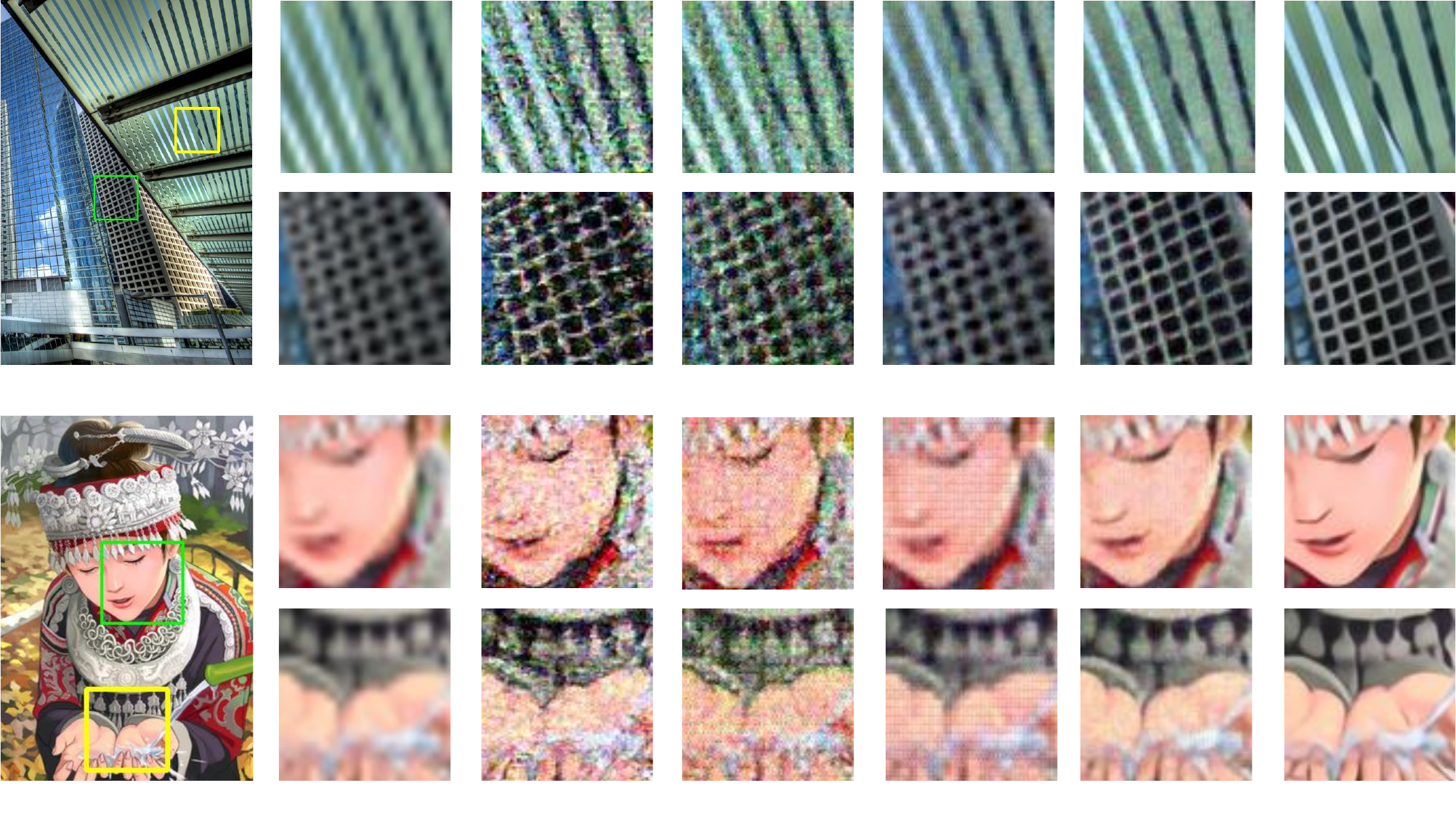}
    \put(4.3,30.2){Img061 ($\times4$)}
    \put(4.3,1.4){Comic ($\times4$)}

    \put(22.5,30.2){Bicubic}
    \put(34.3,30.2){MinMax~\cite{jacob2018quantization}}
    \put(47.5,30.2){Percentile~\cite{li2019fully}}
    \put(62,30.2){PTQ4SR~\cite{tu2023toward}}
    \put(78.6,30.2){Ours}
    \put(89,30.2){Full-Precision}
    
    \put(22.5,1.4){Bicubic}
    \put(34.3,1.4){MinMax~\cite{jacob2018quantization}}
    \put(47.5,1.4){Percentile~\cite{li2019fully}}
    \put(62,1.4){PTQ4SR~\cite{tu2023toward}}
    \put(78,1.4){Ours}
    \put(89.3,1.4){Full-Precision}
    
    \end{overpic}
    \vspace{-3mm}
    \caption{Visual results of different quantization methods on 4-bit CRAFT model with a magnification factor of 4.}
    \vspace{-4mm}
    \label{fig:visulizationQuantization}
\end{figure*}

%% file: tables/ptq_ab_strategy.tex
\begin{table*}[!t]
  \caption{{Efficacy of quantization methods across two stages, using DBDC and PaC from PTQ4SR~\cite{tu2023toward} as baselines for comparison with our ADC and BR strategies. The symbol $\Delta$ represents the PSNR improvement over the baseline methods at each stage. FGO denotes the proposed frequency-guided optimization. The dashed line separates the two stages for clarity.}}
  \vspace{-3mm}
  \centering
  \footnotesize
  \setlength{\tabcolsep}{1.4mm}
  \begin{tabular}{cccccccccc}
      \toprule
      Scale & Model & FGO & Set5 ($\Delta$) & Set14 ($\Delta$) & B100 ($\Delta$) & Urban100 ($\Delta$) & Manga109 ($\Delta$) & DIV2K ($\Delta$) & LSDIR ($\Delta$) \\
      \midrule
      \multirow{6}{*}{$\times4$}& DBDC~\cite{tu2023toward} & \ding{55} & 23.69 & 22.18 & 21.90 & 20.24 & 22.49 & 21.36 & 20.77\\
      & ADC (Ours) & \ding{55} & 26.21 (+2.52) & 24.43 (+2.25) & 24.05 (+2.15) & 22.65 (+2.41) & 23.23 (+0.74) & 24.90 (+3.54) & 23.21 (+2.44)\\
      & ADC (Ours) & \ding{51} & \textcolor{black}{26.51} (+2.82) & \textcolor{black}{24.61} (+2.43) & \textcolor{black}{24.11} (+2.21) & \textcolor{black}{22.77} (+2.53) & \textcolor{black}{24.94} (+2.45) & \textcolor{black}{25.06} (+3.70) & \textcolor{black}{23.30} (+2.53) \\
      \cdashline{2-10}
      & PaC~\cite{tu2023toward} & \ding{55} & 28.94 & 26.52 & 26.03 & 24.10 & 25.63 & 27.89 & 24.76 \\
      & BR (Ours) & \ding{55} & 28.95 (+0.01) & 26.79 (+0.27) & 26.27 (+0.24) & 24.42 (+0.32) & 27.07 (+1.44) & 27.96 (+0.07) & 24.87 (+0.11)\\
      & BR (Ours) & \ding{51} & 29.46 (+0.52) & 26.92 (+0.40) & 26.32 (+0.29) & 24.50 (+0.40) & 27.03 (+1.40) & 28.10 (+0.21) & 24.97 (+0.21)\\
      \bottomrule
  \end{tabular}
  \vspace{-2mm}
  \label{tab:ablationquantstrategy}
\end{table*}

%% file: tables/ptq_ab_dual_clip.tex
\newcommand{\AddImgQuant}[1]{\includegraphics[width=.3\linewidth]{#1}}
\begin{figure*}[!t]
  \AddImgQuant{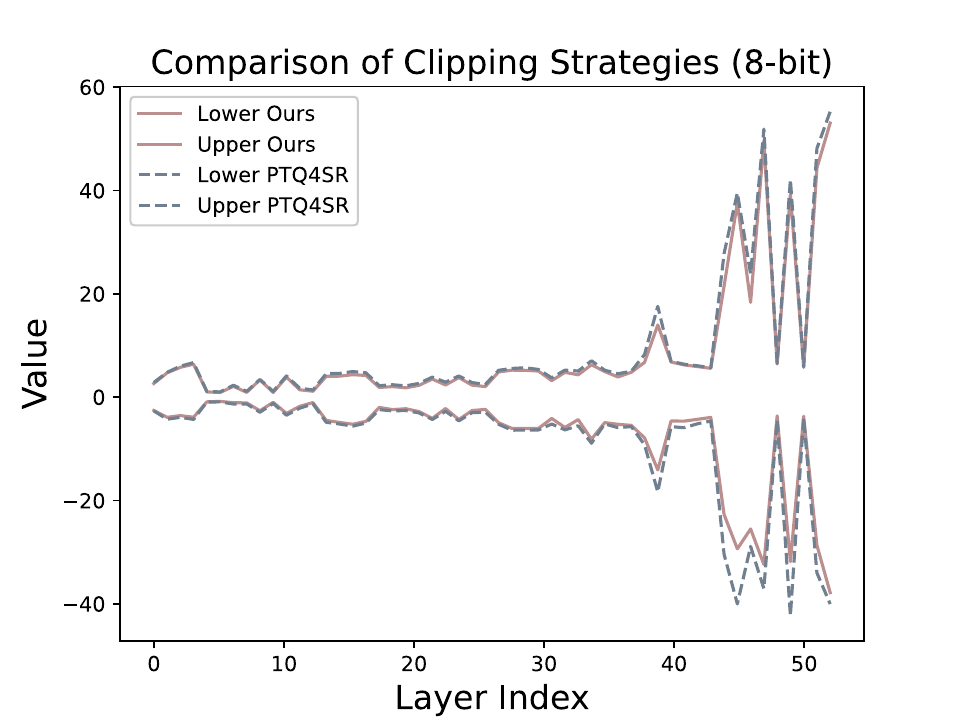} \hfill
  \AddImgQuant{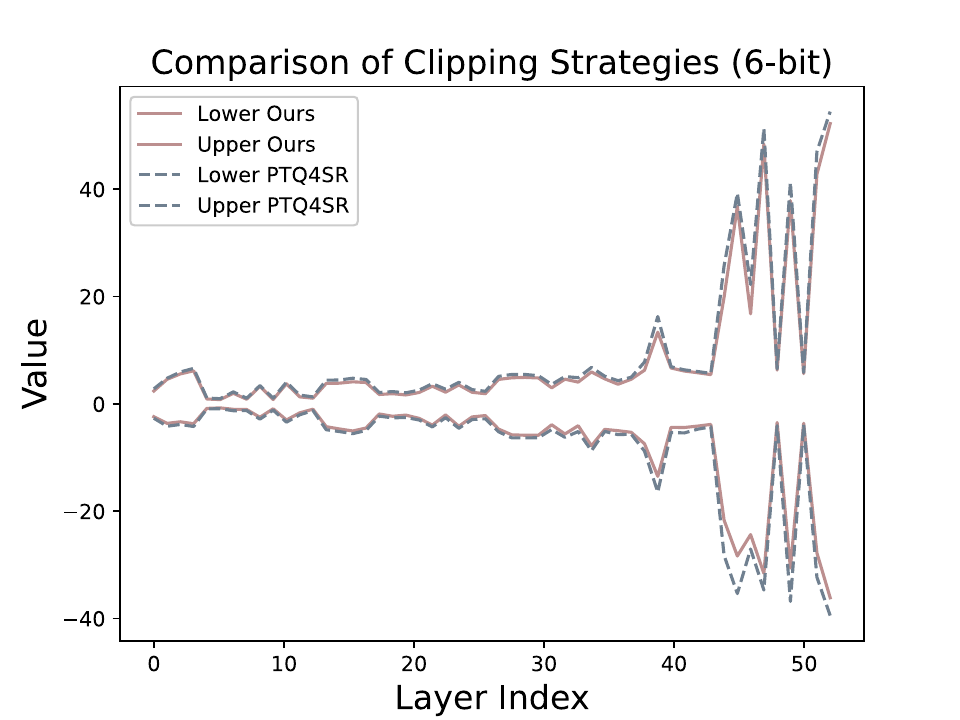} \hfill \AddImgQuant{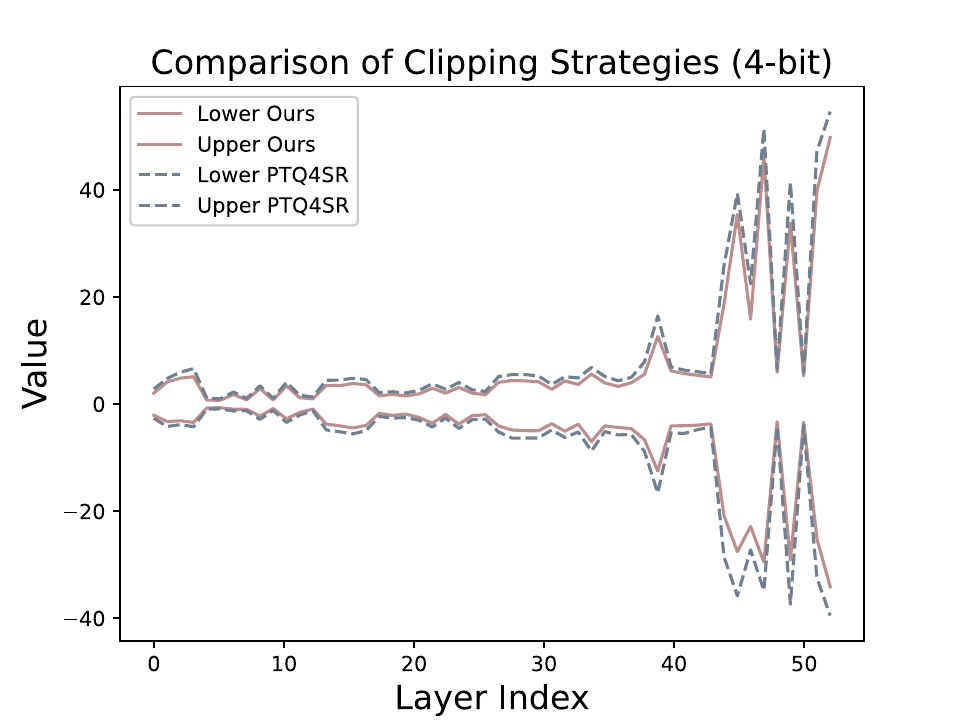}
  \vspace{-3mm}
  \caption{{Clipping strategies comparison between PTQ4SR~\cite{tu2023toward} and our method applied to CRAFT with a magnification factor of 4. The layer index represents progressively deeper layers in the model. From left to right, the results show quantization at 8 bits, 6 bits, and 4 bits, respectively.}}
  \vspace{-3mm}
  \label{fig:boudary_comparison}
\end{figure*}

%% file: tables/ptq_ab_quantization_recon_hf.tex
\renewcommand{\AddImg}[1]{\includegraphics[width=.1\linewidth]{#1}}

\begin{figure*}[!t]
  \centering
  \subfloat[Full-precise]{%
    \AddImg{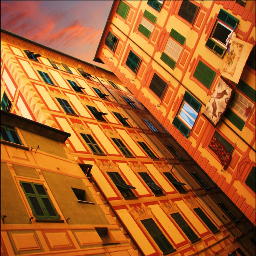}
    \AddImg{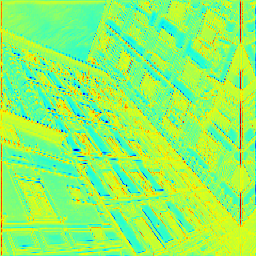}
    \AddImg{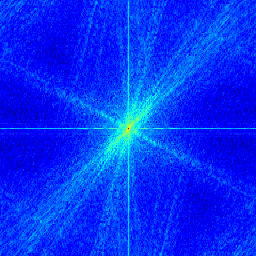}
    \label{fig:filtered_float}
  }
  \hfill
  \subfloat[Ours w/o FGO]{%
    \AddImg{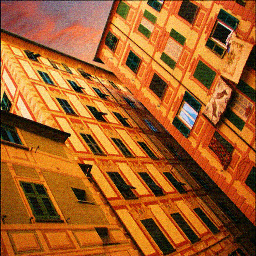}
    \AddImg{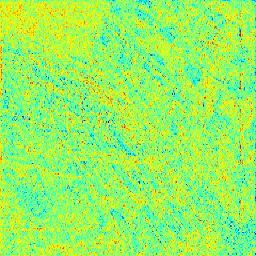}
    \AddImg{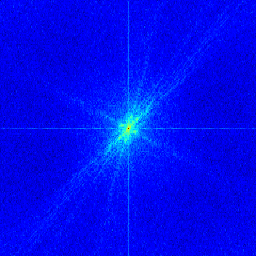}
    \label{fig:filtered_ori}
  }
  \hfill
  \subfloat[Ours]{%
    \AddImg{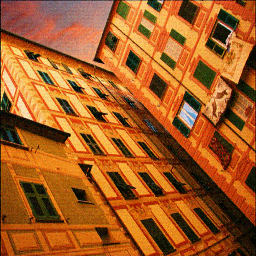}
    \AddImg{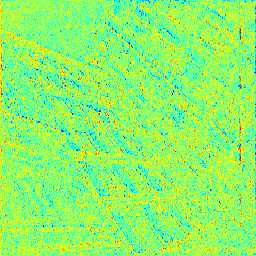}
    \AddImg{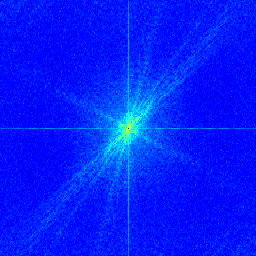}
    \label{fig:filtered_opt}
  }
  \vspace{-2mm}
  \caption{{Visualization of HFERB outputs under different quantization strategies. FGO refers to the proposed frequency-guided optimization. (a) Feature distribution of the full-precision model in both spatial and frequency domains. (b) Output without FGO, showing significant frequency loss. (c) Output after applying FGO, demonstrating notably improved frequency restoration.}}
  \vspace{-1mm}
  \label{fig:ablation_QUANTIZATION}
\end{figure*}

%% file: tables/ptq_ab_general_quant.tex
\begin{table*}[!t]
  \caption{{PSNR/SSIM comparison with transformer-based methods for the magnification factor of 4. All methods are quantized by the proposed general PTQ strategy. ``W/A'' represents weight and activation bit-width, while FGO refers to the proposed frequency-guided optimization.}}
  \vspace{-3mm}
  \centering
  \footnotesize 
  \setlength{\tabcolsep}{2.2mm}
  \begin{tabular}{cccccccccc}
      \toprule
      Scale & Model & W/A & Set5 & Set14 & BSD100 & Urban100 & Manga109 & LSDIR & DIV2K \\
      \midrule
      \multirow{5}{*}{${\times 4}$}& SwinIR~\cite{liang2021swinir} & 8/8 & 32.36/.8952 & 28.71/.7837 & 27.65/.7386 & 26.41/.7951 & 30.76/.9115 & 26.36/.7649 & 30.55/.8402\\
      & CRAFT w/o FGO & 8/8 & 32.45/.8967 & 28.77/.7848 & 27.69/.7398 & 26.47/.7968 & 30.98/.9134 & 26.38/.7659 & 30.58/.8402\\
      \cdashline{2-10}
    
      & SwinIR~\cite{liang2021swinir} & 6/6 & 31.94/.8775 & 28.43/.7686 & 27.45/.7246 & 26.10/.7784 & 29.98/.8876 & 26.16/.7490 & 30.14/.8225  \\
      & CRAFT w/o FGO & 6/6 & 32.00/.8816 & 28.49/.7718 & 27.49/.7276 & 26.14/.7800 & 30.21/.8924 & 26.15/.7504 & 30.18/.8242   \\
      \cdashline{2-10}
      
      & SwinIR~\cite{liang2021swinir} & 4/4 & 28.99/.7583 & 26.59/.6659 & 25.97/.6238 & 24.37/.6586 & 26.49/.7383 & 24.83/.6346 & 27.75/.6897 \\
      & CRAFT w/o FGO & 4/4 & 28.95/.7771 & 26.79/.6892 & 26.27/.6505 & 24.42/.6772 & 27.07/.7731 & 24.87/.6541 & 27.96/.7199\\
      \bottomrule
  \end{tabular}
  \vspace{-3mm}
  \label{tab:ablationgeneralQuantization}
\end{table*}

%% file: extended_craft.bbl
\begin{thebibliography}{10}
\providecommand{\url}[1]{#1}
\csname url@samestyle\endcsname
\providecommand{\newblock}{\relax}
\providecommand{\bibinfo}[2]{#2}
\providecommand{\BIBentrySTDinterwordspacing}{\spaceskip=0pt\relax}
\providecommand{\BIBentryALTinterwordstretchfactor}{4}
\providecommand{\BIBentryALTinterwordspacing}{\spaceskip=\fontdimen2\font plus
\BIBentryALTinterwordstretchfactor\fontdimen3\font minus \fontdimen4\font\relax}
\providecommand{\BIBforeignlanguage}[2]{{%
\expandafter\ifx\csname l@#1\endcsname\relax
\typeout{** WARNING: IEEEtran.bst: No hyphenation pattern has been}%
\typeout{** loaded for the language `#1'. Using the pattern for}%
\typeout{** the default language instead.}%
\else
\language=\csname l@#1\endcsname
\fi
#2}}
\providecommand{\BIBdecl}{\relax}
\BIBdecl

\bibitem{mudunuri2015low}
S.~P. Mudunuri and S.~Biswas, ``Low resolution face recognition across variations in pose and illumination,'' \emph{IEEE Transactions on Pattern Analysis and Machine Intelligence}, vol.~38, no.~5, pp. 1034--1040, 2015.

\bibitem{greenspan2009super}
H.~Greenspan, ``Super-resolution in medical imaging,'' \emph{The Computer Journal}, vol.~52, no.~1, pp. 43--63, 2009.

\bibitem{liang2021swinir}
J.~Liang, J.~Cao, G.~Sun, K.~Zhang, L.~Van~Gool, and R.~Timofte, ``Swinir: Image restoration using swin transformer,'' in \emph{Proceedings of the IEEE/CVF Conference on Computer Vision Workshop (ICCVW)}, 2021, pp. 1833--1844.

\bibitem{Ledig2017}
C.~Ledig, L.~Theis, F.~Husz{\'{a}}r, J.~Caballero, A.~Cunningham, A.~Acosta, A.~Aitken, A.~Tejani, J.~Totz, Z.~Wang, and W.~Shi, ``Photo-realistic single image super-resolution using a generative adversarial network,'' in \emph{Proceedings of the IEEE/CVF Conference on Computer Vision and Pattern Recognition (CVPR)}, 2017, pp. 105--114.

\bibitem{Zhang2017}
Z.~Zhang and V.~Sze, ``Fast: A framework to accelerate super-resolution processing on compressed videos,'' in \emph{Proceedings of the IEEE/CVF Conference on Computer Vision and Pattern Recognition (CVPR)}, 2017, pp. 1015--1024.

\bibitem{dong2016srcnn}
C.~Dong, C.~C. Loy, K.~He, and X.~Tang, ``Image super-resolution using deep convolutional networks,'' \emph{IEEE Transactions on Pattern Analysis and Machine Intelligence}, vol.~38, no.~2, pp. 295--307, 2016.

\bibitem{zhang2021edge}
X.~Zhang, H.~Zeng, and L.~Zhang, ``Edge-oriented convolution block for real-time super resolution on mobile devices,'' in \emph{Proceedings of the ACM International Conference on Multimedia (ACMMM)}, 2021, pp. 4034--4043.

\bibitem{Liang2021}
J.~Liang and L.~V. Gool, ``{Flow-based Kernel Prior with Application to Blind Super-Resolution},'' in \emph{Proceedings of the IEEE/CVF Conference on Computer Vision and Pattern Recognition (CVPR)}, 2021, pp. 10\,601--10\,610.

\bibitem{ledig2017photo}
C.~Ledig, L.~Theis, F.~Husz{\'a}r, J.~Caballero, A.~Cunningham, A.~Acosta, A.~Aitken, A.~Tejani, J.~Totz, Z.~Wang \emph{et~al.}, ``Photo-realistic single image super-resolution using a generative adversarial network,'' in \emph{Proceedings of the IEEE/CVF Conference on Computer Vision and Pattern Recognition (CVPR)}, 2017, pp. 4681--4690.

\bibitem{Tong2017iccv}
T.~Tong, G.~Li, X.~Liu, and Q.~Gao, ``{Image Super-Resolution Using Dense Skip Connections},'' in \emph{Proceedings of the IEEE/CVF Conference on Computer Vision (ICCV)}, 2017, pp. 4809--4817.

\bibitem{Zhang2018ECCV}
Y.~Zhang, K.~Li, K.~Li, L.~Wang, B.~Zhong, and Y.~Fu, ``Image super-resolution using very deep residual channel attention networks,'' in \emph{Proceedings of the European Conference on Computer Vision (ECCV)}, 2018, pp. 294--310.

\bibitem{Xia2022}
B.~Xia, Y.~Hang, Y.~Tian, W.~Yang, Q.~Liao, and J.~Zhou, ``Efficient non-local contrastive attention for image super-resolution,'' in \emph{Proceedings of the Association for the Advancement of Artificial Intelligence (AAAI)}, 2022, pp. 2759--2767.

\bibitem{Mei2021cvpr}
Y.~Mei, Y.~Fan, and Y.~Zhou, ``{Image Super-Resolution with Non-Local Sparse Attention},'' in \emph{Proceedings of the IEEE/CVF Conference on Computer Vision and Pattern Recognition (CVPR)}, 2021, pp. 3517--3526.

\bibitem{Mei2020cvpr}
Y.~Mei, Y.~Fan, Y.~Zhou, L.~Huang, T.~S. Huang, and H.~Shi, ``Image super-resolution with cross-scale non-local attention and exhaustive self-exemplars mining,'' in \emph{Proceedings of the IEEE/CVF Conference on Computer Vision and Pattern Recognition (CVPR)}, 2020.

\bibitem{Chen2022nips}
Z.~Chen, Y.~Zhang, J.~Gu, Y.~Zhang, and L.~Kong, ``Cross aggregation transformer for image restoration,'' in \emph{Proceedings of the Conference and Workshop on Neural Information Processing Systems (NeurIPS)}, 2022.

\bibitem{chen2021pre}
H.~Chen, Y.~Wang, T.~Guo, C.~Xu, Y.~Deng, Z.~Liu, S.~Ma, C.~Xu, C.~Xu, and W.~Gao, ``Pre-trained image processing transformer,'' in \emph{Proceedings of the IEEE/CVF Conference on Computer Vision and Pattern Recognition (CVPR)}, 2021, pp. 12\,299--12\,310.

\bibitem{li2021efficient}
W.~Li, X.~Lu, J.~Lu, X.~Zhang, and J.~Jia, ``On efficient transformer and image pre-training for low-level vision,'' \emph{arXiv preprint arXiv:2112.10175}, 2021.

\bibitem{lu2022transformer}
Z.~Lu, J.~Li, H.~Liu, C.~Huang, L.~Zhang, and T.~Zeng, ``Transformer for single image super-resolution,'' in \emph{Proceedings of the IEEE/CVF Conference on Computer Vision and Pattern Recognition Workshop (CVPRW)}, 2022, pp. 457--466.

\bibitem{li2023feature}
A.~Li, L.~Zhang, Y.~Liu, and C.~Zhu, ``Feature modulation transformer: Cross-refinement of global representation via high-frequency prior for image super-resolution,'' in \emph{Proceedings of the IEEE/CVF International Conference on Computer Vision (ICCV)}, 2023, pp. 12\,514--12\,524.

\bibitem{Lim2017}
B.~Lim, S.~Son, H.~Kim, S.~Nah, and K.~M. Lee, ``Enhanced deep residual networks for single image super-resolution,'' in \emph{Proceedings of the IEEE/CVF Conference on Computer Vision and Pattern Recognition Workshop (CVPRW)}, 2017, pp. 1132--1140.

\bibitem{li2020pams}
H.~Li, C.~Yan, S.~Lin, X.~Zheng, B.~Zhang, F.~Yang, and R.~Ji, ``Pams: Quantized super-resolution via parameterized max scale,'' in \emph{Proceedings of the European Conference on Computer Vision (ECCV)}, 2020, pp. 564--580.

\bibitem{wang2021fully}
H.~Wang, P.~Chen, B.~Zhuang, and C.~Shen, ``Fully quantized image super-resolution networks,'' in \emph{Proceedings of the 29th ACM International Conference on Multimedia (ACMMM)}, 2021, pp. 639--647.

\bibitem{hong2022cadyq}
C.~Hong, S.~Baik, H.~Kim, S.~Nah, and K.~M. Lee, ``Cadyq: Content-aware dynamic quantization for image super-resolution,'' in \emph{Proceedings of the European Conference on Computer Vision (ECCV)}.\hskip 1em plus 0.5em minus 0.4em\relax Springer, 2022, pp. 367--383.

\bibitem{hong2022daq}
C.~Hong, H.~Kim, S.~Baik, J.~Oh, and K.~M. Lee, ``Daq: Channel-wise distribution-aware quantization for deep image super-resolution networks,'' in \emph{Proceedings of the IEEE/CVF Winter Conference on Applications of Computer Vision (WACV)}, 2022, pp. 2675--2684.

\bibitem{zhong2022dynamic}
Y.~Zhong, M.~Lin, X.~Li, K.~Li, Y.~Shen, F.~Chao, Y.~Wu, and R.~Ji, ``Dynamic dual trainable bounds for ultra-low precision super-resolution networks,'' in \emph{Proceedings of the European Conference on Computer Vision (ECCV)}, 2022, pp. 1--18.

\bibitem{tu2023toward}
Z.~Tu, J.~Hu, H.~Chen, and Y.~Wang, ``Toward accurate post-training quantization for image super resolution,'' in \emph{Proceedings of the IEEE/CVF Conference on Computer Vision and Pattern Recognition (CVPR)}, 2023, pp. 5856--5865.

\bibitem{kim2016accurate}
J.~Kim, J.~K. Lee, and K.~M. Lee, ``Accurate image super-resolution using very deep convolutional networks,'' in \emph{Proceedings of the IEEE/CVF Conference on Computer Vision and Pattern Recognition (CVPR)}, 2016, pp. 1646--1654.

\bibitem{ioffe2015batch}
S.~Ioffe and C.~Szegedy, ``Batch normalization: Accelerating deep network training by reducing internal covariate shift,'' in \emph{Proceedings of the International Conference on Machine Learning (ICML)}, 2015, pp. 448--456.

\bibitem{Ahn2018}
N.~Ahn, B.~Kang, and K.~A. Sohn, ``Fast, accurate, and lightweight super-resolution with cascading residual network,'' in \emph{Proceedings of the European Conference on Computer Vision (ECCV)}, 2018, pp. 256--272.

\bibitem{Hui2019}
Z.~Hui, Y.~Yang, X.~Gao, and X.~Wang, ``Lightweight image super-resolution with information multi-distillation network,'' in \emph{Proceedings of the ACM International Conference on Multimedia (ACMMM)}, 2019, pp. 2024--2032.

\bibitem{Li2020}
W.~Li, K.~Zhou, L.~Qi, N.~Jiang, J.~Lu, and J.~Jia, ``Lapar: Linearly-assembled pixel-adaptive regression network for single image super-resolution and beyond,'' in \emph{Proceedings of the Conference and Workshop on Neural Information Processing Systems (NeurIPS)}, 2020, pp. 1--13.

\bibitem{sun2022hybrid}
B.~Sun, Y.~Zhang, S.~Jiang, and Y.~Fu, ``Hybrid pixel-unshuffled network for lightweight image super-resolution,'' in \emph{Proceedings of the Association for the Advancement of Artificial Intelligence (AAAI)}, 2023.

\bibitem{liu2021swin}
Z.~Liu, Y.~Lin, Y.~Cao, H.~Hu, Y.~Wei, Z.~Zhang, S.~Lin, and B.~Guo, ``Swin transformer: Hierarchical vision transformer using shifted windows,'' in \emph{Proceedings of the IEEE/CVF Conference on Computer Vision (ICCV)}, 2021, pp. 10\,012--10\,022.

\bibitem{zhang2022efficient}
X.~Zhang, H.~Zeng, S.~Guo, and L.~Zhang, ``Efficient long-range attention network for image super-resolution,'' in \emph{Proceedings of the European Conference on Computer Vision (ECCV)}, 2022.

\bibitem{chen2023activating}
X.~Chen, X.~Wang, J.~Zhou, Y.~Qiao, and C.~Dong, ``Activating more pixels in image super-resolution transformer,'' in \emph{Proceedings of the IEEE/CVF Conference on Computer Vision and Pattern Recognition (CVPR)}, 2023, pp. 22\,367--22\,377.

\bibitem{ma2019efficient}
Y.~Ma, H.~Xiong, Z.~Hu, and L.~Ma, ``Efficient super resolution using binarized neural network,'' in \emph{Proceedings of the IEEE/CVF Conference on Computer Vision and Pattern Recognition Workshops (CVPRW)}, 2019.

\bibitem{xin2020binarized}
J.~Xin, N.~Wang, X.~Jiang, J.~Li, H.~Huang, and X.~Gao, ``Binarized neural network for single image super resolution,'' in \emph{Proceedings of the European Conference on Computer Vision (ECCV)}, 2020, pp. 91--107.

\bibitem{Shi2016}
W.~Shi, J.~Caballero, F.~Husz{\'{a}}r, J.~Totz, A.~Aitken, R.~Bishop, D.~Rueckert, and Z.~Wang, ``Real-time single image and video super-resolution using an efficient sub-pixel convolutional neural network,'' in \emph{Proceedings of the IEEE/CVF Conference on Computer Vision and Pattern Recognition (CVPR)}, 2016, pp. 1874--1883.

\bibitem{rossmann1969point}
K.~Rossmann, ``Point spread-function, line spread-function, and modulation transfer function: tools for the study of imaging systems,'' \emph{Radiology}, vol.~93, no.~2, pp. 257--272, 1969.

\bibitem{shechtman2014optimal}
Y.~Shechtman, S.~J. Sahl, A.~S. Backer, and W.~E. Moerner, ``Optimal point spread function design for 3d imaging,'' \emph{Physical review letters}, vol. 113, no.~13, p. 133902, 2014.

\bibitem{wang2021crossformericlr}
W.~Wang, L.~Yao, L.~Chen, B.~Lin, D.~Cai, X.~He, and W.~Liu, ``Crossformer: A versatile vision transformer hinging on cross-scale attention,'' in \emph{Proceedings of the International Conference on Learning Representations (ICLR)}, 2022.

\bibitem{Zamir2022}
S.~W. Zamir, A.~Arora, S.~Khan, M.~Hayat, F.~S. Khan, and M.-H. Yang, ``Restormer: Efficient transformer for high-resolution image restoration,'' in \emph{Proceedings of the IEEE/CVF Conference on Computer Vision and Pattern Recognition (CVPR)}, 2022, pp. 5718--5729.

\bibitem{jacob2018quantization}
B.~Jacob, S.~Kligys, B.~Chen, M.~Zhu, M.~Tang, A.~Howard, H.~Adam, and D.~Kalenichenko, ``Quantization and training of neural networks for efficient integer-arithmetic-only inference,'' in \emph{Proceedings of the IEEE/CVF Conference on Computer Vision and Pattern Recognition (CVPR)}, 2018, pp. 2704--2713.

\bibitem{bengio2013estimating}
Y.~Bengio, N.~L{\'e}onard, and A.~Courville, ``Estimating or propagating gradients through stochastic neurons for conditional computation,'' \emph{arXiv preprint arXiv:1308.3432}, 2013.

\bibitem{luo2020latticenet}
X.~Luo, Y.~Xie, Y.~Zhang, Y.~Qu, C.~Li, and Y.~Fu, ``Latticenet: Towards lightweight image super-resolution with lattice block,'' in \emph{Proceedings of the European Conference on Computer Vision (ECCV)}, 2020, pp. 272--289.

\bibitem{gao2022lightweight}
G.~Gao, Z.~Wang, J.~Li, W.~Li, Y.~Yu, and T.~Zeng, ``Lightweight bimodal network for single-image super-resolution via symmetric cnn and recursive transformer,'' \emph{Proceedings of the International Joint Conference on Artificial Intelligence (IJCAI)}, 2022.

\bibitem{agustsson2017ntire}
E.~Agustsson and R.~Timofte, ``Ntire 2017 challenge on single image super-resolution: Dataset and study,'' in \emph{Proceedings of the IEEE/CVF Conference on Computer Vision and Pattern Recognition Workshop (CVPRW)}, 2017, pp. 126--135.

\bibitem{bevilacqua2012low}
M.~Bevilacqua, A.~Roumy, C.~Guillemot, and M.~L. Alberi-Morel, ``Low-complexity single-image super-resolution based on nonnegative neighbor embedding,'' in \emph{Proceedings of the British Machine Vision Conference (BMVC)}, 2012, pp. 1--10.

\bibitem{zeyde2010single}
R.~Zeyde, M.~Elad, and M.~Protter, ``On single image scale-up using sparse-representations,'' in \emph{International Conference on Curves and Surfaces (ICCS)}, 2010, pp. 711--730.

\bibitem{martin2001database}
D.~Martin, C.~Fowlkes, D.~Tal, and J.~Malik, ``A database of human segmented natural images and its application to evaluating segmentation algorithms and measuring ecological statistics,'' in \emph{Proceedings of the IEEE/CVF Conference on Computer Vision (ICCV)}, 2001, pp. 416--423.

\bibitem{huang2015single}
J.-B. Huang, A.~Singh, and N.~Ahuja, ``Single image super-resolution from transformed self-exemplars,'' in \emph{Proceedings of the IEEE/CVF Conference on Computer Vision and Pattern Recognition (CVPR)}, 2015, pp. 5197--5206.

\bibitem{matsui2017sketch}
Y.~Matsui, K.~Ito, Y.~Aramaki, A.~Fujimoto, T.~Ogawa, T.~Yamasaki, and K.~Aizawa, ``Sketch-based manga retrieval using manga109 dataset,'' \emph{Multimedia Tools and Applications}, vol.~76, no.~20, pp. 21\,811--21\,838, 2017.

\bibitem{wang2004image}
Z.~Wang, A.~C. Bovik, H.~R. Sheikh, and E.~P. Simoncelli, ``Image quality assessment: from error visibility to structural similarity,'' \emph{IEEE Transactions on Image Processing}, vol.~13, no.~4, pp. 600--612, 2004.

\bibitem{li2023lsdir}
Y.~Li, K.~Zhang, J.~Liang, J.~Cao, C.~Liu, R.~Gong, Y.~Zhang, H.~Tang, Y.~Liu, D.~Demandolx \emph{et~al.}, ``Lsdir: A large scale dataset for image restoration,'' in \emph{Proceedings of the IEEE/CVF Conference on Computer Vision and Pattern Recognition Workshop (CVPRW)}, 2023, pp. 1775--1787.

\bibitem{gu2021interpreting}
J.~Gu and C.~Dong, ``Interpreting super-resolution networks with local attribution maps,'' in \emph{Proceedings of the IEEE/CVF Conference on Computer Vision and Pattern Recognition (CVPR)}, 2021, pp. 9199--9208.

\bibitem{li2019fully}
R.~Li, Y.~Wang, F.~Liang, H.~Qin, J.~Yan, and R.~Fan, ``Fully quantized network for object detection,'' in \emph{Proceedings of the IEEE/CVF Conference on Computer Vision and Pattern Recognition (CVPR)}, 2019, pp. 2810--2819.

\bibitem{son2022real}
H.~Son, J.~Lee, S.~Cho, and S.~Lee, ``Real-time video deblurring via lightweight motion compensation,'' in \emph{Computer Graphics Forum}, vol.~41, no.~7.\hskip 1em plus 0.5em minus 0.4em\relax Wiley Online Library, 2022, pp. 177--188.

\bibitem{micikevicius2017mixed}
P.~Micikevicius, S.~Narang, J.~Alben, G.~Diamos, E.~Elsen, D.~Garcia, B.~Ginsburg, M.~Houston, O.~Kuchaiev, G.~Venkatesh \emph{et~al.}, ``Mixed precision training,'' \emph{arXiv preprint arXiv:1710.03740}, 2017.

\end{thebibliography}
